%% file: camera_ready_1_main.tex
\documentclass{article}

\usepackage[main, final]{style/neurips_2026/neurips_2026}

\usepackage[utf8]{inputenc} % allow utf-8 input
\usepackage[T1]{fontenc}    % use 8-bit T1 fonts
\usepackage{hyperref}       % hyperlinks
\usepackage{url}            % simple URL typesetting
\usepackage{booktabs}       % professional-quality tables
\usepackage{amsfonts}       % blackboard math symbols
\usepackage{nicefrac}       % compact symbols for 1/2, etc.
\usepackage{microtype}      % microtypography
\input{math_commands.tex}

\usepackage{multirow}
\usepackage{tablefootnote}
\usepackage{makecell}
\usepackage{enumitem}
\usepackage{soul}
\usepackage[export]{adjustbox}
\usepackage{bbm}
\usepackage{bbding}
\usepackage{minitoc}
\usepackage{graphicx}
\usepackage[table, svgnames]{xcolor}
\usepackage{subfig}
\usepackage{cleveref}
\usepackage{tikz}
\usepackage{mathtools}
\usepackage{tcolorbox}
\usepackage{lscape}
\usepackage{wrapfig}
\usepackage{longtable}

\usepackage{amsmath}
\usepackage{amssymb}
\usepackage{mathtools}
\usepackage{amsthm}

\usepackage{algorithm} 
\usepackage{algorithmic}
\usepackage[algo2e]{algorithm2e} 

\usepackage{silence}
\crefname{appendix}{Appendix}{Appendices}
\newcommand{\modelname}{\mbox{TabFORGE}\xspace}
\theoremstyle{plain}

\theoremstyle{definition}

\theoremstyle{remark}

\newcommand{\titlecontent}{\modelname: A Tabular Foundation Model\\for Generative Modelling}
\title{\titlecontent}

\author{%
  Xiangjian Jiang\textsuperscript{1},
  \space Mingxuan Liu\textsuperscript{2},
  \space Nikola Simidjievski\textsuperscript{3,1},
  \space Tassilo Klein\textsuperscript{2},
  \space Mateja Jamnik\textsuperscript{1} \\
  \textsuperscript{1}Department of Computer Science and Technology, University of Cambridge, UK \\
  \textsuperscript{2}SAP SE
  \space \textsuperscript{3}Télécom Paris, Institut Polytechnique de Paris, France \\
  {\scriptsize \texttt{xj265@cam.ac.uk, mingxuan.liu01@sap.com}} \\ {\scriptsize \texttt{nikola.simidjievski@ip-paris.fr, tassilo.klein@sap.com, mateja.jamnik@cl.cam.ac.uk}} \\
}

\begin{document}
\doparttoc
\faketableofcontents

\maketitle

\vspace{-6mm}
\begin{abstract}
\vspace{-1mm}
% {
Generative modelling is a demanding test of foundation models, because it requires robust, holistic representation learning for a given data modality, rather than optimisation for a supervised prediction target alone. While recent work on tabular foundation models has achieved remarkable progress in predictive modelling, generative tabular foundation models remain underexplored. Existing tabular foundation generators, in particular, have not yet consistently matched strong dataset-specific generators in synthetic data quality.
A key reason is their insufficient alignment with the inter-feature dependency structure of tabular data, which can reflect underlying causal relationships.
In this paper, we address this gap by introducing a novel tabular foundation model, \textbf{\modelname}, built on pretrained \textbf{Tab}ular \textbf{FO}undational \textbf{R}epresentations for \textbf{GE}neration. \modelname is designed to utilise transferable representations induced by a pretrained structure-aware feature encoder, which captures information useful for characterising inter-feature dependencies.
It further decouples latent modelling from decoding through a two-stage design: we first pretrain a score-based diffusion transformer, and then pretrain a denoising-aligned decoder using the denoised latent embeddings. This design elegantly mitigates the distribution shifts in latent embeddings that typically arise between training and inference.
We evaluate \modelname comprehensively against 22 benchmark methods on 45 real-world datasets. Our results show that \modelname effectively learns and leverages generalisable tabular representations, enabling efficient generation of high-quality synthetic tabular data, particularly with strong structural fidelity.
% Code is available at \textcolor{blue}{\url{https://github.com/SilenceX12138/TabFORGE}}.
% }
\end{abstract}

\vspace{-2mm}
\section{Introduction}
\vspace{-1.5mm}
\label{section:intro}
% {
Tabular data is ubiquitous across a wide range of real-world applications~\citep{borisov2022deep, shwartz2022tabular, gorishniy2021revisiting, jiang2026representation, somvanshi2026survey}, spanning domains such as healthcare~\citep{jiang2024protogate} and scientific discovery~\citep{margeloiu2024tabebm}. However, acquiring high-quality tabular data can often be intricate and expensive~\citep{hernandez2022synthetic}, which underscores the need for powerful generative methods tailored to the tabular modality.
Generative modelling has long been recognised as challenging~\citep{bond2021deep, salakhutdinov2015learning}, as it requires not merely recognising patterns of a single predictive target, but the learning of highly generalisable representations that capture the comprehensive data structures~\citep{jiang2026tabstruct}. As such, conventional dataset-specific {\em tabular} generators~\citep{xu2019modeling, shi2025tabdiff}, typically trained from scratch on each individual dataset, often yield suboptimal performance because they are unable to improve generalisability with transferable knowledge across diverse datasets~\citep{van2024position}.
In contrast, for other modalities like text and images, several generative foundation models have been developed to address this limitation~\citep{guo2025deepseek, lu2025vision}. As defined in prior studies~\citep{bommasani2021opportunities, van2024position}, these foundation models are first pretrained on broad upstream datasets and then fitted on downstream datasets via mechanisms such as finetuning~\citep{anisuzzaman2025fine, hu2022lora} and in-context learning~\citep{grinsztajn2025tabpfn, qu2026tabiclv2}.
This success in text and vision highlights the importance of exploring foundation generators for tabular data.
% }

\begin{figure}[!t]
    \centering
    \includegraphics[width=\linewidth]{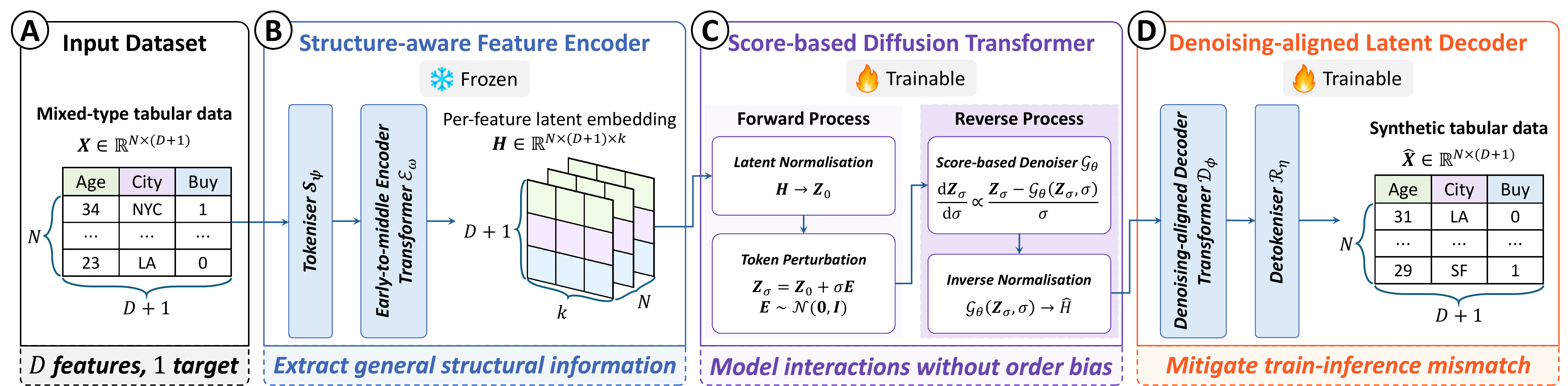}
    % \vspace{-5mm}
    \caption{\textbf{The architecture of \modelname}. \textbf{(A)}~Given a tabular dataset with $D$ features and one target, \modelname treats the target as an additional feature for generative modelling. \textbf{(B)}~The frozen structure-aware feature encoder maps the table into per-feature latent embeddings by first tokenising the features and then contextualising them with inter-feature dependencies. \textbf{(C)}~The trainable score-based diffusion transformer learns the latent distribution by denoising noisy latent embeddings while preserving the token structure. \textbf{(D)}~Once the diffusion transformer has been fitted, the trainable denoising-aligned latent decoder is trained to map the denoised embeddings to the original feature space. During inference, noisy latent embeddings are denoised and decoded into synthetic data.}
    \label{fig:framework}
\vspace{-6mm}
\end{figure}

% {
Prior work~\citep{borisov2022language, lin2025ctsyn, grinsztajn2025tabpfn} has attempted to address tabular data generation via tabular foundation models. However, existing foundation models still struggle to match strong dataset-specific generators and exhibit three primary limitations stemming from insufficient alignment with the inter-feature dependency structure of tabular data:
\textit{(i)~Misaligned generative process.} Some models~\citep{borisov2022language, grinsztajn2025tabpfn} treat tables as a sequential modality (i.e., each sample as a sentence and each feature as a token), thereby performing autoregressive generation with large language models (LLMs) or tabular foundation predictors. However, autoregressive generation inevitably introduces feature-order bias. This can be problematic because the imposed order may deviate substantially from the topological order derived by the underlying causal structures~\citep{tugnoli2026improving, jiang2026tabstruct}. 
\textit{(ii)~Neglect of global dependency structure.} Other studies~\citep{margeloiu2024tabebm, ma2023tabpfgen} construct Energy-Based Models (EBMs) using tabular foundation predictors. However, existing EBMs are inherently biased towards the local dependency structures surrounding the prediction target, while largely overlooking the interactions across all features~\citep{jiang2026tabstruct}.
\textit{(iii)~Limited handling of structural information and latent space.} Earlier work~\citep{lin2025ctsyn} proposes pretraining a cross-dataset variational autoencoder (VAE) for tabular data, along with a latent diffusion model conditioned on LLM embeddings. Despite their susceptibility to metadata quality and LLM capability~\citep{lin2025ctsyn}, they may obscure the inter-feature dependency structures by obscuring feature identity with a predetermined fixed-length query token sequence. Moreover, the latent decoder is trained on clean embeddings by the encoder, whereas at inference time, it receives denoised embeddings from the diffusion model. Such training-inference shifts further degrade model performance.
% }

% {
\vspace{-0.75mm}
\looseness-1
In this paper, we aim to bridge these gaps by introducing \modelname, a novel generative tabular foundation model.
\modelname is distinguished by three complementary components:
(i)~\modelname constructs a transferable latent space with per-feature embeddings using a frozen Prior-Data Fitted Network (PFN) encoder. 
We refer to this latent space as \textit{structure-aware} to describe the motivation behind this design choice: prior evidence~\citep{swelam2025does} has shown that intermediate PFN representations contain information useful for recovering inter-feature dependency structures.
Since such dependencies are central to the data-generating processes underlying tabular data~\citep{zhang2025mitra, jiang2026tabstruct}, we hypothesise that incorporating these structural signals can provide an effective inductive bias for tabular data generation.
% Thus, \modelname utilises the early-to-middle layers of a trained PFN model and further freezes them to avoid latent collapse or overfitting to any specific dataset.
(ii)~\modelname then models interactions among these per-feature latent embeddings using a score-based diffusion transformer pretrained across diverse datasets.
This diffusion-based modelling allows \modelname to capture global data structures effectively~\citep{jiang2026tabstruct}, without imposing a potentially misaligned feature order.
(iii)~\modelname is robust to the latent distribution shifts between training and inference phases. As the latent space is designed to be frozen, \modelname can train a denoising-aligned decoder directly on the denoised embeddings produced by the trained diffusion transformer, rather than on clean embeddings by the feature encoder. This is fundamentally different from VAE-based latent diffusion models~\citep{lin2025ctsyn}, in which the encoder and decoder are jointly trained to establish the latent space.

\vspace{-0.75mm}
Furthermore, we benchmark \modelname against 22 representative tabular generators across 45 real-world datasets. The results suggest that \modelname can capture and utilise fundamental representations for tabular data generation with reduced overfitting risk. 
Importantly, the three core components of \modelname make complementary contributions to preserving the inter-feature dependency structures of real data, rather than serving as interchangeable design choices.
% }

\begin{table}[!t]
\centering
\caption{\textbf{Model design comparison between \modelname and prior tabular foundation models.} We use ``$-$'' to indicate dimensions that are not applicable to particular models. \modelname is distinguished by identity-preserved feature encoding and denoising-aligned decoding without imposing feature-order bias, and offers strong practicability for tabular data generation.}
\label{tab:model_design_comparison}

\resizebox{\textwidth}{!}{
\begin{tabular}{l|ccc|ccc|cc}
\toprule

\multirow{3}{*}{\textbf{Foundation model}} 
& \multicolumn{3}{c|}{\textbf{Pretraining \& Fitting}} 
& \multicolumn{3}{c|}{\textbf{Generation}} 
& \multicolumn{2}{c}{\textbf{Practicability}} \\
& \makecell{Generative\\objective}
& Metadata-free                       
& \makecell{Feature identity\\preservation}        
& Paradigm         
& \makecell{Feature-order\\bias avoidance}                         
& \makecell{Denoising-aligned\\latent decoding}                       
& \makecell{No iterative\\in-context inference}   
& \makecell{Unlimited feature\\cardinality} \\

\midrule

GReaT~\cite{borisov2022language}                            & {\color[HTML]{3A7B21}   \CheckmarkBold} & {\color[HTML]{E6341C}   \XSolidBrush}   & {\color[HTML]{3A7B21}   \CheckmarkBold} & Autoregressive   & {\color[HTML]{E6341C}   \XSolidBrush}   & $-$                                     & $-$                                        & {\color[HTML]{3A7B21}   \CheckmarkBold} \\

TabPFN~\cite{grinsztajn2025tabpfn}                           & {\color[HTML]{E6341C}   \XSolidBrush} & {\color[HTML]{3A7B21}   \CheckmarkBold} & {\color[HTML]{3A7B21}   \CheckmarkBold} & Autoregressive   & {\color[HTML]{E6341C}   \XSolidBrush}   & $-$                                     & {\color[HTML]{E6341C}   \XSolidBrush}      & {\color[HTML]{E6341C}   \XSolidBrush}   \\

TabDPT~\cite{ma2025tabdpt}                           & {\color[HTML]{E6341C}   \XSolidBrush} & {\color[HTML]{3A7B21}   \CheckmarkBold} & {\color[HTML]{E6341C}   \XSolidBrush}   & Autoregressive   & {\color[HTML]{E6341C}   \XSolidBrush}   & $-$                                     & {\color[HTML]{E6341C}   \XSolidBrush}      & {\color[HTML]{E6341C}   \XSolidBrush}   \\
\bottomrule

\end{tabular}
}
\end{table}

% \vspace{-1mm}
\section{Related Work}
\vspace{-1mm}

\textbf{Tabular generative modelling.}
Tabular data generation has evolved from classic resampling towards expressive deep generative models~\cite{shi2025comprehensive, fang2024large, jiang2026tabstruct}. Conventional non-parametric methods~\cite{chawla2002smote, neto2025tabsds}, such as SMOTE~\cite{chawla2002smote}, synthesise samples by interpolation with observed data. These methods are efficient and easy to deploy, yet they can struggle to capture complex interactions across heterogeneous numerical and categorical features. Subsequent studies adapt deep generative models to tabular data~\cite{jiang2026tabstruct, qian2023synthcity, du2025systematic}, such as VAE-based methods~\cite{xu2019modeling, liu2023goggle} and diffusion models~\cite{kotelnikov2023tabddpm, zhang2023mixed, mueller2025continuous, shi2025tabdiff}. However, these methods are typically trained from scratch on each dataset, limiting their ability to utilise transferable knowledge across diverse tabular datasets.

Motivated by the success of foundation models in other modalities~\cite{achiam2023gpt, zheng2026diffusion, guo2025deepseek}, existing tabular foundation generators~\cite{borisov2022language, lin2025ctsyn, margeloiu2024tabebm, grinsztajn2025tabpfn, ma2025tabdpt, zhang2025limix, zhang2025mitra, qu2026tabiclv2} mainly follow three paths, as summarised in \Cref{tab:model_design_comparison}.
First, autoregressive methods serialise tabular data as a sequential modality~\cite{borisov2022language, seedat2024curated} and repurpose language models~\cite{fang2024large} or tabular foundation predictors~\cite{grinsztajn2025tabpfn, zhang2025limix} for generation. Although these methods can leverage pretrained representations, they impose a potentially misaligned feature order, which can degrade synthetic data quality~\cite{jiang2026tabstruct}.
Second, energy-based methods~\cite{margeloiu2024tabebm, ma2023tabpfgen}, such as TabEBM~\cite{margeloiu2024tabebm}, generate data from energy landscapes induced by tabular foundation predictors, achieving strong predictive utility while tending to emphasise local target-related structures over global inter-feature dependencies.
Third, CTSyn~\cite{lin2025ctsyn} explores latent diffusion with metadata-conditioned latent space and fixed-length tokenisation, which may obscure feature identity and induce train-inference mismatch in latent space.
We further provide a comprehensive review of prior work on tabular data generation and position \modelname as a tabular foundation model that supports cross-dataset pretraining and lightweight downstream adaptation in~\Cref{appendix:related_work}.
Overall, \modelname addresses these limitations through a metadata-free, structure-aware latent diffusion framework with per-feature embeddings and denoising-aligned decoding, while avoiding feature-order bias.

% \vspace{-1mm}
\section{\modelname: Tabular Foundation Model for Generative Modelling}
\label{sec:method}
\vspace{-1mm}

\Cref{fig:framework} illustrates the architecture of \modelname. We first describe the problem setup (\Cref{sec:problem}). We then introduce the core components of \modelname: a feature encoder that extracts general structural information (\Cref{sec:encoder}), a score-based diffusion model over latent embeddings (\Cref{sec:diffusion}), and a denoising-aligned decoder that maps latent embeddings back to the original feature space (\Cref{sec:decoder}). Next, we present the training and inference strategies (\Cref{sec:pretraining_and_inference}), corresponding pseudocode (\Cref{appendix:pseudocode}), and extended implementation details (\Cref{appendix:implementation_details}).

% \vspace{-1mm}
\subsection{\texorpdfstring{Problem Setup (\Cref{fig:framework}A)}{}}
\label{sec:problem}
% \vspace{-1mm}
% {
Let $\{(\rvx^{(i)}, y^{(i)})\}_{i=1}^{N} \sim p(\rvx, y)$ denote a tabular dataset, where each sample consists of $D$ mixed-type features (i.e., numerical and categorical) and one target. We use $\vx_d$ to denote the $d$-th feature (i.e., a column or variable), and $x^{(i)}_{d}$ to denote the value of the $d$-th feature in the $i$-th sample (i.e., a cell).
For notational clarity, we assume that all samples are used for training, and we treat the target as an additional feature, denoted by $\vx_{D+1} \coloneqq \{y^{(i)}\}_{i=1}^{N}$. Accordingly, we refer to the entire dataset by $\mX \in \mathbb{R}^{N \times (D+1)}$. We further use $\vx_\text{num}$ and $\vx_{\text{cat}}$ to denote numerical and categorical feature sets.
% }

\subsection{\texorpdfstring{Structure-aware Feature Encoder (\Cref{fig:framework}B)}{}}
\label{sec:encoder}
% \vspace{-1mm}
% {
\looseness-1
\modelname first maps a mixed-type tabular dataset from the original feature space $\mX \in \mathbb{R}^{N \times (D+1)}$ to structure-aware embeddings $\mH \in \mathbb{R}^{N \times (D+1) \times k}$ within a frozen latent space. Specifically, it tokenises ($\mathcal{S}_{\psi}$) each feature into a shared continuous latent space and then applies the early-to-middle layers ($\mathcal{E}_{\omega}$) of a pretrained PFN model to contextualise the tokens with inter-feature dependencies. This design is inspired by the observation that PFN models are intentionally exposed to abundant causal structures during their pretraining~\cite{grinsztajn2025tabpfn, hollmann2025accurate, zhang2025mitra, qu2026tabiclv2}.
Motivated by this property, \modelname can construct a transferable and structure-aware latent space across heterogeneous datasets for generative purposes.

\textbf{Tokeniser $\mathcal{S}_{\psi}$ handles heterogeneity without relying on metadata.}
% {
Since feature sets often vary across tabular datasets, learning a dataset-specific latent space would substantially limit cross-dataset transferability. Therefore, \modelname follows the TabPFN tokenisation strategy~\cite{hollmann2025accurate} and maps any table into a shared $k$-dimensional latent space while preserving the identity of each feature, that is, $\mX \in \mathbb{R}^{N \times (D+1)} \rightarrow \mT \in \mathbb{R}^{N \times (D+1) \times k}$. The per-feature tokenisation is computed directly from feature values (further details are in \Cref{appendix:implementation_details}) and does not rely on metadata such as column names or descriptions, thereby enabling flexible cross-dataset training and inference.
% }

\textbf{Encoder layers $\mathcal{E}_{\omega}$ leverage transferable pretrained representations.}
% {
\modelname employs the early-to-middle layers of a PFN predictor (i.e., the first $L_{\text{enc}}$ transformer layers), motivated by prior work~\cite{swelam2025does} suggesting that such representations implicitly capture inter-feature dependencies, thereby inducing the mapping $\mT \rightarrow \mH \in \mathbb{R}^{N \times (D+1) \times k}$.
Unlike in-context PFN predictors, \modelname does not require additional unseen query data to obtain latent embeddings for the available training data. Concretely, \modelname derives latent embeddings $\mH$ by applying the encoder layers with a leave-one-fold-out feature extraction strategy~\cite{ye2025closer}, which provides important insights into the effectiveness of randomised feature tokenisation, showing how TabPFN can infer meaningful attribute relationships across heterogeneous datasets without relying on dataset-specific feature embeddings (further details are in \Cref{appendix:implementation_details}).
Thus, the encoder serves purely as a representation extractor for the training data, rather than as a predictor that depends on access to query samples at inference time. 

A key design choice in \modelname is to preserve the identity of each feature embedding, operating on $\mH^{(i)} \in \mathbb{R}^{(D+1) \times k}$ rather than a flattened vector~\cite{zhang2023mixed, shi2025tabdiff} or a predetermined number of latent query tokens~\cite{lin2025ctsyn}. With one-to-one correspondence between each feature and its embedding, \modelname mitigates the disruption of per-feature semantics and the information bottleneck stemming from a fixed-length latent token sequence.
% }

\textbf{Frozen latent space mitigates distribution shifts.}
% {
In \modelname, we construct a frozen latent space by inheriting the tokeniser and the first $L_\text{enc}$ transformer layers from a pretrained PFN model. By default, we adopt \mbox{Real-TabPFN-2.5}~\citep{grinsztajn2025tabpfn}.
Specifically, we freeze the latent space for three main reasons:~(1)~A~frozen latent space uniquely allows \modelname to decouple latent modelling from decoding, thereby mitigating the train-inference mismatch common in latent diffusion models (\Cref{sec:decoder}). 
(2)~The early-to-middle layers of a pretrained PFN model can provide strong inductive bias for capturing general structural information~\citep{swelam2025does}.
(3)~A~frozen encoder stabilises optimisation and mitigates latent collapse by preventing the latent space from drifting across datasets~\citep{erdogan2025layerlock}.
% }

% \vspace{-2mm}
\subsection{\texorpdfstring{Score-based Diffusion Transformer (\Cref{fig:framework}C)}{}}
\label{sec:diffusion}
% \vspace{-1mm}
% {
\modelname models latent distribution with a score-based denoiser $\mathcal{G}_{\theta}$ that learns to recover clean latent embeddings $\widehat{\mH} \in \mathbb{R}^{N \times (D+1) \times k}$ from their noisy counterparts. Specifically, the diffusion transformer maintains the one-to-one correspondence between each feature and its embedding, enabling \modelname to refine inter-feature interactions. Furthermore, as the transformer architecture is agnostic to the length of the token sequence (i.e., number of features), \modelname supports large-scale pretraining across heterogeneous datasets with varying feature sets.
% }

\textbf{Forward process preserves token structure.}
% {
The forward process constructs noisy latent embeddings by perturbing clean latent embeddings while preserving token structure. Specifically, given the clean latent embeddings $\mH$, we first obtain the normalised latent embeddings \mbox{$\mZ_0 \in \mathbb{R}^{N \times (D+1) \times k}$} via Z-score normalisation. We then generate noisy latent embeddings according to
\begin{equation}
    \mZ_\sigma = \mZ_0 + \sigma \mE, 
    \quad
    \sigma = \exp(p_{\text{mean}} + p_{\text{std}} \, \epsilon_\sigma), 
    \quad 
    \epsilon_\sigma \sim \mathcal{N}(0, 1),
\end{equation}
where $\mE \sim \mathcal{N}(\mathbf{0}, \mI)$ denotes isotropic Gaussian noise, and $(p_{\text{mean}}, p_{\text{std}})$ parametrise the log-normal distribution from which the noise level $\sigma$ is sampled.
% }

\textbf{Reverse process supports varying feature sets.}
% {
Given noisy embeddings $\mZ_\sigma$, the score-based denoiser estimates the clean latent embedding via the EDM~\cite{karras2022elucidating} parametrisation: 
\mbox{
$
\mathcal{G}_\theta(\mZ_\sigma,\sigma) = c_{\text{skip}}(\sigma)\mZ_\sigma + c_{\text{out}}(\sigma)\, \mathcal{F}_\theta\left(c_{\text{in}}\left(\sigma\right)\mZ_\sigma,\; c_{\text{noise}}\left(\sigma\right)\right)
$
}, where $\mathcal{F}_\theta$ is a transformer model and $\sigma_{\text{data}}$ denotes the standard deviation of the clean latent embeddings. The corresponding preconditioning coefficients are
$
c_{\text{skip}}(\sigma)=\frac{\sigma_{\text{data}}^2}{\sigma^2+\sigma_{\text{data}}^2}
$,
$
c_{\text{out}}(\sigma)=\frac{\sigma \sigma_{\text{data}}}{\sqrt{\sigma^2+\sigma_{\text{data}}^2}}
$,
$
c_{\text{in}}(\sigma)=\frac{1}{\sqrt{\sigma^2+\sigma_{\text{data}}^2}}
$,
$
c_{\text{noise}}(\sigma)=\frac{\ln \sigma}{4}
$.
Under this parametrisation, the denoiser induces an approximate score field $\vs_\theta(\mZ_\sigma,\sigma)$ and the associated reverse-time dynamics:
\begin{equation}
    \vs_\theta(\mZ_\sigma,\sigma)
    =
    \nabla_{\mZ_\sigma}\log p_\sigma(\mZ_\sigma),
    \quad
    \frac{\text{d}\mZ_\sigma}{\text{d}\sigma}
    =
    -\sigma\vs_\theta(\mZ_\sigma,\sigma)
    \propto
    \frac{\mZ_\sigma-\mathcal{G}_\theta(\mZ_\sigma,\sigma)}{\sigma}.
\end{equation} 
As $\sigma$ decreases, the latent sample is progressively guided from a noisy state towards the clean latent manifold. Once the diffusion model has been trained, we apply the inverse Z-score normalisation to $\mathcal{G}_\theta(\mZ_\sigma, \sigma)$ to recover the latent embeddings $\widehat{\mH}$ on the same scale as the clean embeddings.
% }

\textbf{Objective of latent modelling.}
% {
The score-based diffusion transformer is optimised using the EDM weighted denoising objective~\cite{karras2022elucidating}, which trains the denoiser to recover the clean normalised latent embeddings from their noisy counterparts. The diffusion loss is computed via
\begin{equation}
    \mathcal{L}_{\text{diff}}
    =
    \frac{1}{N}
    \sum_{i=1}^{N}
    \frac{\sigma_i^2 + \sigma_{\text{data}}^2}{(\sigma_i \sigma_{\text{data}})^2}
    \left\|
    \mathcal{G}_\theta(\mZ_{\sigma_i}^{(i)}, \sigma_i) - \mZ_0^{(i)}
    \right\|_2^2
    \label{eq:objective_diff}
\end{equation}
which reweights training samples according to their noise levels, yielding a balanced learning signal across the full range of noise scales considered during diffusion training.
% }

\subsection{\texorpdfstring{Denoising-aligned Latent Decoder (\Cref{fig:framework}D)}{}}
\label{sec:decoder}
% {
The denoising-aligned latent decoder first refines the denoised latent embeddings $\widehat{\mH}$ with a decoder transformer ($\mathcal{D}_{\phi}$) and then projects individual tokens back to the original feature space $\widehat{\mX} \in \mathbb{R}^{N \times (D+1) }$ with a lightweight detokeniser ($\mathcal{R}_{\eta}$).
By training the decoder directly on denoised embeddings produced by the fitted diffusion transformer, \modelname aligns decoder training with the latent distribution at inference time, thereby mitigating the train-inference mismatch that commonly affects latent diffusion models.
% }

\textbf{Decoder transformer $\mathcal{D}_{\phi}$ refines embeddings for generation.}
% {
Although the denoised latent embeddings encode contextualised information, they can still be too coarse to directly generate accurate values for individual features (see \Cref{sec:ablation_impacts}). Therefore, \modelname further refines them with a transformer $\mathcal{D}_{\phi}$ of $L_{\text{dec}}$ layers to obtain generation-ready token representations \mbox{$\mU = [\vu^{(1)}, \dots, \vu^{(N)}]^{\mathsf{T}} \in \mathbb{R}^{N \times (D+1) \times k}$}. As a result, $\mathcal{D}_{\phi}$ remains agnostic to the number of features and can thus be pretrained across heterogeneous datasets with varying feature sets.
% }

\textbf{Detokeniser $\mathcal{R}_{\eta}$ enables flexible generation.}
% {
\modelname employs a lightweight detokeniser to map the refined embeddings $\mU$ back to the original feature space. For the $j$-th feature of the $i$-th sample, the detokeniser produces either a reconstructed numerical value or a categorical logit vector:
\begin{equation}
    \widehat{o}_{j}^{(i)}
    =
    \begin{cases}
        \langle \vu_{j}^{(i)}, \vw_{j} \rangle,
        & \text{if } \vx_j \in \vx_{\text{num}}, \\[4pt]
        \mW_{j}\vu_{j}^{(i)} + \vb_{j},
        & \text{if } \vx_j,
    \end{cases}
    \label{eq:detoken}
\end{equation}
where $\langle \cdot, \cdot \rangle$ denotes the dot product, $\vw_j \in \mathbb{R}^{k}$ is a learnable reconstruction vector for the $j$-th numerical feature, and $\mW_j \in \mathbb{R}^{C_j \times k}$ and $\vb_j \in \mathbb{R}^{C_j}$ are the parameters of the linear head for the $j$-th categorical feature, whose cardinality is $C_j$. The decoding process is further detailed in ~\Cref{appendix:feature_decoding}.

\looseness-1
Unlike the fixed output heads in PFN models~\cite{grinsztajn2025tabpfn, qu2026tabiclv2, zhang2025limix}, which are often limited to feature cardinalities no greater than 10, \modelname is designed to use a lightweight detokeniser for each dataset, while keeping the decoder transformer shared across datasets. As a result, \modelname can naturally generate high-cardinality categorical features without more complex techniques such as hierarchical classification~\cite{qu2026tabiclv2, qu2025tabicl}, which can add non-trivial computational overhead. This dataset-specific detokeniser therefore improves flexibility in synthetic data generation at only minimal computational cost (see \Cref{sec:practicability}).
% }

\textbf{Objective of denoising-aligned decoding.}
% {
The denoising-aligned latent decoder is optimised using a mixed-type reconstruction objective defined over the decoder outputs:
\begin{equation}
    \mathcal{L}_{\text{recon}}
    =
    \frac{1}{N}\sum_{i=1}^{N}
    \left(
    \left\|
        \widehat{\vx}_{\text{num}}^{(i)} - \vx_{\text{num}}^{(i)}
    \right\|_2^2
    +
    \sum_{j=1}^{|\vx_{\text{cat}}|}
    \operatorname{CE}\!\left(
        \widehat{\vp}_{\text{cat},j}^{(i)},
        x_{\text{cat},j}^{(i)}
    \right)
    \right)
    \label{eq:objective_recon}
\end{equation}
where $\operatorname{CE}(\cdot,\cdot)$ denotes cross-entropy, and $\widehat{\vp}_{\text{cat},j}^{(i)}=\operatorname{softmax}\left( \widehat{\vo}_{\text{cat},j}^{(i)}\right) \in \mathbb{R}^{C_j}$, applied independently to each categorical feature.
A core characteristic of \modelname is that $\mathcal{D}_{\phi}$ are trained solely on denoised embeddings $\widehat{\mH}$ by the fitted diffusion transformer $\mathcal{G}_{\theta}$. As such, $\mathcal{D}_{\phi}$ is directly trained on the latent distribution consistent with what it will encounter at inference time.
% }

\subsection{Pretraining and Inference}
\label{sec:pretraining_and_inference}

\textbf{Pretraining setup.}
% {
We pretrain \modelname on the same 43 real-world datasets (see \Cref{appendix:pretraining_dataset}) as used by \mbox{Real-TabPFN-2.5}~\cite{grinsztajn2025tabpfn}, avoiding additional real-world datasets to reduce the risk of data leakage when evaluating on unseen datasets.
For each dataset, we compute frozen latent embeddings for all samples only once using the structure-aware feature encoder and cache the embeddings. Then we optimise \modelname in two sequential stages:~(1)~we train the score-based diffusion transformer for 5{,}000 steps; (2)~we train the denoising-aligned decoder for another 5{,}000 steps. One full pass over all 43 datasets constitutes a pretraining round, and we run 10 rounds in total.
We provide detailed model implementation details in \Cref{appendix:implementation_details}.
% }

\textbf{Fitting to unseen datasets.}
% {
Given an unseen dataset, we first compute frozen latent embeddings for the available training data only once using the structure-aware feature encoder, then load the pretrained diffusion transformer $\mathcal{G}_{\theta}$ and decoder transformer $\mathcal{D}_{\phi}$, and initialise a dataset-specific detokeniser $\mathcal{R}_{\eta}$.
We first finetune the diffusion transformer for 100 steps. We then finetune the decoder for 100 steps, jointly updating the transferable decoder transformer and the newly instantiated detokeniser. We provide detailed optimisation configurations in \Cref{appendix:model_configurations}.
% }

\textbf{Synthetic data generation.}
% {
After fitting, \modelname generates synthetic mixed-type tabular data by denoising latent embeddings with random noise and therefore no longer requires the structure-aware feature encoder. This yields a key practical advantage over other in-context tabular foundation generators: as the encoder performs in-context inference only once to obtain the embeddings during fitting, \modelname avoids the substantial cost of repeated in-context inference when generating data.
% }

\section{Experiments}
\label{sec:experiment}

We evaluate \modelname by answering three research questions:

\begin{itemize}[topsep=0pt, leftmargin=10pt, itemsep=0pt]
    \item \textbf{Generation Quality (Q1, \Cref{sec:generation_quality}, \Cref{appendix:extended_analysis_quality})}:
    How effectively does \modelname generate high-quality synthetic tabular data while mitigating the risk of overfitting?

    \item \textbf{Ablation Impacts (Q2, \Cref{sec:ablation_impacts}, \Cref{appendix:extended_analysis_ablation})}:
    How do individual components of \modelname make complementary contributions to the overall model performance?

    \item \textbf{Practicability (Q3, \Cref{sec:practicability}, \Cref{appendix:extended_analysis_practicability})}:
    How readily can \modelname be applied to unseen real-world tasks at practical computational cost?
\end{itemize}

\textbf{Real-world benchmark datasets.}
We curate 45 challenging datasets from the TabArena~\cite{erickson2025tabarena} and TabStruct~\cite{jiang2026tabstruct} benchmark suites, comprising 31 classification datasets with 748-150,000 samples and 5-119 features, and 14 regression datasets with 907-53,940 samples and 6-82 features.
Specifically, we exclude the 43 pretraining datasets to assess \modelname on unseen datasets and manually verify that all selected datasets include metadata suitable for fitting benchmark generators (i.e., CTSyn~\cite{lin2025ctsyn} and GReaT~\cite{borisov2022language}). Full dataset descriptions are provided in \Cref{appendix:benchmark_dataset}.

\textbf{Data splitting.}
For each dataset of $N$ samples, we perform nested cross-validation with repeated shuffle (details are provided in \Cref{appendix:data_processing}). In each repetition, we split the dataset into four disjoint subsets: 30\% train set, 30\% test set, 30\% holdout set, and 10\% validation set. We shuffle the dataset to repeat the splitting 10 times, summing up to 10 runs per dataset.
All benchmark generators are fitted on the training split, denoted by $\mX_{\text{Train}}$, and each generator produces a synthetic dataset with $N_{\text{Train}}$ samples. In contrast to prior studies~\cite{jiang2026tabstruct, margeloiu2024tabebm, zhang2023mixed, shi2025tabdiff}, we further preserve a holdout set, denoted by $\mX_{\text{Holdout}}$, which remains unused during training, validation, and hyperparameter tuning.

\textbf{Benchmark setups.}
We compare \modelname against 22 benchmark methods, including 8 foundation models and 14 dataset-specific methods.
By default, ``TabPFN'' refers to \mbox{Real-TabPFN-2.5}~\cite{grinsztajn2025tabpfn}, and ``TabICL'' refers to TabICLv2~\cite{qu2026tabiclv2}.
We further report the performance of $\mX_{\text{Train}}$ and $\mX_{\text{Holdout}}$ as reference points for assessing overfitting risk. In particular, for metrics where $\mX_{\text{Train}}$ represents a lower/upper-bound reference, synthetic data that are closer to $\mX_{\text{Train}}$ than $\mX_{\text{Holdout}}$ may indicate memorisation of the $\mX_{\text{Train}}$, rather than generalisation. 
We provide further details on implementation (\Cref{appendix:benchmark_generator}) and experimental setup (\Cref{appendix:exp_setup}).

\begin{figure}[!t]
    \centering
    \includegraphics[width=\linewidth]{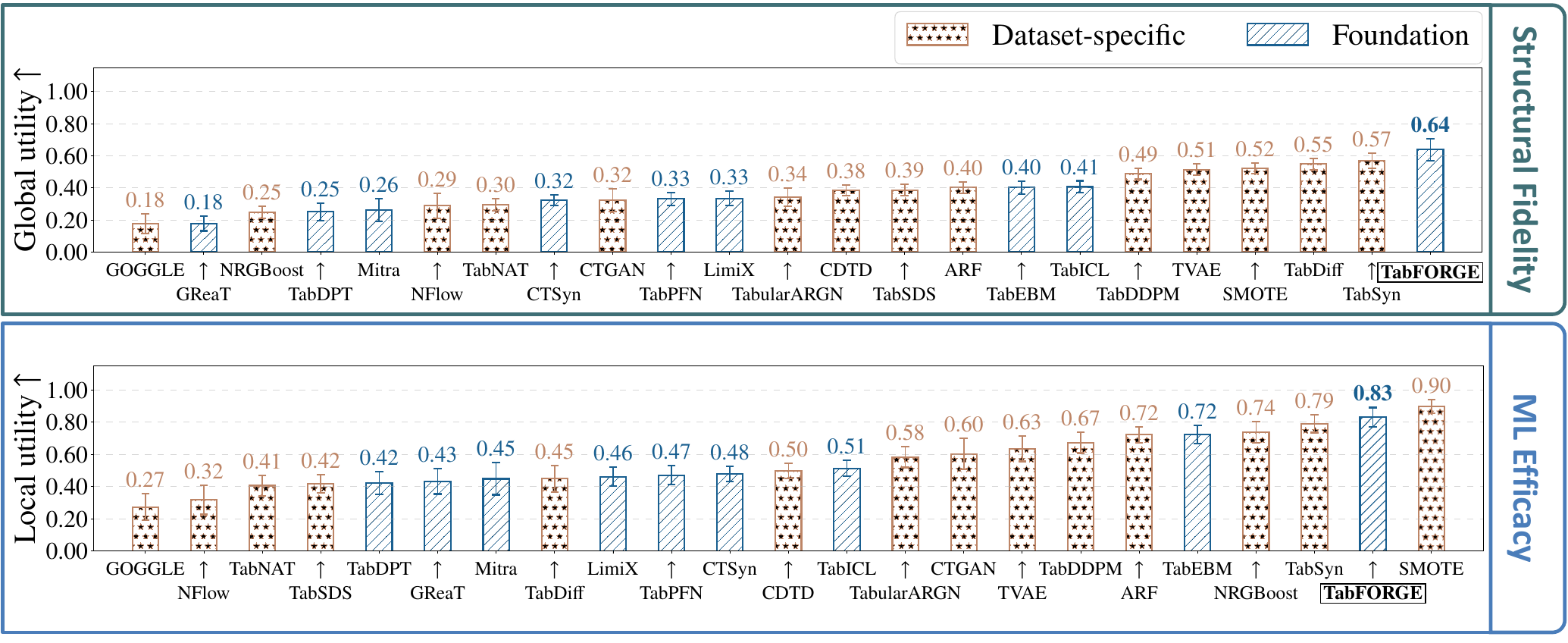}
    % \vspace{-5mm}
    \caption{\textbf{Benchmark results of 23 generators on 45 real-world tabular datasets.} We report the normalised mean $\pm$ std metric values across datasets. \textbf{Top:} Global utility~\citep{jiang2026tabstruct} -- a higher value typically indicates that the generator better captures global dependency structures across all features (structural fidelity). \textbf{Bottom:} Local utility~\citep{xu2019modeling, zhang2023mixed, jiang2026tabstruct} -- a higher value typically indicates that synthetic data yields stronger predictive performance when used to train downstream predictors (ML efficacy). \modelname achieves the best overall performance, particularly surpassing benchmark methods in structural fidelity.\looseness-1}
    \label{fig:utility}
% \vspace{-7mm}
\end{figure}

\textbf{Evaluation scope.}
\modelname is a tabular foundation generator that leverages a causally-informed structural inductive biases, rather than an approach for explicit causal modelling. Therefore, we follow established evaluation protocols on tabular data generation~\cite{jiang2026tabstruct, zhang2023mixed, qian2023synthcity}. Specifically, we adopt eight well-defined metrics from prior work spanning four complementary evaluation dimensions: structural fidelity, ML efficacy, density estimation, and privacy preservation.
For structural fidelity, ground-truth causal structures are rarely available for real-world tabular datasets, even precluding the possibility of theoretical validation. Note that, direct causal verification, such as comparing explicitly discovered causal graphs from real and synthetic data, may yield misleading conclusions because existing causal discovery methods cannot reliably infer the governing causal structures of real-world tabular datasets~\cite{kaddour2022causal, tu2024causality, glymour2019review, nastl2024causal, jiang2026tabstruct}.
Therefore, we mainly evaluate structural fidelity with the empirically effective metric (i.e., global utility~\cite{jiang2026tabstruct}), rather than relying on direct causal verification. Accordingly, our evaluation does not aim to establish that \modelname learns explicit causal structures, such as identifiable DAGs, valid interventional distributions, or counterfactually correct mechanisms.
We provide full details on the evaluation metrics in~\Cref{appendix:related_work} and discuss future directions for evaluating tabular generators in~\Cref{appendix:future_work}.

% \vspace{-1mm}
\subsection{Generation Quality (Q1)}
\label{sec:generation_quality}
% \vspace{-1mm}

\begin{figure}[!t]
    \centering
    \includegraphics[width=\linewidth]{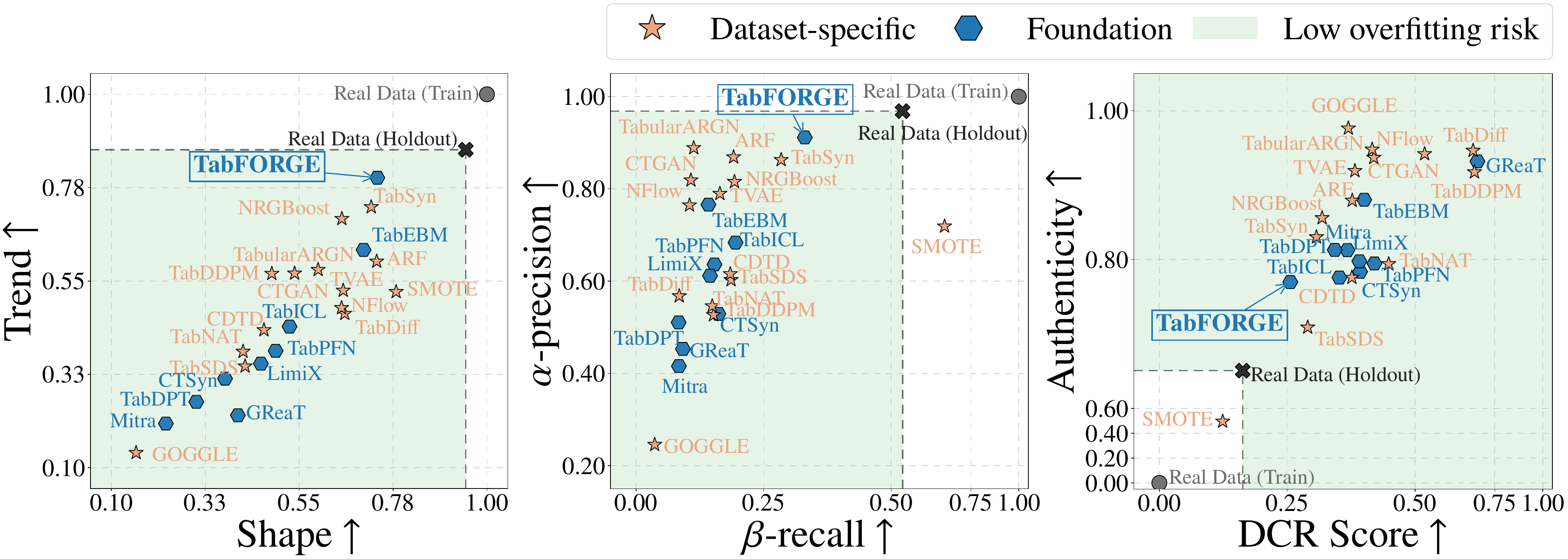}
    % \vspace{-5mm}
    \caption{\textbf{Comparison of the fitting behaviour of 23 tabular data generators across 45 real-world datasets.} We report the normalised mean metric values across datasets, with the axis display scales adjusted for visual clarity. \textbf{Left:} Low-order density estimation, which assesses the preservation of marginal distributions (\textit{Shape}~\cite{wust2011sdmetrics}) and inter-feature correlations (\textit{Trend}~\cite{wust2011sdmetrics}). \textbf{Middle:} High-order density estimation, which quantifies sample-level similarity (\mbox{\textit{$\alpha$-precision}}~\cite{alaa2022faithful}) and distributional coverage (\mbox{\textit{$\beta$-recall}}~\cite{alaa2022faithful}).  \textbf{Right:} Privacy preservation, which assesses whether synthetic data avoids memorising (\textit{Authenticity}~\cite{alaa2022faithful}) or closely duplicating training records (\textit{DCR Score}~\cite{wust2011sdmetrics}). Privacy metrics should be interpreted together with the other dimensions, since overly high values may indicate a poor fit to real data. Detailed explanations for considered evaluation metrics are in Appendix~\ref{appendix:tabular_data_evaluation}. Synthetic data that scores closer to $\mX_{\text{Train}}$ than $\mX_{\text{Holdout}}$ indicates greater similarity to the training data than holdout data, suggesting a higher risk of overfitting. Accordingly, the shaded region denotes a low risk of overfitting. \modelname is a strong tabular generator that achieves high performance while reducing the risk of overfitting.}
    \label{fig:overfitting}
\vspace{-6mm}
\end{figure}

\textbf{Existing tabular foundation generators still struggle to match strong dataset-specific generators.}
In~\Cref{fig:utility}, dataset-specific generators account for seven of the top-10 methods in global utility and eight of the top-10 methods in local utility. This dominance suggests that existing foundation generators have not yet fully realised the benefits of transferable tabular representations for generation. It further showcases the key limitations of mainstream tabular foundation generators:
{(i)}~the weak overall performance of LLM-based generators (e.g., GReaT) and tabular foundation predictors (e.g., TabICL) indicates that treating tabular data as a sequential modality can be suboptimal for generation;
{(ii)}~the high local utility but low global utility of TabEBM suggests that it overemphasises the prediction target, rather than modelling the global dependency structures across all features;
{(iii)}~the limited effectiveness of CTSyn further highlights the importance of integrating structure-aware representations for tabular generative modelling. Moreover, its instability further shows how to integrate metadata into tabular data generation remains an open research question.

\textbf{\modelname achieves strong global structural fidelity.}
\Cref{fig:utility} (Top) shows that \modelname substantially outperforms the previous best tabular foundation model, TabICL, improving global utility by a clear margin (0.41$\rightarrow$0.64). Notably, \modelname is the only foundation model that surpasses the strongest dataset-specific method, TabSyn, improving from 0.57 to 0.64 in global utility. 
Further results and analyses on datasets with known causal structures, high-dimensional datasets, and generators with explicit causal modelling are provided in~\Cref{appendix:extended_analysis_quality}.
These results indicate that \modelname better preserves global inter-feature structural relationships than both dataset-specific generators and existing foundation models, showing that its design provides an effective inductive bias for modelling global dependency structures in tabular data.

\textbf{\modelname achieves strong ML efficacy.}
As shown in \Cref{fig:utility} (Bottom), in terms of local utility, \modelname outperforms the previous best tabular foundation model, TabEBM, by a clear margin (0.72$\rightarrow$0.83), and ranks second overall, closely approaching SMOTE. We further note that \Cref{fig:overfitting} shows SMOTE can be susceptible to overfitting. These results suggest that the synthetic data generated by \modelname remains highly effective for downstream predictive modelling. More importantly, in contrast to previous foundation methods that primarily optimise predictive power, such as TabEBM, \modelname also achieves the highest global utility. As noted in prior work~\cite{jiang2026tabstruct}, balancing global and local utility reflects the ability to capture both global and local inter-feature dependency structures. Therefore, \modelname achieves strong ML efficacy while maintaining high structural fidelity to the underlying structures of real tabular data.

\textbf{\modelname reduces overfitting risk while preserving synthetic data quality.}
The fitting-behaviour analysis in \Cref{fig:overfitting} shows that \modelname achieves strong performance while remaining in the low-overfitting-risk region. Specifically, \modelname lies within the shaded region across all six metrics, covering low-order density estimation, high-order density estimation, and privacy preservation.
In contrast, several high-performing benchmark methods exhibit signs of overfitting. For instance, SMOTE achieves the highest local utility (\Cref{fig:utility}), and outperforms the real holdout data across multiple metrics (\Cref{fig:overfitting}), including $\beta$-recall, authenticity, and DCR score. This pattern suggests that SMOTE's strong performance is likely driven by memorisation of the training data, whereas \modelname generalises beyond the training data rather than duplicating it.

\begin{figure}[!t]
%    \vspace{-5mm}
    \centering
    \includegraphics[width=\linewidth]{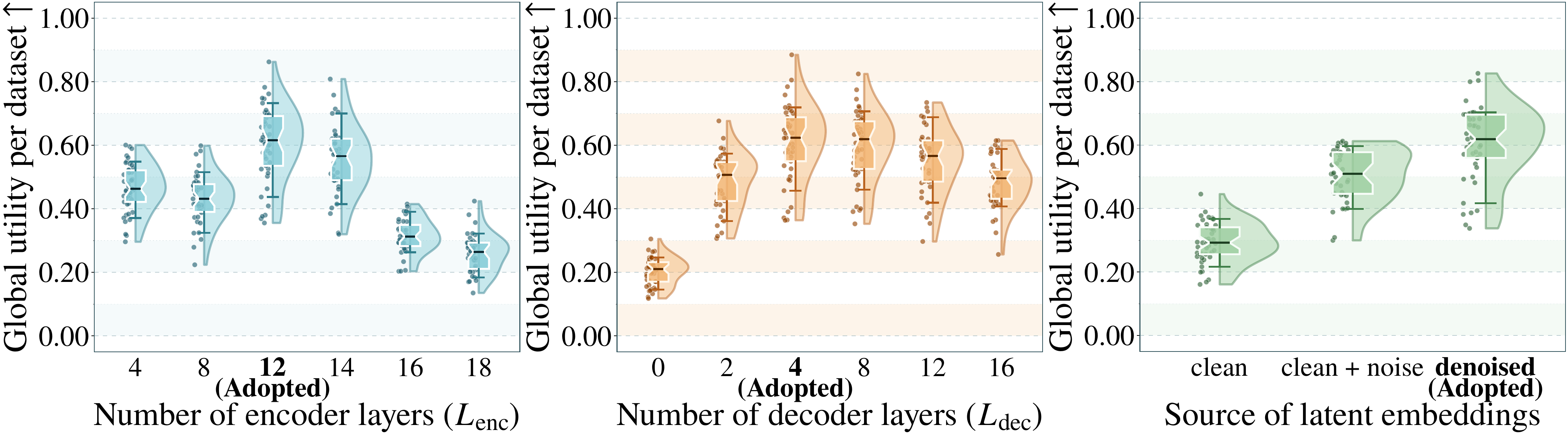}
    \caption{\textbf{Ablation studies of \modelname on 45 real-world datasets.} We report the global utility on each dataset and highlight the adopted configurations in \modelname. More analyses on the role of each component are provided in~\Cref{appendix:extended_analysis_ablation}. \textbf{Left:} Effect of $L_{\text{enc}}$, with the best performance at 12 layers, whereas deeper encoders perform worse because they can provide weaker signals of global dependency structures. \textbf{Middle:} Effect of $L_{\text{dec}}$, where 4 layers yield the highest global utility, indicating that the decoder transformer is necessary to refine the denoised embeddings for generation. \textbf{Right:} Sources of latent embeddings, showing that denoised embeddings outperform clean embeddings, even when the latter are augmented with noise.}
    \label{fig:ablation_arch}
\vspace{-2mm}
\end{figure}

\subsection{Ablation Studies (Q2)}
\label{sec:ablation_impacts}

\looseness-1
\textbf{Early-to-middle encoder layers provide useful global structural information.}
As shown in \Cref{fig:ablation_arch} (Left), using $L_{\text{enc}}=12$ encoder layers yields the highest global utility, whereas both shallower and deeper encoders degrade performance, with deeper encoders performing even worse than shallower ones. This suggests that early-to-middle PFN layers provide the most useful latent representations for tabular data generation. 
Shallow layers may lack the capacity to capture rich inter-feature interactions, and deeper layers, while more powerful in predictive modelling, become increasingly biased towards the prediction target, thereby reducing global structural information. This interpretation is consistent with prior work showing that representations from both shallower and deeper layers yield less accurate causal discovery than those from intermediate layers~\cite{swelam2025does}.
Further results of the encoder architecture and learned representations are provided in~\Cref{appendix:extended_analysis_ablation}.
Thus, rather than merely reproducing prior observations about where structural information emerges within PFNs~\cite{swelam2025does}, our ablation provides complementary evidence that the intermediate representations are also more effective for generation. These findings further support our hypothesis that preserving dependencies across features can yield broadly useful pretrained representations that extend beyond predictive modelling.

\textbf{Decoder refinement is necessary for generation-ready representations.}
\Cref{fig:ablation_arch} (Middle) shows that using no decoder layer leads to the weakest performance, whereas $L_{\text{dec}}=4$ achieves the highest global utility. These results demonstrate that denoised latent embeddings cannot be directly detokenised into high-quality tabular data without further refinement ($L_{\text{dec}}=0$).
The results further suggest that the structural prior induced by PFN pretraining alone is insufficient to account for the performance gains of \modelname. Instead, our experiments provide complementary evidence that structurally informative pretrained representations serve as an effective basis for generation, while a transferable decoder transformer remains necessary to refine them into generation-ready representations that are better aligned with the value space of each feature.

\textbf{Denoising-aligned decoding mitigates training-inference mismatch.}
As shown in \Cref{fig:ablation_arch} (Right), training the decoder with denoised embeddings achieves the highest global utility, outperforming both clean embeddings and clean embeddings augmented with noise. Specifically, following prior work~\cite{zheng2026diffusion}, we augment clean embeddings with additive Gaussian noise. The results indicate that generic noise augmentation only partially addresses the mismatch between the training and inference latent distributions. In contrast, directly training the decoder on denoised embeddings exposes it to representations that are more consistent with those encountered during inference. This finding further supports a key principle of \modelname: denoising-aligned decoding mitigates the train-inference distribution shift that commonly arises in latent diffusion models.

\textbf{The three core components of \modelname are complementary, rather than interchangeable.}
First, the encoder and diffusion-decoder framework play distinct but complementary roles: the encoder produces structure-aware representations, while the diffusion-decoder framework transforms these representations into high-quality synthetic tabular data. Since removing any of these components would disrupt the generation pipeline, their contributions are not naturally interpretable through percentage-style attribution.
Second, the frozen encoder provides a transferable latent space that preserves feature identity. Specifically, it maps heterogeneous tabular data into a unified representation space while retaining feature-level semantics and structural information.
Third, the diffusion-decoder framework further makes this latent space generative. The diffusion model learns the joint distribution of the latent embeddings, while the decoder maps sampled latent representations back to the original feature space. Removing the decoder ($L_{\text{dec}}=0$) leads to the weakest performance (\Cref{fig:ablation_arch}, Middle), highlighting its importance for effective reconstruction. In addition, the denoising-alignment design helps reduce the train-inference mismatch, thereby improving the robustness of the overall diffusion-decoder framework (\Cref{fig:ablation_arch}, Right).

\subsection{Practicability (Q3)}
\label{sec:practicability}

\begin{figure}[!ht]
    \centering
%    \vspace{-3mm}
    \includegraphics[width=\linewidth]{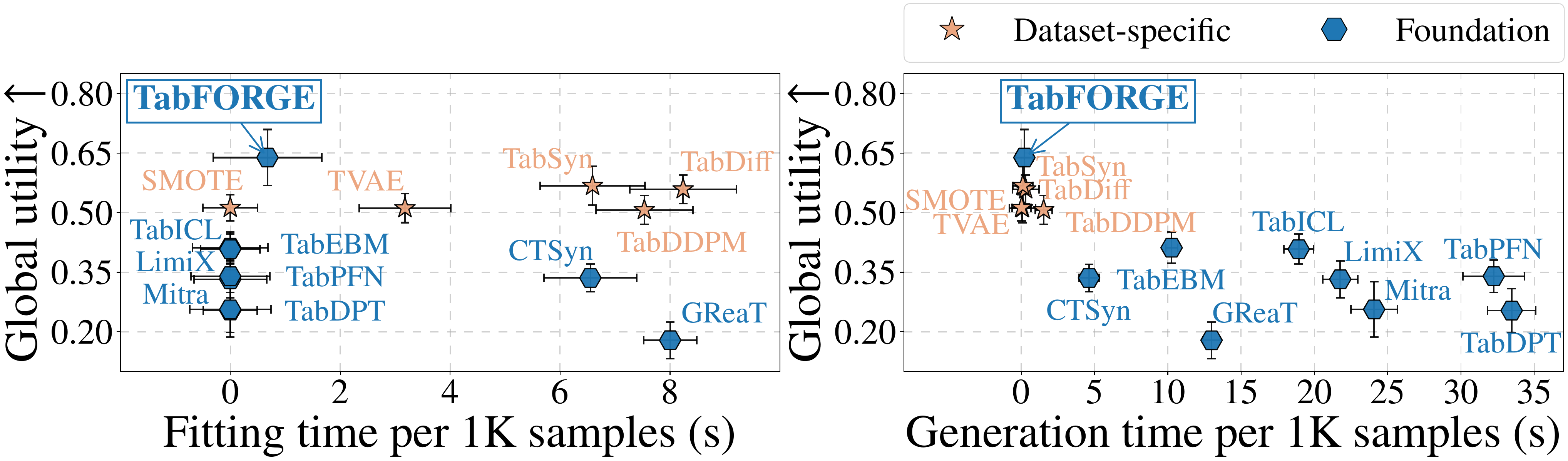}
    \caption{\textbf{Computation efficiency on 45 real-world datasets.} For visual clarity, we report all nine foundation models and Top-5 dataset-specific methods in global utility. \textbf{Left:} Median fitting time per 1,000 samples vs. mean normalised global utility. \textbf{Right:} Median generation time per 1,000 samples vs.\ mean normalised global utility. \modelname generally achieves better structural fidelity with higher computation efficiency than benchmark methods.}
    \label{fig:computation_time}
\end{figure}

\textbf{\modelname incurs low fitting and generation costs on unseen datasets.}
As shown in \Cref{fig:computation_time}, \modelname is substantially more efficient than most benchmark methods in both fitting and generation. On unseen datasets, its fitting time is only around 9.71\% of that required to train TabSyn from scratch, as most learning capacity is acquired during pretraining. Moreover, \modelname requires only around 1.66\% of the generation time of the fastest in-context foundation generator, TabEBM, because it avoids iterative in-context inference. Further discussion is provided in~\Cref{appendix:extended_analysis_practicability}.

\looseness-1
\textbf{Limitations and future work.}
Although \modelname performs strongly across benchmark datasets, its current scale can be constrained by available computational resources, as we have not observed clear signs of performance saturation during pretraining.
Our evaluation investigates whether meaningful cross-dataset transfer can already emerge at this moderate pretraining scale.
These observations motivate us to further scale the pretraining data collection, increase model capacity, and develop more computationally efficient implementations. A detailed discussion on future explorations is provided in Appendix~\ref{appendix:future_work}.

\section{Conclusion}
We introduce \modelname, a tabular foundation model for generative modelling. \modelname builds on pretrained tabular representations and introduces a structure-aware latent diffusion framework tailored to heterogeneous tabular data.
Through our comprehensive evaluation against 22 benchmark generators across 45 real-world datasets, \modelname generates high-quality synthetic tabular data with strong structural fidelity. In particular, \modelname substantially improves global utility over existing tabular foundation generators and is the only foundation model in our study to surpass the strongest dataset-specific generator in global utility.
Our findings suggest that pretrained structure-aware representations and denoising-aligned decoding form an effective basis for preserving complex inter-feature dependencies in tabular generative modelling.
We will open-source \modelname to facilitate community collaboration and further advance high-fidelity tabular foundation generators.

\begin{ack}
The authors would like to express their gratitude to Prof.~Han-Jia Ye and Dr.~Johannes Hoffart for their insightful discussions on tabular foundation models during the early stages of this project.
XJ acknowledges the generous support of the Google PhD Fellowship.
MJ and NS acknowledge the support of the U.S. Army Medical Research and Development Command of the Department of Defense; through the FY22 Breast Cancer Research Program of the Congressionally Directed Medical Research Programs, Clinical Research Extension Award GRANT13769713. Opinions, interpretations, conclusions, and recommendations are those of the authors and are not necessarily endorsed by the Department of Defense.
\end{ack}

%%%%%%%%%%%%%%%%%%%%%%%%%%%%%%%%
% References
%%%%%%%%%%%%%%%%%%%%%%%%%%%%%%%%
\bibliography{reference}
\bibliographystyle{plain}

%%%%%%%%%%%%%%%%%%%%%%%%%%%%%%%%
% Appendix
%%%%%%%%%%%%%%%%%%%%%%%%%%%%%%%%
\input{camera_ready_2_appendix}
\input{camera_ready_3_checklist}

\end{document}

%% file: math_commands.tex
\usepackage{amsmath,amsfonts,bm}

\def\eqref#1{equation~\ref{#1}}
\def\1{\bm{1}}

\def\rvx{{\mathbf{x}}}

\def\vb{{\bm{b}}}

\def\vo{{\bm{o}}}
\def\vp{{\bm{p}}}

\def\vr{{\bm{r}}}
\def\vs{{\bm{s}}}
\def\vt{{\bm{t}}}
\def\vu{{\bm{u}}}

\def\vw{{\bm{w}}}
\def\vx{{\bm{x}}}

\def\mE{{\bm{E}}}

\def\mH{{\bm{H}}}
\def\mI{{\bm{I}}}

\def\mP{{\bm{P}}}

\def\mR{{\bm{R}}}

\def\mT{{\bm{T}}}
\def\mU{{\bm{U}}}

\def\mW{{\bm{W}}}
\def\mX{{\bm{X}}}

\def\mZ{{\bm{Z}}}

\DeclareMathAlphabet{\mathsfit}{\encodingdefault}{\sfdefault}{m}{sl}
\SetMathAlphabet{\mathsfit}{bold}{\encodingdefault}{\sfdefault}{bx}{n}

%% file: camera_ready_2_appendix.tex
\clearpage
\appendix

\addcontentsline{toc}{section}{Appendix}
\part{Appendix}

{\Large{\textbf{\modelname: A Tabular Foundation Model for Generative Modelling}}}
\mtcsetdepth{parttoc}{3} 
\parttoc
\clearpage

\section{Summary of Related Work}
\label[appendix]{appendix:related_work}

\subsection{Tabular Data Generator}

\subsubsection{Dataset-specific Tabular Generator}
The development of dataset-specific tabular data generators has largely progressed from conventional resampling and non-parametric estimation towards increasingly expressive deep generative models~\cite{hansen2023reimagining, du2024systematic, tu2024causality, livieris2024evaluation, lautrup2025syntheval, kapar2025what, jiang2026tabstruct, jacob2026tabscm}.

Standard and non-parametric methods generate synthetic data by directly manipulating the observed data points~\cite{sauber2022use, chawla2002smote, neto2025tabsds}. For instance, SMOTE~\cite{chawla2002smote} synthesises new samples through interpolation between real samples, while TabSDS~\cite{neto2025tabsds} estimates feature-wise marginal distributions with interpolated order statistics and joint-probability-preserving shuffling. These methods are simple and often efficient, yet their expressiveness can be limited when modelling complex interactions across heterogeneous numerical and categorical features.

Another extensive line of work adapts deep generative models to the tabular modality~\cite{xu2019modeling, watson2023adversarial, kotelnikov2023tabddpm, shi2025tabdiff, zhang2025tabnat}. VAE-based methods, such as TVAE~\cite{xu2019modeling} and GOGGLE~\cite{liu2023goggle}, learn continuous latent representations of tabular data, with GOGGLE further incorporating graph neural networks to model feature dependencies. GAN-based methods, such as CTGAN~\cite{xu2019modeling}, use adversarial training and conditional generation to handle imbalanced discrete columns. Normalising-flow methods, such as Neural Spline Flows (NFlow)~\cite{durkan2019neural}, learn invertible transformations between tabular data and a tractable base distribution. Tree-based density estimators, such as ARF~\cite{watson2023adversarial}, iteratively refine synthetic data through adversarial random forests.
More recent methods explore generative objectives that can better capture complex mixed-type distributions. Diffusion-based generators, such as TabDDPM~\cite{kotelnikov2023tabddpm}, CDTD~\cite{mueller2025continuous}, TabSyn~\cite{zhang2023mixed}, and TabDiff~\cite{shi2025tabdiff}, progressively denoise corrupted tabular representations and have shown strong performance on heterogeneous tabular data. Specifically, TabDDPM introduces separate diffusion processes for numerical and categorical variables; CDTD unifies mixed-type generation through continuous diffusion with learned categorical embeddings; TabSyn applies diffusion in the latent space of a VAE; and TabDiff proposes a joint continuous-time diffusion framework for numerical and categorical features.
Alongside diffusion models, autoregressive tabular generators such as TabNAT~\cite{zhang2025tabnat} and TabularARGN~\cite{tiwald2025tabularargn} factorise the joint distribution into conditional distributions over features.

Despite this progress, dataset-specific generators remain inherently limited. Since they are fitted independently on each dataset, they cannot leverage transferable knowledge from other tabular datasets. This limitation is particularly restrictive for tabular data, where individual datasets are often small and domain-specific. Consequently, dataset-specific generators have to relearn from scratch on every dataset, which can lead to weak generalisation and overfitting to the training data.

\subsubsection{Tabular Foundation Generator}
The limitations of dataset-specific models motivate tabular foundation generators~\cite{lin2025ctsyn, grinsztajn2025tabpfn, borisov2022language}, which aim to acquire generalisable representations through pretraining across broad upstream data and then adapt to downstream tasks through techniques like finetuning or in-context inference.

In this work, we strictly follow the widely adopted definition of ``foundation models'' in prior work~\cite{bommasani2021opportunities, sun2026survey}. Specifically, a foundation model is defined as \textit{``any model that is trained on broad data (generally using self-supervision at scale) that can be adapted (e.g., fine-tuned) to a wide range of downstream tasks''}~\cite{bommasani2021opportunities}, and further note that \textit{``on a technical level, foundation models are enabled by transfer learning and scale.''}~\cite{bommasani2021opportunities}

Accordingly, our use of the term ``tabular foundation model'' does not require a model to operate exclusively in a zero-shot manner, such as through in-context learning. Instead, the key characteristics include broad pretraining, transfer learning, scale, and adaptability to downstream tasks. This interpretation is also consistent with recent discussions of foundation models for structured data, which recognise a range of adaptation mechanisms. For instance, prior work notes that \textit{``Typical adaptation strategies for tabular foundation models include in-context learning, prompt-based adaptation, and fine-tuning.''}~\cite{sun2026survey}, while studies of graph foundation models similarly highlight that \textit{``They (graph foundation models) enable strong performance with minimal task-specific supervision.''}~\cite{wang2025graph}.

One popular perspective to approach tabular data generation is to leverage foundation models from other modalities~\cite{borisov2022language, seedat2024curated, nguyen2024generating, zhao2025tabula, li2024tabsal}, such as Large Language Models (LLMs). For instance, GReaT~\cite{borisov2022language} converts each table row into a text sequence and finetunes an autoregressive language model to generate synthetic rows. This formulation can benefit from the strong generative ability of pretrained language models. However, it treats tabular data as a sequential modality, so the imposed feature order can introduce harmful order bias.

Specifically, prior work~\cite{jiang2026tabstruct} has shown that feature ordering can have a notable impact on autoregressive tabular generators. Autoregressive generators model the joint data distribution by factorising features according to a column ordering $\pi$. For tabular data, an ideal ordering $\pi^*$ may correspond to a topological ordering induced by the underlying structural causal model (SCM). However, since $\pi^*$ is typically unavailable in practice, using a mismatched ordering $\pi$ may lead to an autoregressive factorisation that is poorly aligned with the underlying causal structure, potentially compromising structural fidelity. Although prior work~\cite{borisov2022language} attempts to improve robustness to feature ordering by finetuning LLMs on randomly permuted column orders, such approaches can be computationally expensive and do not explicitly encourage alignment with the causal structure of the dataset. In other words, random shuffling exposes the model to multiple factorisations and can improve robustness to different orders, but it does not necessarily improve performance by identifying an order that is consistent with the underlying causal structure (i.e., the ground-truth topological order).
For instance, when a sampled ordering $\pi$ substantially deviates from, or even reverses, a topological ordering induced by the ground-truth causal structure, the resulting autoregressive factorisation may require the model to capture complex dependencies in directions that are inconsistent with the underlying causal mechanisms, thereby making it more difficult to preserve the global dependency structure~\cite{jiang2026tabstruct}.
Its generation quality may also depend heavily on textual serialisation and metadata~\cite{fang2024large}. As a result, LLM-based tabular foundation generators can be suboptimal for high-quality tabular data generation.

Another mainstream paradigm repurposes tabular foundation predictors as generators, such as TabPFN~\cite{grinsztajn2025tabpfn}, TabDPT~\cite{ma2025tabdpt}, Mitra~\cite{zhang2025mitra}, LimiX~\cite{zhang2025limix}, and TabICL~\cite{qu2026tabiclv2}. These models are originally designed for in-context tabular prediction. They can be converted into autoregressive generators by treating each feature as a prediction target and iteratively sampling feature values conditioned on previously generated features~\cite{prior2026tabpfnext, hollmann2025accurate}. However, their autoregressive generation again imposes a feature order like LLM-based tabular foundation generators, which may conflict with the underlying causal structures. Moreover, because such an autoregressive paradigm requires repeated in-context inference to sequentially generate each feature, these methods can be computationally expensive. Furthermore, their predictive pretraining objectives can also potentially bias the learned representations towards supervised target prediction, rather than holistic modelling of all features.

A further direction converts tabular foundation predictors into energy-based generators~\cite{margeloiu2024tabebm, ma2023tabpfgen}. For instance, TabEBM~\cite{margeloiu2024tabebm} builds class-specific energy surfaces from pretrained tabular predictors and samples synthetic data through the induced energy landscape. This design can achieve strong downstream predictive utility. However, because the energy functions are derived from predictive backbones, the model tends to overly emphasise local structures around the prediction target~\cite{jiang2026tabstruct}. As a result, it may overlook global dependency structures across all features, which are essential for high-quality tabular data generation.

One recent attempt at a generative tabular foundation model is CTSyn~\cite{lin2025ctsyn}, which pretrains a cross-dataset tabular generator by handling heterogeneous tables with a VAE and then modelling this latent space using diffusion, with metadata embedded by a pretrained LLM. Nevertheless, CTSyn still faces several limitations. Its dependence on metadata and LLM embeddings makes the model sensitive to metadata quality and LLM capability. Moreover, mapping variable-length feature sets into a predetermined fixed-length latent token sequence can obscure feature identity and weaken feature-specific semantics. In addition, as in many VAE-based latent diffusion models, the decoder is trained on clean encoder embeddings, whereas inference uses denoised embeddings produced by the diffusion model, creating a train-inference mismatch in latent space.

\textbf{Positioning of \modelname as a tabular foundation model.}
Due to these limitations, existing tabular foundation generators still struggle to match strong dataset-specific generators (\Cref{sec:experiment}). \modelname is designed to address this gap through a structure-aware latent diffusion framework for tabular foundation generation. Specifically, \modelname constructs a frozen structure-aware latent space and trains a denoising-aligned decoder directly on the denoised embeddings produced by the fitted diffusion transformer. Therefore, \modelname provides a more aligned and practical route towards tabular foundation models for generative modelling.

We further emphasise that \modelname is pretrained across the suggested real-world datasets~\cite{grinsztajn2025tabpfn}, followed by lightweight target-dataset finetuning, and employs a dataset-specific detokeniser within a transfer-learning framework. Under the widely adopted definition~\cite{bommasani2021opportunities}, the use of dataset-specific adaptation does not, by itself, preclude a model such as \modelname from being considered a foundation model. Indeed, several methods explicitly introduced as tabular foundation models also incorporate lightweight dataset-specific adaptation. For example, CTSyn~\cite{lin2025ctsyn} uses dataset-specific finetuning together with a dataset-specific reconstruction head; TabSTAR~\cite{arazi2026tabstar} employs dataset-specific LoRA finetuning; iLTM~\cite{bonet2026iltm} uses dataset-specific finetuning; and LaTable~\cite{van2024latable} studies cross-dataset pretraining followed by dataset-specific finetuning.
Therefore, we view \modelname as an instantiation of a generative tabular foundation model, following the common paradigm of cross-dataset pretraining with lightweight downstream adaptation.

\subsection{Tabular Data Evaluation}
\label[appendix]{appendix:tabular_data_evaluation}

Following prior studies~\cite{zhang2023mixed, shi2025tabdiff, hansen2023reimagining, tu2024causality, livieris2024evaluation, lautrup2025syntheval, kapar2025what, jiang2026tabstruct, margeloiu2024tabebm}, we evaluate the quality of synthetic tabular data with well-defined metrics along four complementary dimensions: structural fidelity, ML efficacy, density estimation, and privacy preservation. These dimensions are selected because each captures a distinct and important aspect of data quality, as detailed below:

\textbf{Structural fidelity} evaluates whether synthetic data preserves the global inter-feature dependency structures of real data~\cite{jiang2026tabstruct}. This dimension is particularly important for tabular generation because dependency structures have been shown to provide an effective prior for tabular data, whereas preserving only distributional properties may fail to capture inter-feature dependency interactions.
We quantify structural fidelity using \textit{global utility}~\cite{jiang2026tabstruct}, which aggregates the predictive performance of an ensemble of nine tuned downstream models rather than relying on a single predictor.
Specifically, the ensemble comprises three standard baselines: Logistic Regression (LR)~\cite{cox1958regression}, KNN~\cite{fix1985discriminatory}, and MLP~\cite{gorishniy2021revisiting}; five tree-based methods: Random Forest (RF)~\cite{breiman2001random}, Extra Trees~\cite{agtabular}, LightGBM~\cite{ke2017lightgbm}, CatBoost~\cite{prokhorenkova2018catboost}, and XGBoost~\cite{chen2016xgboost}; and one PFN-based method, TabPFN~\cite{hollmann2025accurate}.
Prior work~\cite{jiang2026tabstruct} has shown that aggregating performance across a diverse set of downstream predictors can mitigate evaluation bias associated with any individual model when assessing tabular generators.

To avoid further ambiguity, we provide a high-level summary of the three-step argument established by TabStruct~\cite{jiang2026tabstruct} for why global utility can serve as an empirically effective measure for the causal structural fidelity of synthetic data with respect to the underlying SCM. The full details are available in their original paper.
\begin{itemize}[topsep=0pt, leftmargin=10pt, itemsep=0pt]
\item \textbf{Step 1: Decomposition into local and global structures.}
Structural fidelity is decomposed into local conditional independence (\textit{local CI}), which concerns relationships around an individual feature, and global conditional independence (\textit{global CI}), which concerns relationships across all features. Global CI evaluates whether synthetic data preserve the conditional-independence structure across all features at the level of Markov equivalence class~\cite{tu2024causality}.
\item \textbf{Step 2: Local CI and local utility.}
Correctly distinguishing informative from uninformative features through d-separation and d-connection is closely related to Markov-blanket-based feature selection~\cite{fu2010markov} and can improve predictive performance for the target feature, i.e., local utility~\cite{cherepanova2023performance}. TabStruct~\cite{jiang2026tabstruct} empirically demonstrates a strong correlation between local CI and local utility.
\item \textbf{Step 3: Empirical aggregation to the global level.}
Motivated by the strong relationship between local CI and local utility, TabStruct~\cite{jiang2026tabstruct} further investigates whether an analogous relationship holds between global CI and global utility. Across 13 generators and six datasets with expert-validated SCMs, TabStruct~\cite{jiang2026tabstruct} shows that global utility exhibits a substantially stronger correlation with global CI than conventional statistical, privacy, or downstream-utility metrics. These results provide empirical evidence that global utility can serve as an effective proxy for global structural fidelity when the underlying SCM is unknown.
\end{itemize}
Therefore, global utility is designed by prior work~\cite{jiang2026tabstruct} to evaluate causal structural fidelity at the Markov-equivalence level. It does not guarantee the orientation or full identifiability of every causal edge, and we do not aim to make such claims in this work. Instead, a higher global utility indicates that the synthetic data can better preserve the global dependency structures of real data empirically.

\textbf{ML efficacy} measures whether synthetic data can support downstream predictive modelling as effectively as real data~\cite{margeloiu2024tabebm, zhang2023mixed}. This dimension reflects the practical utility of synthetic data when it is used as a substitute for real data in downstream tasks.
Following prior studies~\cite{xu2019modeling, zhang2023mixed, shi2025tabdiff, jiang2026tabstruct, zhang2025tabnat}, we adopt the ``train-on-synthetic, test-on-real'' strategy: downstream predictors are trained on synthetic data and evaluated on real test data. 
To mitigate evaluation bias from any single downstream model, we quantify ML efficacy using \textit{local utility}, which aggregates the performance of an ensemble of nine tuned downstream predictors~\cite{jiang2026tabstruct}. We use the same set of predictors as that employed to compute global utility, ensuring a consistent evaluation protocol across the two dimensions.
A higher local utility indicates that the synthetic data better supports downstream predictive modelling, i.e., the synthetic data can retain predictive performance when used to train downstream predictors.

\textbf{Density estimation} assesses the distributional discrepancy between real and synthetic data. We include this dimension because high-quality synthetic data should capture both low-order statistics, such as feature-level marginals and pairwise correlations, and high-order distributional properties, such as sample-level similarity and diversity.
Following prior studies~\cite{zhang2023mixed, shi2025tabdiff, jiang2026tabstruct}, we evaluate density estimation using four metrics from two categories.
For low-order metrics, we use \textit{Shape} and \textit{Trend}~\cite{wust2011sdmetrics}: \textit{Shape} measures how well synthetic data replicates the marginal density of each column, and \textit{Trend} measures how well the synthetic data captures pairwise correlations between features.
For high-order metrics, we use \mbox{\textit{$\alpha$-precision}} and \mbox{\textit{$\beta$-recall}}~\cite{alaa2022faithful}: \textit{$\alpha$-precision} measures whether generated samples lie within the support of the real-data distribution and reflects sample similarity, and \textit{$\beta$-recall} measures how comprehensively the synthetic data cover the real-data distribution and reflects diversity and mode coverage.

\textbf{Privacy preservation} evaluates the extent to which synthetic data avoids directly copying or memorising samples from the real training data~\cite{mckenna2019graphical, jordon2018pate, truda2023generating, stoian2025survey, hu2024sok}. This dimension is essential because a generator with strong overall performance may still pose privacy risks if it closely duplicates training samples.
Following prior studies~\cite{jiang2026tabstruct, jiang2025well}, we measure privacy preservation using two metrics. 
First, we use \textit{Authenticity}~\cite{alaa2022faithful}, which estimates whether generated samples represent novel draws from the learned distribution rather than memorised training examples, and a higher value indicates that the generator is less likely to overfit to the real training data.
Second, we use the \textit{Distance to Closest Record Score (DCR score)}~\cite{wust2011sdmetrics}, which measures the proximity of synthetic samples to individual real records and therefore helps identify potential direct copying or memorisation. A higher DCR score indicates that synthetic samples are further from their nearest real training records and are therefore less likely to be direct duplications.
We note that privacy metrics should be interpreted together with the other dimensions, since excessive distance from the real data may also indicate a poor fit to the real data distribution.

\clearpage
\section{Extended Model Design of \modelname}
\label[appendix]{appendix:model_design}

In this section, we provide the complete training and inference procedures of \modelname, together with detailed implementation choices and model configurations. We first present the pseudocode for pretraining, fitting, and generation (\Cref{appendix:pseudocode}). We then describe the implementation details of certain components (\Cref{appendix:implementation_details}). Next, we summarise the configurations used throughout the experiments of \modelname.

\subsection{Pseudocode for Training and Inference}
\label[appendix]{appendix:pseudocode}

As illustrated in~\Cref{sec:experiment}, \modelname follows a three-phase workflow. First, during pretraining (\Cref{alg:pretraining}), it learns transferable latent representations from multiple real-world tabular datasets. Second, when applied to an unseen dataset (\Cref{alg:fitting}), it performs lightweight fitting by updating the diffusion transformer and decoder using the available training data. Third, during inference (\Cref{alg:inference}), it generates synthetic tabular data by progressively denoising latent embeddings with noise, and decoding the final denoised embeddings into mixed-type feature values.

\begin{algorithm}[!htbp]
\caption{Pretraining \modelname across multiple tabular datasets}
\label{alg:pretraining}

\DontPrintSemicolon

\KwIn{
Pretraining datasets $\{\mX_m\}_{m=1}^{M}$;
frozen tokeniser $\mathcal{S}_{\psi}$;
frozen encoder $\mathcal{E}_{\omega}$;
diffusion denoiser $\mathcal{G}_{\theta}$;
decoder transformer $\mathcal{D}_{\phi}$;
diffusion steps $S_{\text{diff}}$;
decoder steps $S_{\text{dec}}$;
pretraining rounds $R$
}
\KwOut{Pretrained diffusion denoiser $\mathcal{G}_{\theta}$ and decoder transformer $\mathcal{D}_{\phi}$}

\ForEach{pretraining dataset $\mX_m$}{
    $\mX_m \leftarrow$ preprocess$(\mX_m)$ \tcp*{imputation and feature encoding}

    $\mH_m \leftarrow \mathcal{E}_{\omega}(\mathcal{S}_{\psi}(\mX_m))$ \tcp*{compute frozen latent embeddings once}

    $(\boldsymbol{\mu}_{H,m},\boldsymbol{s}_{H,m}) \leftarrow$ statistics$(\mH_m)$ \tcp*{dataset-specific latent statistics}

    $\mZ_{0,m} \leftarrow (\mH_m-\boldsymbol{\mu}_{H,m})/\boldsymbol{s}_{H,m}$ \tcp*{normalised clean embeddings}

    Initialise dataset-specific detokeniser $\mathcal{R}_{\eta_m}$ according to feature types and categorical cardinalities\;
}

\For{$r \leftarrow 1$ \KwTo $R$}{

    \ForEach{pretraining dataset $\mX_m$}{

        \For{$s \leftarrow 1$ \KwTo $S_{\text{diff}}$}{

            $\mZ_0 \leftarrow$ sample\_batch$(\mZ_{0,m})$\;

            $\sigma \leftarrow \exp(p_{\text{mean}}+p_{\text{std}}\epsilon_{\sigma})$,
            where $\epsilon_{\sigma}\sim\mathcal{N}(0,1)$ \tcp*{sample noise level}

            $\mE \sim \mathcal{N}(\mathbf{0},\mI)$ \tcp*{sample Gaussian noise}

            $\mZ_{\sigma} \leftarrow \mZ_0+\sigma\mE$ \tcp*{forward diffusion}

            $\widehat{\mZ}_0 \leftarrow \mathcal{G}_{\theta}(\mZ_{\sigma},\sigma)$ \tcp*{denoise latent embeddings}

            $\theta \leftarrow \theta-\nabla_{\theta}\mathcal{L}_{\text{diff}}(\widehat{\mZ}_0,\mZ_0)$ \tcp*{update diffusion denoiser}
        }

        \For{$s \leftarrow 1$ \KwTo $S_{\text{dec}}$}{

            $(\mX,\mZ_0) \leftarrow$ sample\_batch$(\mX_m,\mZ_{0,m})$\;

            $\sigma \leftarrow \exp(p_{\text{mean}}+p_{\text{std}}\epsilon_{\sigma})$,
            where $\epsilon_{\sigma}\sim\mathcal{N}(0,1)$\;

            $\mE \sim \mathcal{N}(\mathbf{0},\mI)$\;

            $\mZ_{\sigma} \leftarrow \mZ_0+\sigma\mE$\;

            $\widehat{\mZ} \leftarrow \mathcal{G}_{\theta}(\mZ_{\sigma},\sigma)$ \tcp*{produce denoised embeddings}

            $\widehat{\mH} \leftarrow \widehat{\mZ}\odot\boldsymbol{s}_{H,m}+\boldsymbol{\mu}_{H,m}$ \tcp*{inverse latent normalisation}

            $\mU \leftarrow \mathcal{D}_{\phi}(\widehat{\mH})$ \tcp*{refine denoised embeddings}

            $\widehat{\mX} \leftarrow \mathcal{R}_{\eta_m}(\mU)$ \tcp*{detokenise into mixed-type features}

            $(\phi,\eta_m) \leftarrow (\phi,\eta_m)-\nabla_{\phi,\eta_m}\mathcal{L}_{\text{recon}}(\widehat{\mX},\mX)$ \tcp*{update decoder and detokeniser}
        }
    }
}

\Return{$\mathcal{G}_{\theta},\mathcal{D}_{\phi}$}
\end{algorithm}

\begin{algorithm}[!htbp]
\caption{Fitting \modelname to an unseen dataset}
\label{alg:fitting}

\DontPrintSemicolon

\KwIn{
Unseen training data $\mX_{\text{Train}}$;
pretrained diffusion denoiser $\mathcal{G}_{\theta}$;
pretrained decoder transformer $\mathcal{D}_{\phi}$;
frozen tokeniser $\mathcal{S}_{\psi}$;
frozen encoder $\mathcal{E}_{\omega}$;
fitting steps $S_{\text{fit,diff}}$ and $S_{\text{fit,dec}}$
}
\KwOut{
Fitted diffusion denoiser $\mathcal{G}_{\theta^{\star}}$;
fitted decoder transformer $\mathcal{D}_{\phi^{\star}}$;
dataset-specific detokeniser $\mathcal{R}_{\eta^{\star}}$;
latent statistics $(\boldsymbol{\mu}_{H,\star},\boldsymbol{s}_{H,\star})$
}

$\mX_{\text{Train}} \leftarrow$ preprocess$(\mX_{\text{Train}})$ \tcp*{imputation and feature encoding}

$\mH_{\text{Train}} \leftarrow \mathcal{E}_{\omega}(\mathcal{S}_{\psi}(\mX_{\text{Train}}))$ \tcp*{compute frozen latent embeddings once}

$(\boldsymbol{\mu}_{H,\star},\boldsymbol{s}_{H,\star}) \leftarrow$ statistics$(\mH_{\text{Train}})$ \tcp*{dataset-specific latent statistics}

$\mZ_{0,\star} \leftarrow (\mH_{\text{Train}}-\boldsymbol{\mu}_{H,\star})/\boldsymbol{s}_{H,\star}$ \tcp*{normalised clean embeddings}

Initialise dataset-specific detokeniser $\mathcal{R}_{\eta^{\star}}$ according to feature types and categorical cardinalities\;

\For{$s \leftarrow 1$ \KwTo $S_{\text{fit,diff}}$}{

    $\mZ_0 \leftarrow$ sample\_batch$(\mZ_{0,\star})$\;

    $\sigma \leftarrow \exp(p_{\text{mean}}+p_{\text{std}}\epsilon_{\sigma})$,
    where $\epsilon_{\sigma}\sim\mathcal{N}(0,1)$ \tcp*{sample noise level}

    $\mE \sim \mathcal{N}(\mathbf{0},\mI)$ \tcp*{sample Gaussian noise}

    $\mZ_{\sigma} \leftarrow \mZ_0+\sigma\mE$ \tcp*{forward diffusion}

    $\widehat{\mZ}_0 \leftarrow \mathcal{G}_{\theta}(\mZ_{\sigma},\sigma)$ \tcp*{denoise latent embeddings}

    $\theta \leftarrow \theta-\nabla_{\theta}\mathcal{L}_{\text{diff}}(\widehat{\mZ}_0,\mZ_0)$ \tcp*{fit diffusion denoiser}
}

$\theta^{\star} \leftarrow \theta$ \tcp*{save fitted diffusion denoiser}

\For{$s \leftarrow 1$ \KwTo $S_{\text{fit,dec}}$}{

    $(\mX,\mZ_0) \leftarrow$ sample\_batch$(\mX_{\text{Train}},\mZ_{0,\star})$\;

    $\sigma \leftarrow \exp(p_{\text{mean}}+p_{\text{std}}\epsilon_{\sigma})$,
    where $\epsilon_{\sigma}\sim\mathcal{N}(0,1)$\;

    $\mE \sim \mathcal{N}(\mathbf{0},\mI)$\;

    $\mZ_{\sigma} \leftarrow \mZ_0+\sigma\mE$\;

    $\widehat{\mZ} \leftarrow \mathcal{G}_{\theta^{\star}}(\mZ_{\sigma},\sigma)$ \tcp*{produce denoised embeddings}

    $\widehat{\mH} \leftarrow \widehat{\mZ}\odot\boldsymbol{s}_{H,\star}+\boldsymbol{\mu}_{H,\star}$ \tcp*{inverse latent normalisation}

    $\mU \leftarrow \mathcal{D}_{\phi}(\widehat{\mH})$ \tcp*{refine denoised embeddings}

    $\widehat{\mX} \leftarrow \mathcal{R}_{\eta^{\star}}(\mU)$ \tcp*{detokenise into mixed-type features}

    $(\phi,\eta^{\star}) \leftarrow (\phi,\eta^{\star})-\nabla_{\phi,\eta^{\star}}\mathcal{L}_{\text{recon}}(\widehat{\mX},\mX)$ \tcp*{fit decoder and detokeniser}
}

$\phi^{\star} \leftarrow \phi$ \tcp*{save fitted decoder transformer}

\Return{$\mathcal{G}_{\theta^{\star}},\mathcal{D}_{\phi^{\star}},\mathcal{R}_{\eta^{\star}},(\boldsymbol{\mu}_{H,\star},\boldsymbol{s}_{H,\star})$}
\end{algorithm}

\begin{algorithm}[!htbp]
\caption{Synthetic data generation with fitted \modelname}
\label{alg:inference}

\DontPrintSemicolon

\KwIn{
Fitted diffusion denoiser $\mathcal{G}_{\theta^{\star}}$;
fitted decoder transformer $\mathcal{D}_{\phi^{\star}}$;
dataset-specific detokeniser $\mathcal{R}_{\eta^{\star}}$;
latent embeddings $\mZ_{0,\star}$ and statistics $(\boldsymbol{\mu}_{H,\star},\boldsymbol{s}_{H,\star})$;
desired synthetic sample size $N_{\text{syn}}$;
feature count $D+1$;
latent dimension $k$;
reverse diffusion steps $T$
}
\KwOut{Synthetic tabular dataset $\mX_{\text{Syn}}$}

$\widetilde{\mZ}_0 \leftarrow
\operatorname{sample\_rows}(\mZ_{0,\star},N_{\text{syn}})$
\tcp*{sample clean latent embeddings}

$\mZ_{\sigma_1}\mid\widetilde{\mZ}_0
\sim\mathcal{N}\left(
\widetilde{\mZ}_0,\sigma_{\max}^{2}\mI
\right)$ with shape
$\mathbb{R}^{N_{\text{syn}}\times(D+1)\times k}$
\tcp*{initial noisy latents}

Construct decreasing schedule $\sigma_1>\sigma_2>\cdots>\sigma_T$, with $\sigma_1=\sigma_{\max}$ and $\sigma_T=\sigma_{\min}$\;

\For{$t \leftarrow 1$ \KwTo $T-1$}{

    $\widehat{\mZ}_t \leftarrow \mathcal{G}_{\theta^{\star}}(\mZ_{\sigma_t},\sigma_t)$ \tcp*{predict denoised latent embeddings}

    $\mathbf{d}_t \leftarrow (\mZ_{\sigma_t}-\widehat{\mZ}_t)/\sigma_t$ \tcp*{EDM probability-flow direction}

    $\mZ_{\sigma_{t+1}} \leftarrow \mZ_{\sigma_t}+(\sigma_{t+1}-\sigma_t)\mathbf{d}_t$ \tcp*{reverse diffusion update}
}

$\widehat{\mH}_{\text{syn}} \leftarrow \mZ_{\sigma_T}\odot\boldsymbol{s}_{H,\star}+\boldsymbol{\mu}_{H,\star}$ \tcp*{inverse latent normalisation}

$\mU_{\text{syn}} \leftarrow \mathcal{D}_{\phi^{\star}}(\widehat{\mH}_{\text{syn}})$ \tcp*{generation-ready embeddings}

$\widehat{\mX}_{\text{syn}} \leftarrow \mathcal{R}_{\eta^{\star}}(\mU_{\text{syn}})$ \tcp*{decode mixed-type features}

$\mX_{\text{Syn}} \leftarrow$ inverse\_preprocess$(\widehat{\mX}_{\text{syn}})$ \tcp*{recover numerical scales and categorical labels}

\Return{$\mX_{\text{Syn}}$}
\end{algorithm}

\subsection{Implementation Details}
\label[appendix]{appendix:implementation_details}

\subsubsection{Feature Tokenisation}
Following TabPFN~\cite{grinsztajn2025tabpfn}, the tokeniser constructs distinguishable feature tokens within a shared latent space by introducing a learnable base vector $\vu \in \mathbb{R}^{k}$ that is shared across all features. It then builds feature-specific perturbations as
\begin{equation}
    \mR = \mW\mP \in \mathbb{R}^{k \times (D+1)},
\end{equation}
where $\mW \in \mathbb{R}^{k \times k'}$ is a learnable projection matrix with $k' < k$, and $\mP \in \mathbb{R}^{k' \times (D+1)}$ is a randomly generated matrix. The $j$-th column $\vr_j \in \mathbb{R}^{k}$ of $\mR$ serves as a feature-specific perturbation for the $j$-th feature. For the $i$-th sample, the token corresponding to the $j$-th feature is then computed as
\begin{equation}
    \vt_{j}^{(i)}
    =
    x_{j}^{(i)}(\vu+\vr_j)
    \in
    \mathbb{R}^{k}.
\end{equation}
We note that, for the $j$-th numerical feature, the value after Z-score normalisation is denoted by $x_j^{(i)}$. For the $j$-th categorical feature with cardinality $C_j$, ordinal encoding maps each category to a unique integer identifier, also denoted by $x_j^{(i)}$. Consequently, the number of input tokens is independent of feature cardinality, as each feature contributes exactly one token.
Thus, the tokenised representation of the $i$-th sample is
\begin{equation}
    \mT^{(i)}
    =
    [
    \vt_{1}^{(i)},
    \dots,
    \vt_{D+1}^{(i)}
    ]^{\mathsf{T}}
    \in
    \mathbb{R}^{(D+1)\times k},
\end{equation}
and the tokenised representation of the full dataset is
\begin{equation}
    \mT
    =
    [\mT^{(1)},\dots,\mT^{(N)}]^{\mathsf{T}}
    \in
    \mathbb{R}^{N\times(D+1)\times k}.
\end{equation}
This construction allows all features to be embedded into a common latent space while retaining feature identity through feature-specific perturbations, without relying on metadata such as column names or textual descriptions.

\subsubsection{Leave-one-fold-out Latent Extraction}
Standard PFN models are originally designed for in-context prediction~\cite{grinsztajn2025tabpfn, qu2026tabiclv2, ma2025tabdpt}, where query samples (i.e., test data) are embedded in the context of observed samples (i.e., training data). Thus, standard PFN models require query samples to extract latent embeddings. To obtain latent embeddings for the available training samples themselves, \modelname adopts a leave-one-fold-out extraction strategy~\cite{ye2025closer}. Specifically, the training rows are partitioned into multiple folds. For each fold, the remaining folds are used as context rows, while the holdout fold is treated as query rows whose latent representations are extracted. The latent representations from all holdout folds are then concatenated to form the complete latent embeddings. For classification datasets, we use the TabPFN classifier checkpoint to extract latent embeddings; for regression datasets, we use the TabPFN regressor checkpoint.
\modelname uses $L_{\text{enc}}=12$ encoder layers for both classification and regression tasks.
This strategy enables \modelname to extract latent embeddings for the training data without requiring additional unseen query data.

\subsubsection{Latent Normalisation and Caching}
For each dataset, the frozen latent embeddings are computed only once and cached before optimising the diffusion transformer or the decoder. This caching strategy avoids repeated PFN forward passes, thereby substantially reducing the computational cost of both pretraining and fitting.

Given cached embeddings $\mH \in \mathbb{R}^{N\times(D+1)\times k}$, we compute the latent normalisation statistics, i.e., the mean and standard deviation, over the sample dimension:
\begin{equation}
    \boldsymbol{\mu}_{H,j,a}
    =
    \frac{1}{N}\sum_{i=1}^{N} H_{i,j,a},
    \quad
    \boldsymbol{s}_{H,j,a}
    =
    \sqrt{
    \frac{1}{N}\sum_{i=1}^{N}
    (H_{i,j,a}-\boldsymbol{\mu}_{H,j,a})^2
    +\epsilon
    }
\end{equation}
where $j$ indexes features and $a$ indexes latent dimensions. The normalised latent embeddings are then computed as
\begin{equation}
    \mZ_0
    =
    \frac{\mH-\boldsymbol{\mu}_{H}}{\boldsymbol{s}_{H}}.
\end{equation}
The diffusion model is trained in this normalised latent space. During decoding, denoised latent embeddings are mapped back to the original latent scale by
\begin{equation}
    \widehat{\mH}
    =
    \widehat{\mZ}\odot\boldsymbol{s}_{H}
    +
    \boldsymbol{\mu}_{H}
\end{equation}
where $\odot$ denotes element-wise multiplication (i.e., the Hadamard product). The latent normalisation statistics are saved with the fitted generator, since they are required during synthetic data generation.

\subsubsection{Feature Decoding}
\label{appendix:feature_decoding}
During detokenisation, \modelname instantiates a dataset-specific detokeniser with feature-specific output heads.
For each numerical feature, the corresponding detokenisation head maps its latent representation to a scalar via~\Cref{eq:detoken}. We then apply the inverse Z-score transformation using the mean and standard deviation computed from the training data to recover the value in the original feature space.
For each categorical feature with cardinality $C$, the corresponding detokenisation head produces a $C$-dimensional logit vector.

The detokeniser is trained using the mixed-type reconstruction objective in~\Cref{eq:objective_recon}, which combines mean squared error for numerical features with cross-entropy loss for categorical features.
During generation, we apply $\arg\max$ to the logits to obtain the predicted category identifier and then map this identifier back to the corresponding original category using the identifier-to-category mapping cached during preprocessing.
This design enables features with different cardinalities to be handled within a unified pipeline while requiring only lightweight dataset-specific adaptations.

\subsubsection{Pretraining Cost}
% {
All experiments were conducted on a machine with four NVIDIA H100 Tensor Core GPUs (80\,GB), a 64-core Intel Xeon CPU at 2.20\,GHz, and Ubuntu 24.04.4 LTS. Under this setup, pretraining \modelname required approximately 605 GPU hours (i.e., about 25 H100-days).
% }

\subsubsection{Detailed Model Configurations}
\label[appendix]{appendix:model_configurations}

We further provide the default configurations of \modelname, including the model architecture (\Cref{tab:arch_config}), diffusion-process setup (\Cref{tab:diffusion_config}), and optimisation settings (\Cref{tab:optimisation_config}).
We note that all model configurations are fixed based solely on validation performance on the 43 pretraining datasets. Consequently, the ablation studies conducted on the 45 evaluation datasets do not influence any model design or hyperparameter choices.

\begin{table}[!htbp]
\centering
\caption{Default architecture configurations of \modelname.}
\label{tab:arch_config}
\begin{tabular}{lr}
\toprule
\textbf{Component} & \textbf{Default configuration} \\
\midrule
\rowcolor{Gainsboro!60}
Feature encoder backbone & \mbox{Real-TabPFN-2.5} \\
Encoder depth $L_{\text{enc}}$ & $12$ \\
\rowcolor{Gainsboro!60}
Latent dimension $k$ & 192 \\
Diffusion transformer layers & $4$ \\
\rowcolor{Gainsboro!60}
Diffusion attention heads & $4$ \\
Decoder transformer layers $L_{\text{dec}}$ & $4$ \\
\bottomrule
\end{tabular}
\end{table}

\begin{table}[!htbp]
\centering
\caption{Default diffusion and inference configurations of \modelname.}
\label{tab:diffusion_config}
\begin{tabular}{lr}
\toprule
\textbf{Hyperparameter} & \textbf{Default value} \\
\midrule
\rowcolor{Gainsboro!60}
$\sigma_{\min}$ & $0.002$ \\
$\sigma_{\max}$ & $80.0$ \\
\rowcolor{Gainsboro!60}
$\sigma_{\text{init}}$ & $0.1$ \\
$\rho$ & $7.0$ \\
\rowcolor{Gainsboro!60}
$\sigma_{\text{data}}$ & $1.0$ \\
Noise conditioning & $\log\sigma/4$ \\
\rowcolor{Gainsboro!60}
Reverse diffusion steps & $10$ \\
Inference-time encoder usage & None \\
\bottomrule
\end{tabular}
\end{table}

\begin{table}[!htbp]
\centering
\caption{Default optimisation configurations of \modelname.}
\label{tab:optimisation_config}
\begin{tabular}{llr}
\toprule
\textbf{Stage} & \textbf{Configuration} & \textbf{Value} \\
\midrule
\rowcolor{Gainsboro!60}
Pretraining & Number of rounds & $10$ \\
Diffusion pretraining & Optimiser & AdamW \\
\rowcolor{Gainsboro!60}
Diffusion pretraining & Learning rate & $10^{-3}$ \\
Diffusion pretraining & Weight decay & $10^{-5}$ \\
\rowcolor{Gainsboro!60}
Diffusion pretraining & Steps per dataset & $5{,}000$ \\
Decoder pretraining & Optimiser & SGD \\
\rowcolor{Gainsboro!60}
Decoder pretraining & Learning rate & $10^{-1}$ \\
Decoder pretraining & Weight decay & $5\times10^{-5}$ \\
\rowcolor{Gainsboro!60}
Decoder pretraining & Steps per dataset & $5{,}000$ \\
Pretraining and fitting & Batch size & $3{,}172$ \\
\rowcolor{Gainsboro!60}
Pretraining and fitting & Gradient clipping & Enabled \\
Pretraining and fitting & LR scheduler & Plateau-based scheduler \\
\rowcolor{Gainsboro!60}
Fitting on unseen datasets & Learning rate & $10^{-6}$ \\
Fitting diffusion on unseen datasets & Steps & $100$ \\
\rowcolor{Gainsboro!60}
Fitting decoder on unseen datasets & Steps & $100$ \\
\bottomrule
\end{tabular}
\end{table}

\clearpage
\section{Reproducibility}
\label[appendix]{appendix:reproducibility}

\subsection{Pretraining Datasets}
\label[appendix]{appendix:pretraining_dataset}

To reduce the risk of data leakage during evaluation on the benchmark datasets, we pretrain \modelname on the same 43 real-world datasets used by \mbox{Real-TabPFN-2.5}. \Cref{tab:pretraining_dataset} lists the datasets curated for pretraining, along with their sources and access links.

\begin{table}[!ht]
\centering
\caption{The 43 pretraining datasets and their sources, adopted from \mbox{Real-TabPFN-2.5}~\cite{grinsztajn2025tabpfn}.}
\label{tab:pretraining_dataset}
\vspace{-4mm}
\begin{longtable}{p{10cm}p{4cm}}
\toprule
Name & Source \\
\midrule
\endfirsthead
\toprule
Name & Source \\
\midrule
\endhead

\href{https://www.openml.org/search?type=data&sort=runs&status=active&id=1459}{artificial-characters} & OpenML \\
\href{https://www.openml.org/search?type=data&status=active&id=251}{BNG(breast-w)} & OpenML \\
\href{https://www.openml.org/search?type=data&status=active&id=137}{BNG(tic-tac-toe)} & OpenML \\
\href{https://www.openml.org/d/40668}{connect\_4} & OpenML \\
\href{https://www.openml.org/search?type=data&sort=runs&status=active&id=1471}{eeg-eye-state} & OpenML \\
\href{https://openml.org/search?type=data&status=active&id=43551}{Employee-Turnover-at-TECHCO} & OpenML \\
\href{https://openml.org/search?type=data&status=active&id=1044}{eye\_movements} & OpenML \\
\href{https://www.openml.org/search?type=data&status=active&id=41787&sort=runs}{FOREX\_eurpln-hour-High} & OpenML \\
\bottomrule
\end{longtable}
\end{table}

\subsection{Benchmark Datasets}
\label[appendix]{appendix:benchmark_dataset}

We curate 45 challenging real-world benchmark datasets from public tabular-data repositories and benchmark suites, including TabArena~\citep{erickson2025tabarena}, TabStruct~\citep{jiang2026tabstruct}, and OpenML (\url{https://www.openml.org/search?type=data&sort=runs}).

Following the well-established benchmarks~\citep{erickson2025tabarena, jiang2026tabstruct}, our experiments mainly aim to demonstrate generalisation to unseen datasets rather than to claim robustness under controlled distribution shifts.

Therefore, the dataset selection follows four major criteria:
First, the datasets are non-trivial, meaning that generative models cannot readily achieve evaluation results comparable to those obtained using the real training data.
Second, the datasets span diverse application domains. For example, ``Credit'' concerns business applications, whereas ``Biodeg'' relates to biological studies.
Third, the datasets were not used during the meta-validation stage of TabPFN~\citep{hollmann2025accurate}, reducing the risk that their underlying structures were implicitly exposed during the development or pretraining of TabPFN.
Fourth, we manually verify that none of the 45 benchmark datasets was used during \modelname pretraining, i.e., that none duplicates any dataset listed in~\Cref{tab:pretraining_dataset}.
Together, these criteria ensure that the benchmark provides a fair, diverse, and comprehensive evaluation of \modelname. All datasets are publicly available, with further details provided in~\Cref{tab:benchmark_dataset_classification} and~\Cref{tab:benchmark_dataset_regression}.

\begin{table}[!htbp]
\centering
\caption{Details of 31 real-world benchmark classification datasets.}
\label{tab:benchmark_dataset_classification}

\resizebox{\textwidth}{!}{
\begin{tabular}{lrrrrrrr}

\toprule

Dataset & Source & ID    & \# Samples ($N$) & \# Features ($D + 1$) & \# Numerical & \# Categorical & \# Classes \\

\midrule

\rowcolor{Gainsboro!60}
BankChurn                                              & OpenML & 46911 & 10,000          & 11                   & 6           & 5             & 2         \\
Bankruptcy                                             & OpenML & 46962 & 6,819           & 95                   & 94          & 1             & 2         \\
\rowcolor{Gainsboro!60}
Biodeg                                                 & OpenML & 46952 & 1,054           & 42                   & 36          & 6             & 2         \\
Card                                                   & OpenML & 46919 & 30,000          & 24                   & 20          & 4             & 2         \\
\rowcolor{Gainsboro!60}
Churn                                                  & OpenML & 46915 & 5,000           & 20                   & 15          & 5             & 2         \\
Coil2000                                               & OpenML & 46916 & 9,822           & 86                   & 80          & 6             & 2         \\
\rowcolor{Gainsboro!60}
Company                                                & OpenML & 46950 & 5,910           & 65                   & 64          & 1             & 2         \\
\bottomrule

\end{tabular}
}
\end{table}

\begin{table}[!htbp]
\centering
\caption{Details of 14 real-world benchmark regression datasets.}
\label{tab:benchmark_dataset_regression}

\resizebox{\textwidth}{!}{
\begin{tabular}{lrrrrrrr}

\toprule

Dataset & Source & ID    & \# Samples ($N$) & \# Features ($D + 1$) & \# Numerical & \# Categorical \\

\midrule

\rowcolor{Gainsboro!60}
Airfoil                                                & OpenML & 46904 & 1,503           & 6                    & 5           & 1             \\
California                                             & OpenML & 43939 & 20,640          & 10                   & 9           & 1             \\
\rowcolor{Gainsboro!60}
Concrete                                               & OpenML & 46917 & 1,030           & 9                    & 9           & 0             \\
Diamonds                                               & OpenML & 46923 & 53,940          & 10                   & 7           & 3             \\
\rowcolor{Gainsboro!60}
Fiat500                                                & OpenML & 46907 & 1,538           & 8                    & 7           & 1             \\
Fish                                                   & OpenML & 46954 & 907             & 7                    & 7           & 0             \\
\rowcolor{Gainsboro!60}
FoodDelivery                                           & OpenML & 46928 & 45,451          & 10                   & 7           & 3             \\
\bottomrule

\end{tabular}
}
\end{table}

\subsection{Data Processing}
\label[appendix]{appendix:data_processing}

\textbf{Data splitting strategies for 43 pretraining datasets.}
During pretraining, \modelname treats all pretraining datasets as training and validation data. In other words, all samples in the pretraining datasets are used to develop the model, without further test splits.

\begin{figure}[!ht]
    \centering
    \includegraphics[width=0.9\linewidth]{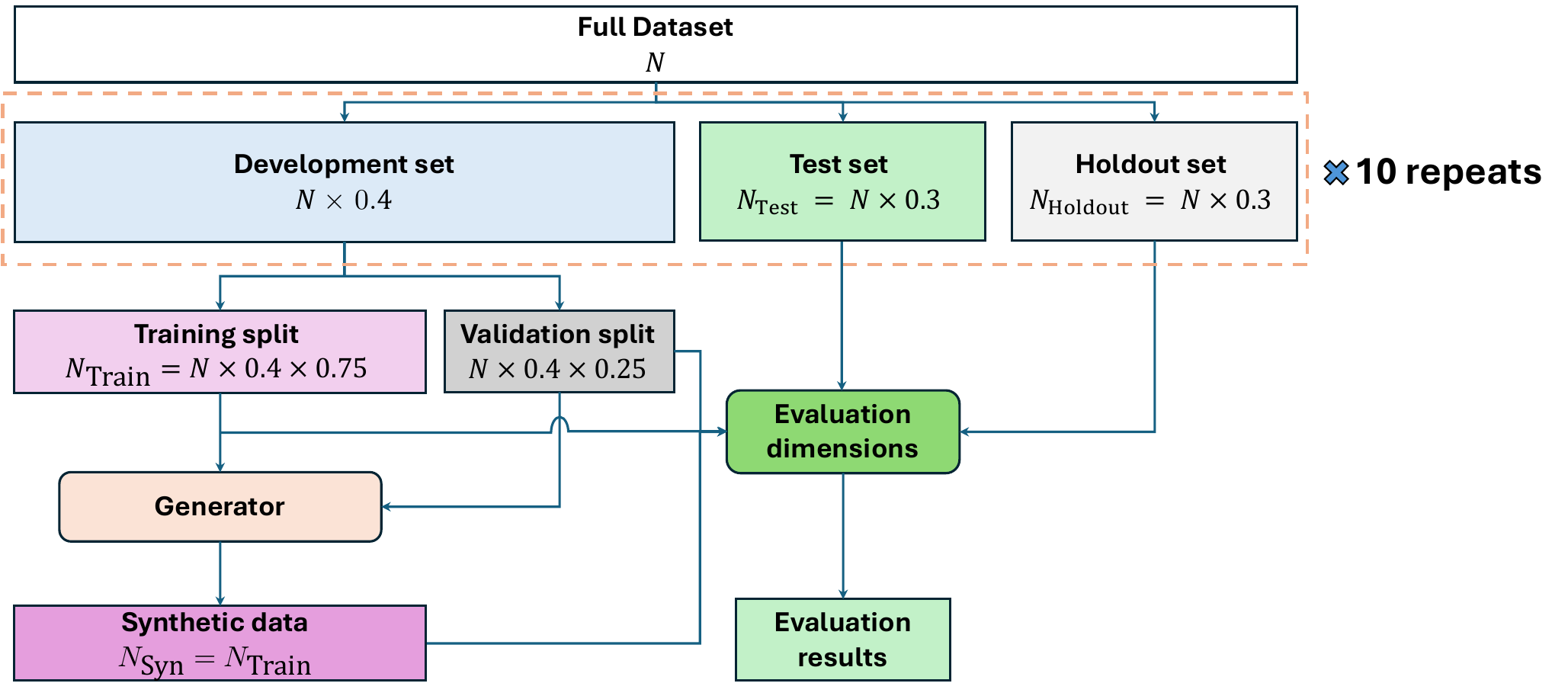}
    \caption{Data splitting strategies for benchmarking tabular data generators.}
    \label{fig:data_splitting}
\end{figure}

\textbf{Detailed data splitting strategies for 45 evaluation datasets.}
During evaluation, for each benchmark dataset of $N$ samples, we perform nested cross-validation with repeated shuffle, as shown in~\Cref{fig:data_splitting}. In each repeat, we first split the dataset into three disjoint subsets: a development set, a test set, and a holdout set, containing 40\%, 30\%, and 30\% of all samples, respectively. We then split the development set into a training split and a validation split, containing 75\% and 25\% of the development samples, respectively. For classification datasets, we perform stratified splitting to preserve the class distribution. We shuffle the dataset to repeat the splitting 10 times, summing up to 10 runs per dataset.
As we further preserve a holdout set with different data splitting strategies, although our benchmark shares some datasets with TabArena and TabStruct, the evaluation results are not directly comparable.

\looseness-1
\textbf{Feature preprocessing for generators.}
Following prior studies~\cite{margeloiu2024tabebm, jiang2026tabstruct}, we perform preprocessing in four steps. 
Firstly, we impute the missing values with the mean value for numerical features and the mode value for categorical features. 
We then compute the required statistics with the training data and then transform the training split. For categorical features, we generally convert them into one-hot encodings.
For \modelname and TabDiff, instead, we perform ordinal encoding by assigning a deterministic integer identifier to each category and cache the corresponding identifier-to-category mapping for inverse transformation.
For numerical features, we perform Z-score normalisation. We compute the mean and standard deviation of each feature in the training data and then transform the training samples to have a mean of zero and a variance of one for each feature.
Next, we apply the same transformation to validation data and train the generators. Once the generators are fitted, we apply the inverse transformation to recover their synthetic tabular data back to the original feature space.
Finally, we conduct evaluations with training data, test data, holdout data, and corresponding synthetic data.

\textbf{Feature preprocessing for downstream predictors.}
When computing global and local utility, the synthetic tabular data is inversely transformed back to the original feature space before being passed to the downstream predictors. In other words, the downstream models receive input data in the original, unprocessed feature space. Following TabStruct~\cite{jiang2026tabstruct}, we use AutoGluon to fit downstream models, thus allowing them to apply their own model-specific preprocessing strategies.

\subsection{Technical Details of Benchmark Generators}
\label[appendix]{appendix:benchmark_generator}

\looseness-1
We include 22 existing tabular data generation methods of ten different categories: 
(i)~two non-parametric methods: SMOTE~\cite{chawla2002smote} and TabSDS~\cite{neto2025tabsds}; 
(ii)~two Variational Autoencoders (VAE) based methods: TVAE~\cite{xu2019modeling} and GOGGLE~\cite{liu2023goggle}; 
(iii)~a Generative Adversarial Networks (GAN) method CTGAN~\cite{xu2019modeling};
(iv)~a normalising flow model Neural Spine Flows (NFlow)~\cite{durkan2019neural};
(v)~a tree-based method Adversarial Random Forests (ARF)~\cite{watson2023adversarial};
(vi)~five diffusion models: TabDDPM~\cite{kotelnikov2023tabddpm}, CDTD~\cite{mueller2025continuous}, TabSyn~\cite{zhang2023mixed}, TabDiff~\cite{shi2025tabdiff}, and CTSyn~\cite{lin2025ctsyn}; 
(vii)~two energy-based models: TabEBM~\cite{margeloiu2024tabebm} and NRGBoost~\cite{bravo2024nrgboost};
(viii)~two tabular-specific autoregressive models: TabNAT~\cite{zhang2025tabnat} and TabularARGN~\cite{tiwald2025tabularargn}.
(ix)~a Large Language Model (LLM) based foundation model GReaT~\cite{borisov2022language}.
(x)~five PFN predictor-based foundation models: TabPFN (i.e., \mbox{Real-TabPFN-2.5}~\cite{grinsztajn2025tabpfn}), TabDPT~\cite{ma2025tabdpt}, Mitra~\cite{zhang2025mitra}, LimiX~\cite{zhang2025limix}, and TabICL (i.e., TabICLv2~\cite{qu2026tabiclv2}).

We note that some of the benchmark generators are also evaluated in TabStruct~\cite{jiang2026tabstruct}. For these methods, we follow the implementations and use the same hyperparameter ranges in TabStruct. For the remaining methods, we define the search ranges for better convergence on the benchmark datasets. The technical details and hyperparameter search space for each method are described below. 

\looseness-1
\textbf{SMOTE} is an interpolation-based oversampling technique~\cite{chawla2002smote}, which generates synthetic samples by interpolating between existing samples. We employ the open-source implementation of SMOTE provided by Imbalanced-learn~\cite{imbalanced-learn}, where the number of nearest neighbours $k$ can be specified. Unless stated otherwise, we use the default setting of $k=5$.

\textbf{TabSDS} is a fully non-parametric and model-free approach for synthetic tabular data generation~\cite{neto2025tabsds}. TabSDS first generates synthetic marginal distributions through interpolated order statistics and then uses sequential joint-probability-preserving data shuffling to recover the dependence structure of the real data. Unlike deep generative models, TabSDS has a small tuning space, primarily controlled by the number of discretisation levels $n_c$, which determines the trade-off between data fidelity and privacy. Unless stated otherwise, we use the default setting of $n_c=200$.

\textbf{TVAE} is a variational autoencoder (VAE) designed for tabular data~\cite{xu2019modeling}. TVAE employs mode-specific normalisation to handle the complex distributions of numerical features. To address the class imbalance problem, TVAE conditions on specific categorical features during generation.

\begin{table}[!ht]
\vspace{-5mm}
\centering
\caption{Hyperparameter search space of TVAE.}
\begin{tabular}{lr}

\toprule

\textbf{Hyperparameter} & \textbf{Range} \\
\midrule
\rowcolor{Gainsboro!60}
encoder\_n\_layers\_hidden & $[1,\,5]$ \\

encoder\_n\_units\_hidden & $[50,\,500]$ \\

\rowcolor{Gainsboro!60}
encoder\_nonlin & $\{\text{relu}, \text{leaky\_relu}, \text{tanh}, \text{elu}\}$ \\

n\_units\_embedding & $[50,\,500]$ \\

\rowcolor{Gainsboro!60}
decoder\_n\_layers\_hidden & $[1,\,5]$ \\

decoder\_n\_units\_hidden & $[50,\,500]$ \\

\rowcolor{Gainsboro!60}
decoder\_nonlin & $\{\text{relu}, \text{leaky\_relu}, \text{tanh}, \text{elu}\}$ \\

\bottomrule
\end{tabular}
\end{table}

\textbf{GOGGLE} is a VAE-based tabular data generator designed to model the dependence relationships between features~\cite{liu2023goggle}. GOGGLE proposes to learn an adjacency matrix to model the dependence relationships between features.
In line with prior studies~\cite{margeloiu2024tabebm, jiang2026tabstruct} and its official codebase~\cite{qian2023synthcity}, we observe that the official implementation of GOGGLE can be unstable and may fail to converge when fitted on large tabular datasets (e.g., more than 10,000 samples).

\begin{table}[!ht]
\centering
\caption{Hyperparameter search space of GOGGLE.}
\begin{tabular}{lr}

\toprule

\textbf{Hyperparameter} & \textbf{Range} \\
\midrule
\rowcolor{Gainsboro!60}
encoder\_dim & $[32,\,128]$ \\
encoder\_l & $[1,\,5]$ \\
\rowcolor{Gainsboro!60}
decoder\_dim & $[32,\,128]$ \\
decoder\_arch & $\{\text{gcn}, \text{het}, \text{sage}\}$ \\
\rowcolor{Gainsboro!60}
n\_iter & $[100,\,500]$ \\
learning\_rate & $[10^{-4},\,5\times10^{-3}]$ (log) \\
\rowcolor{Gainsboro!60}
alpha & $[0.0,\,1.0]$ \\

\bottomrule
\end{tabular}
\end{table}

\textbf{CTGAN} is a conditional generative adversarial network (GAN) designed for tabular data~\cite{xu2019modeling}. CTGAN leverages PacGAN~\cite{lin2018pacgan} framework to mitigate mode collapse. In addition, CTGAN employs the same mode-specific normalisation technique as TVAE.

\begin{table}[!ht]
% \vspace{-5mm}
\centering
\caption{Hyperparameter search space of CTGAN.}
\begin{tabular}{lr}

\toprule

\textbf{Hyperparameter} & \textbf{Range} \\
\midrule
\rowcolor{Gainsboro!60}
generator\_n\_layers\_hidden & $[1,\,4]$ \\
generator\_n\_units\_hidden & $[50,\,150]$ \\
\rowcolor{Gainsboro!60}
generator\_nonlin & $\{\text{relu}, \text{leaky\_relu}, \text{tanh}, \text{elu}\}$ \\
discriminator\_n\_layers\_hidden & $[1,\,4]$ \\
\rowcolor{Gainsboro!60}
discriminator\_n\_units\_hidden & $[50,\,150]$ \\
discriminator\_nonlin & $\{\text{relu}, \text{leaky\_relu}, \text{tanh}, \text{elu}\}$ \\
\rowcolor{Gainsboro!60}
n\_iter & $[100,\,1000]$ \\
discriminator\_n\_iter & $[1,\,5]$ \\
\rowcolor{Gainsboro!60}
lr & $[10^{-4},\,10^{-3}]$ (log) \\
weight\_decay & $[10^{-4},\,10^{-3}]$ (log) \\

\bottomrule
\end{tabular}
% \vspace{-5mm}
\end{table}

\textbf{NFlow} is a normalisation flow model designed for tabular data generation~\cite{durkan2019neural}. NFlow incorporates neural splines as a drop-in replacement for affine or additive transformations in coupling and autoregressive layers, which assists in the modelling of tabular data.

\begin{table}[!ht]
\centering
\caption{Hyperparameter search space of NFlow.}
%
 \\
\rowcolor{Gainsboro!60}
dropout & $[0.0,\,0.2]$ \\
batch\_norm & $\{\text{False}, \text{True}\}$ \\
\rowcolor{Gainsboro!60}
lr & $[2\times10^{-4},\,10^{-3}]$ (log) \\
n\_iter & $[100,\,5000]$ \\

\bottomrule
\end{tabular}
\end{table}

\textbf{ARF} is a tree-based model for tabular data generation~\cite{watson2023adversarial}. ARF employs a recursive adaptation of unsupervised random forests for joint density estimation by iteratively refining synthetic data distributions using adversarial training principles.

\begin{table}[!ht]
\centering
\caption{Hyperparameter search space of ARF.}
\begin{tabular}{lr}

\toprule

\textbf{Hyperparameter} & \textbf{Range} \\
\midrule

\rowcolor{Gainsboro!60}
num\_trees & $\{10,\,20,\,\dotsc,\,100\}$ \\
delta & $\{0,\,2,\,\dotsc,\,50\}$ \\
\rowcolor{Gainsboro!60}
max\_iters & $[1,\,5]$ \\
early\_stop & $\{\text{True}, \text{False}\}$ \\
\rowcolor{Gainsboro!60}
min\_node\_size & $\{2,\,4,\,\dotsc,\,20\}$ \\

\bottomrule
\end{tabular}
\end{table}

\textbf{TabDDPM} is a diffusion-based model for tabular data generation~\cite{kotelnikov2023tabddpm}. TabDDPM introduces two core diffusion processes: (i)~Gaussian noise for numerical features and (ii)~multinomial diffusion with categorical noise for categorical features. TabDDPM directly concatenates numerical and categorical features as the input and output of the denoising function.
We further note that the official implementation of TabDDPM can be unstable and may fail to converge when fitted on tabular data with mixed feature types, which has also been noted in prior studies~\cite{jiang2026tabstruct} and its official codebase~\cite{tabddpm-codebase}.

\begin{table}[!ht]
\centering
\caption{Hyperparameter search space of TabDDPM.}
\begin{tabular}{lr}

\toprule

\textbf{Hyperparameter} & \textbf{Range} \\
\midrule

\rowcolor{Gainsboro!60}
n\_iter & $[10^{3},\,10^{4}]$ \\
lr & $[10^{-5},\,10^{-1}]$ (log) \\
\rowcolor{Gainsboro!60}
weight\_decay & $[10^{-4},\,10^{-3}]$ (log) \\
num\_timesteps & $[10,\,10^{3}]$ \\

\bottomrule
\end{tabular}
\end{table}

\textbf{CDTD} is a continuous diffusion model for mixed-type tabular data~\cite{mueller2025continuous}. CDTD unifies numerical and categorical generation by applying Gaussian diffusion directly to continuous features and learned categorical embeddings. To further account for feature heterogeneity, CDTD introduces adaptive noise schedules, which can be shared across all features by feature type or specified per feature.

\begin{table}[!ht]
% \vspace{-3mm}
\centering
\caption{Hyperparameter search space of CDTD.}
\begin{tabular}{lr}

\toprule

\textbf{Hyperparameter} & \textbf{Range} \\
\midrule

\rowcolor{Gainsboro!60}
timewarp\_type & $\{\texttt{single},\,\texttt{bytype},\,\texttt{all}\}$ \\
generation\_steps & $\{100,\,200,\,500,\,1000\}$ \\
\rowcolor{Gainsboro!60}
timewarp\_weight\_low\_noise & $\{1,\,2,\,3,\,4\}$ \\

\bottomrule
\end{tabular}
% \vspace{-5mm}
\end{table}

\textbf{TabSyn} is a diffusion-based model for tabular data generation~\cite{zhang2023mixed}. It synthesises tabular data by employing a diffusion model within the latent space of a variational autoencoder (VAE). TabSyn supports a wide range of data types by mapping them into a unified representation space and explicitly modelling inter-column dependencies.

\begin{table}[!ht]
\centering
\caption{Hyperparameter search space of TabSyn.}
\begin{tabular}{lr}

\toprule

\textbf{Hyperparameter} & \textbf{Range} \\
\midrule

\rowcolor{Gainsboro!60}
vae.min\_beta & $[10^{-5},\,10^{-4}]$ (log) \\
vae.lambd & $[0.1,\,1.0]$ \\
\rowcolor{Gainsboro!60}
vae.num\_layers & $[1,\,4]$ \\
vae.d\_token & $[1,\,8]$ \\
\rowcolor{Gainsboro!60}
vae.n\_head & $[1,\,4]$ \\
vae.factor & $[1,\,64]$ \\

\bottomrule
\end{tabular}
\end{table}

\textbf{TabDiff} is a diffusion-based model for tabular data generation~\cite{, shi2025tabdiff}. It introduces a joint diffusion framework capable of capturing the mixed-type distributions inherent in tabular data within a single model. In particular, TabDiff utilises a joint continuous-time diffusion process and leverages a transformer architecture to handle both numerical and categorical variables. Consistent with prior work~\cite{jiang2026tabstruct}, we also observe that TabDiff is sensitive to the number of training samples. In particular, when the proportion of the full dataset used for training is reduced, e.g., from 72\% to 30\%, TabDiff exhibits a more pronounced performance degradation than other benchmark generators.

\begin{table}[!ht]
\centering
\caption{Hyperparameter search space of TabDiff.}
\begin{tabular}{lr}

\toprule

\textbf{Hyperparameter} & \textbf{Range} \\
\midrule

\rowcolor{Gainsboro!60}
batch\_size & $\{512,\,1024,\,2048,\,4096,\,8192\}$ \\
c\_lambda & $[0.1,\,10.0]$ \\
\rowcolor{Gainsboro!60}
check\_val\_every & $\{10,\,20,\,30,\,40,\,50\}$ \\
closs\_weight\_schedule & $\{\text{"constant"}, \text{"anneal"}, \text{"linear"}\}$ \\
\rowcolor{Gainsboro!60}
d\_lambda & $[0.1,\,10.0]$ \\
ema\_decay & $[0.9,\,0.9999]$ \\

\bottomrule
\end{tabular}
\end{table}

\looseness-1
\textbf{CTSyn} is a diffusion-based generative foundation model for tabular data~\cite{lin2025ctsyn}. CTSyn first maps heterogeneous tables into a unified latent space using a variational autoencoder, with metadata embedded by a pretrained LLM.
Since the open-source weights of CTSyn are unavailable, we strictly follow its official pretraining code~\cite{ctsyn-codebase} and pretrain CTSyn on the 86 datasets specified in its official report~\cite{lin2025ctsyn}. However, we find that pretraining only on these 86 datasets yields poor performance on the 45 benchmark datasets. We therefore further continue pretraining CTSyn on the 43 pretraining datasets used for \modelname (\Cref{tab:pretraining_dataset}). In other words, CTSyn is given an advantage over \modelname, as it is exposed to 86 more real-world datasets during pretraining.
Moreover, as noted in its official report~\cite{lin2025ctsyn}, CTSyn is highly sensitive to the quality of metadata and the capability of the employed LLM. We observe the same issue in our experiments: using the officially recommended GPT-4o causes failure to converge on many of the 45 benchmark datasets. We therefore switch to the more advanced GPT-5.1 throughout the pretraining and evaluation of CTSyn, which can mitigate the convergence issues.

\begin{table}[!ht]
% \vspace{-3mm}
\centering
\caption{Hyperparameter search space of CTSyn.}
\begin{tabular}{lr}

\toprule

\textbf{Hyperparameter} & \textbf{Range} \\
\midrule

\rowcolor{Gainsboro!60}
num\_latent ($\ell$) & $\{16\}$ \\
aggregated\_dim ($M_{\text{agg}}$) & $\{64\}$ \\
\rowcolor{Gainsboro!60}
diffusion\_learning\_rate & $\{10^{-4}\}$ \\
diffusion\_pretrain\_steps & $\{100{,}000,\ 200{,}000,\ 300{,}000\}$ \\
\rowcolor{Gainsboro!60}
sampling\_steps & $\{100, 200, 300, 400\}$ \\
conditioning\_dropout & $\{0.1\}$ \\

\bottomrule
\end{tabular}
% \vspace{-5mm}
\end{table}

\textbf{TabEBM} is an energy-based model for tabular data generation~\cite{margeloiu2024tabebm}. It transforms a pretrained tabular predictor into a set of class-specific generators. While the original paper only provides TabEBM implementation for classification tasks, we adopt the version in TabStruct~\cite{jiang2026tabstruct} to extend its applicability in TabStruct to regression tasks by treating all reference samples as a single class, and then performing sampling over the energy landscape. Moreover, we update its backbone to \mbox{Real-TabPFN-2.5} for a fair comparison with \modelname and other foundation models.

\begin{table}[!ht]
\centering
\caption{Hyperparameter search space of TabEBM.}
\begin{tabular}{lr}

\toprule

\textbf{Hyperparameter} & \textbf{Range} \\
\midrule

\rowcolor{Gainsboro!60}
starting\_point\_noise\_std & $[10^{-4},\,10^{-1}]$ (log) \\
sgld\_step\_size & $[10^{-3},\,10^{-1}]$ (log) \\
\rowcolor{Gainsboro!60}
sgld\_noise\_std & $[10^{-4},\,10^{-1}]$ (log) \\
sgld\_steps & $\{50,\,100,\,200,\,500\}$ \\

\bottomrule
\end{tabular}
\end{table}

\textbf{NRGBoost} is an energy-based model for tabular data generation~\cite{bravo2024nrgboost}. It is trained by maximising a local second-order approximation to the log-likelihood at each stage of the boosting process. NRGBoost is shown to offer generally good discriminative performance and competitive sampling performance compared to more specialised alternatives.

\begin{table}[!ht]
\centering
\caption{Hyperparameter search space of NRGBoost.}
\begin{tabular}{lr}

\toprule

\textbf{Hyperparameter} & \textbf{Range} \\
\midrule

\rowcolor{Gainsboro!60}
num\_trees & $\{1,\,5,\,10,\,20,\,50\}$ \\
shrinkage & $[0.01,\,0.3]$ \\
\rowcolor{Gainsboro!60}
max\_leaves & $\{32,\,64,\,128,\,256,\,512\}$ \\
max\_ratio\_in\_leaf & $[1,\,5]$ \\
\rowcolor{Gainsboro!60}
num\_model\_samples & $\{10{,}000,\,40{,}000,\,80{,}000,\,160{,}000\}$ \\

\bottomrule
\end{tabular}
\end{table}

\textbf{TabNAT} is a continuous-discrete joint generative framework for tabular data~\cite{zhang2025tabnat}. It combines masked generative modelling with a bidirectional transformer to learn conditional distributions under arbitrary column orders. This design enables TabNAT to model heterogeneous tabular data within a unified architecture and supports both unconditional generation and flexible conditional generation.

\begin{table}[!ht]
% \vspace{-3mm}
\centering
\caption{Hyperparameter search space of TabNAT.}
\begin{tabular}{lr}

\toprule

\textbf{Hyperparameter} & \textbf{Range} \\
\midrule

\rowcolor{Gainsboro!60}
epochs & $\{3000,\,5000,\,6000\}$ \\
batch\_size & $\{512,\,1024,\,2048\}$ \\
\rowcolor{Gainsboro!60}
learning\_rate & $[10^{-4},\,10^{-2}]$ (log) \\
weight\_decay & $[10^{-7},\,10^{-5}]$ (log) \\
\rowcolor{Gainsboro!60}
depth & $\{2,\,4,\,6,\,8\}$ \\

\bottomrule
\end{tabular}
\end{table}

\textbf{TabularARGN} is an autoregressive framework for tabular data generation~\cite{tiwald2025tabularargn}. It first converts heterogeneous features into categorical sub-columns, and then learns their joint distribution through shared embedding, permutation-masking, regressor, and predictor modules. During training, TabularARGN randomly permutes the column order and minimises the summed categorical cross-entropy with teacher forcing, enabling flexible conditional generation over arbitrary observed feature subsets.

\begin{table}[!ht]
% \vspace{-3mm}
\centering
\caption{Hyperparameter search space of TabularARGN.}
\begin{tabular}{lr}

\toprule

\textbf{Hyperparameter} & \textbf{Range} \\
\midrule

\rowcolor{Gainsboro!60}
model\_size & $\{\texttt{S},\,\texttt{M},\,\texttt{L}\}$ \\
max\_training\_epochs & $\{100,\,300,\,500\}$ \\

\bottomrule
\end{tabular}
\end{table}

\textbf{GReaT} leverages large language models (LLMs) to generate synthetic tabular data~\cite{borisov2022language}. GReaT converts each sample into a sentence and finetunes the language model to capture the sentence-level distributions. Additionally, GReaT shuffles the order of features to mitigate the permutation variance in sentence-level distributions.

\begin{table}[!ht]
% \vspace{-3mm}
\centering
\caption{Hyperparameter search space of GReaT.}
\begin{tabular}{lr}

\toprule

\textbf{Hyperparameter} & \textbf{Range} \\
\midrule

\rowcolor{Gainsboro!60}
n\_iter & $\{100,\,300,\,500,\,1000\}$ \\
learning\_rate & $[10^{-4},\,10^{-2}]$ (log) \\
\rowcolor{Gainsboro!60}
weight\_decay & $[10^{-5},\,10^{-2}]$ (log) \\

\bottomrule
\end{tabular}
% \vspace{-5mm}
\end{table}

\textbf{TabPFN} is originally designed as a tabular foundation predictor, and we repurpose it for autoregressive tabular data generation~\cite{grinsztajn2025tabpfn}. Following the official implementation~\cite{prior2026tabpfnext}, we instantiate TabPFN classifiers for categorical features and TabPFN regressors for numerical features, and use them as feature-wise conditional density estimators. To generate synthetic samples, the model initialises an empty table with missing entries and fills the features sequentially. 
Following the default column order of each dataset, we generate features sequentially. At each step, the foundation predictor produces a conditional predictive distribution given the previously generated features, from which the next feature is sampled.
For the first column, no preceding feature is available as context. We therefore sample categorical values from their empirical distribution. For numerical features, we add random noise to the original values.
When a categorical feature has a cardinality exceeding the fixed prediction-head capacity, we apply the hierarchical classification strategy~\cite{qu2026tabiclv2}.
This procedure converts TabPFN in-context predictive capability into an autoregressive generator without updating the pretrained backbone parameters.

\begin{table}[!ht]
% \vspace{-3mm}
\centering
\caption{Hyperparameter search space of TabPFN.}
\begin{tabular}{lr}

\toprule

\textbf{Hyperparameter} & \textbf{Range} \\
\midrule

\rowcolor{Gainsboro!60}
feature\_permutation & $\{\texttt{random}\}$ \\
n\_permutations & $\{1,\,3,\,5,\,10\}$ \\
\rowcolor{Gainsboro!60}
temperature & $\{0.1,\,0.3,\,0.5,\,0.7,\,1.0\}$ \\
categorical\_feature\_inference & $\{\texttt{auto}\}$ \\
\rowcolor{Gainsboro!60}
fast\_mode & $\{\texttt{False}\}$ \\

\bottomrule
\end{tabular}
% \vspace{-5mm}
\end{table}

\textbf{TabDPT} is originally designed as an in-context tabular foundation predictor~\cite{ma2025tabdpt}, and we repurpose it for autoregressive tabular data generation in the same procedure as TabPFN. TabDPT uses a row-based transformer encoder with retrieval-based context selection, and performs inference by conditioning each query row on retrieved context rows without dataset-specific weight updates.

\begin{table}[!ht]
% \vspace{-3mm}
\centering
\caption{Hyperparameter search space of TabDPT.}
\begin{tabular}{lr}

\toprule

\textbf{Hyperparameter} & \textbf{Range} \\
\midrule

\rowcolor{Gainsboro!60}
feature\_permutation & $\{\texttt{random}\}$ \\
n\_ensembles & $\{1,\,4,\,8,\,16\}$ \\
\rowcolor{Gainsboro!60}
temperature & $\{0.5,\,0.8,\,1.0\}$ \\
context\_size & $\{128,\,256,\,512,\,1024,\,2048\}$ \\
\rowcolor{Gainsboro!60}
permute\_classes & $\{\texttt{True},\,\texttt{False}\}$ \\

\bottomrule
\end{tabular}
% \vspace{-5mm}
\end{table}

\textbf{Mitra} is originally proposed as an in-context tabular foundation predictor~\cite{zhang2025mitra}, and we repurpose it for autoregressive tabular data generation, similar to TabPFN. MITRA is pretrained on a curated mixture of synthetic priors, including structural causal models and tree-based priors, to improve generalisation across classification and regression tasks.

\begin{table}[!ht]
% \vspace{-3mm}
\centering
\caption{Hyperparameter search space of Mitra.}
\begin{tabular}{lr}

\toprule

\textbf{Hyperparameter} & \textbf{Range} \\
\midrule

\rowcolor{Gainsboro!60}
n\_estimators & $\{1,\,2,\,4,\,8\}$ \\
max\_classes & $\{10\}$ \\
\rowcolor{Gainsboro!60}
device & $\{\texttt{auto}\}$ \\

\bottomrule
\end{tabular}
% \vspace{-5mm}
\end{table}

\textbf{LimiX} is originally pretrained with context-conditional masked modelling~\cite{zhang2025limix}, and we repurpose it for autoregressive tabular data generation. LimiX is designed to learn a conditional distribution through masked prediction and can generate samples by iteratively masking and refilling subsets of features. As noted in \mbox{Section~7.7} of its official technical report~\cite{zhang2025limix}, LimiX performs generation iteratively, like other in-context tabular foundation predictors such as TabPFN.

\begin{table}[!ht]
% \vspace{-3mm}
\centering
\caption{Hyperparameter search space of LimiX.}
\begin{tabular}{lr}

\toprule

\textbf{Hyperparameter} & \textbf{Range} \\
\midrule

\rowcolor{Gainsboro!60}
inference\_config & $\{\texttt{default},\,\texttt{retrieval-ensemble}\}$ \\
mix\_precision & $\{\texttt{True},\,\texttt{False}\}$ \\
\rowcolor{Gainsboro!60}
softmax\_temperature & $\{0.7,\,0.9,\,1.0\}$ \\
outlier\_remove\_std & $\{8,\,12,\,16\}$ \\

\bottomrule
\end{tabular}
% \vspace{-5mm}
\end{table}

\textbf{TabICL} is proposed as an in-context tabular foundation predictor~\cite{qu2026tabiclv2}, and we repurpose it for autoregressive tabular data generation, similar to TabPFN. TabICLv2 performs prediction by conditioning query samples on labelled in-context examples through a scalable Transformer architecture with repeated feature grouping, target-aware embedding, and query-aware scalable softmax.

\begin{table}[!ht]
% \vspace{-3mm}
\centering
\caption{Hyperparameter search space of TabICL.}
\begin{tabular}{lr}

\toprule

\textbf{Hyperparameter} & \textbf{Range} \\
\midrule

\rowcolor{Gainsboro!60}
n\_estimators & $\{1,\,4,\,8,\,16,\,32\}$ \\
softmax\_temperature & $\{0.7,\,0.9,\,1.0\}$ \\
\rowcolor{Gainsboro!60}
outlier\_threshold & $\{3.0,\,4.0,\,6.0\}$ \\
average\_logits & $\{\texttt{True},\,\texttt{False}\}$ \\
\rowcolor{Gainsboro!60}
support\_many\_classes & $\{\texttt{True}\}$ \\
use\_amp & $\{\texttt{auto}\}$ \\

\bottomrule
\end{tabular}
% \vspace{-5mm}
\end{table}

\subsection{Experimental Setup}
\label[appendix]{appendix:exp_setup}

For all benchmark generators, we tune their hyperparameters using Optuna~\cite{akiba2019optuna} according to their average optimisation objective across the 10 repeated runs. Specifically, we tune parametrised generators to minimise their validation loss. For remaining in-context tabular foundation predictors, whose optimisation objectives are not designed for generation, we instead tune their inference hyperparameters to maximise global utility. Each generator is given at most two hours to complete a single repeat.
Unless otherwise stated, reported results are averaged over 10 repeats. Specifically, we use the average distance to the minimum (ADTM) metric with affine renormalisation between the best-performing and worst-performing models~\cite{jiang2026tabstruct, grinsztajn2022tree, mcelfresh2024neural, hollmann2025accurate, margeloiu2024tabebm, jiang2024protogate}. We note that some generators may fail to converge or may generate unexpected values that cause metric computation to fail. In such cases, we impute their metric values for each repeat using the mean value of the other methods, following prior benchmarking practice~\cite{jiang2026tabstruct, margeloiu2024tabebm, jiang2024protogate}.

\subsection{Software Implementations}

\textbf{Implementation of benchmark generators.}
We implemented SMOTE with Imbalanced-learn~\cite{imbalanced-learn}, an open-source Python library for imbalanced datasets with an MIT license. For TabSyn, TabEBM, TabularARGN, TabDPT, and LimiX, we used their open-source implementations with an Apache-2.0 license. For CDTD, TabDiff, and NRGBoost, we used their open-source implementations with an MIT license. For TabSDS, TabNAT, TabPFN, Mitra, and TabICL, we used their open-source implementations with their customised licenses.
For other benchmark generators, we used their open-source implementations in TabStruct~\cite{jiang2026tabstruct}, an open-source toolbox for tabular data generation with an Apache-2.0 license. 

\textbf{Implementation of data processing and evaluation.}
All data handling, including data loading and preprocessing, was performed with TabCamel~\cite{jiang2025tabcamel}, an open-source Python library for tabular data management. All data-quality evaluation was performed with TabEval~\cite{jiang2025tabeval}, a comprehensive open-source Python framework for evaluating tabular data.

\textbf{Implementation of result analysis and visualisation.}
All numerical plots and graphics have been generated using Matplotlib 3.7~\cite{matplotlib}, a Python-based plotting library with a BSD license. The icons for ``frozen'' and ``trainable'' in~\Cref{fig:framework} are from \url{https://icons8.com/}.

\textbf{Implementation of benchmark pipeline.}
Following prior studies~\cite{jiang2024protogate, jiang2026tabstruct}, we ensure the consistency and reproducibility of experimental results by implementing a uniform pipeline using PyTorch Lightning~\cite{pytorch-lightning}, an open-source library under an Apache-2.0 license. We further fixed the random seeds for data loading and evaluation throughout the training and evaluation process. This ensured that \modelname and all benchmark models were trained and evaluated on the same set of samples. The experimental environment settings, including library dependencies, are specified in the open-source library for reference and reproduction purposes.

\clearpage
\section{Extended Analysis and Discussion}

\subsection{Extended Analysis on Generation Quality}
\label[appendix]{appendix:extended_analysis_quality}

\textbf{\modelname performs strongly on evaluations sensitive to dependency structures.}
To provide complementary evidence for our structural motivation, we evaluate \modelname on datasets with known structural causal models (SCMs).
We select six SCM datasets from TabStruct~\cite{jiang2026tabstruct} with expert-validated, known SCMs with 7-64 features: Hailfinder, Insurance, Sangiovese, Healthcare, MAGIC-IRRI, and MEHRA. We strictly follow Algorithm~2 in Appendix~D.1.1 of TabStruct and use its official codebase to construct datasets with ground-truth SCMs.
We include four causal generators with explicit causal structure learning mechanisms: BN~\cite{qian2023synthcity}, VACA~\cite{sanchez2022vaca}, CAREFL~\cite{khemakhem2021causal}, and DCM~\cite{chao2023modeling}. We first use the performant AutoCD~\cite{chan2024autocd} suggested by prior work~\cite{jiang2026tabstruct} to discover the causal graph from the training data, and then fit the corresponding generative models to the estimated structure.
All other experimental settings, including data preprocessing, model training, and metric aggregation, follow those in~\Cref{sec:experiment}.

As shown in~\Cref{tab:scm_data_eval}, \modelname achieves both the highest global CI and global utility. This result is consistent with the motivation that \modelname can capture and utilise dependency structures for synthetic data generation, suggesting that its pretrained representations encode structural information relevant to the underlying causal data-generating process. The strong performance of \modelname further indicates that pretrained representations with structural information can be highly effective, and may even outperform explicit causal modelling mechanisms (i.e., causal generators) in preserving such inter-feature dependencies.
Consistent with prior work~\cite{jiang2026tabstruct}, we also observe the generator rankings are similar under global CI and global utility. This alignment further supports our use of global utility as an empirical measure for structural fidelity.

\begin{table}[htbp]
\centering
\caption{\textbf{Benchmark results on six SCM datasets with known ground-truth SCMs.} We report the normalised global utility and global CI, and \textbf{bold} the highest score for each metric. \modelname achieves competitive performance in preserving structural relationships induced by the ground-truth SCMs against causal generators.}
\label{tab:scm_data_eval}
\begin{tabular}{lcc}
\toprule
Generator & Global utility $\uparrow$ & Global CI $\uparrow$ \\
\midrule
\multicolumn{3}{l}{\textbf{\textit{Causal generators}}} \\
BN & 0.73$_{\pm\text{0.04}}$ & 0.68$_{\pm\text{0.05}}$ \\
VACA & 0.59$_{\pm\text{0.03}}$ & 0.54$_{\pm\text{0.04}}$ \\
CAREFL & 0.64$_{\pm\text{0.04}}$ & 0.57$_{\pm\text{0.03}}$ \\
DCM & 0.68$_{\pm\text{0.03}}$ & 0.62$_{\pm\text{0.04}}$ \\
\midrule
\multicolumn{3}{l}{\textbf{\textit{Other generators}}} \\
SMOTE & 0.54$_{\pm\text{0.03}}$ & 0.42$_{\pm\text{0.03}}$ \\
TabSyn & 0.80$_{\pm\text{0.02}}$ & 0.74$_{\pm\text{0.03}}$ \\
TabPFN & 0.48$_{\pm\text{0.05}}$ & 0.34$_{\pm\text{0.06}}$ \\
\midrule
\textbf{\modelname (Ours)} & \textbf{0.87}$_{\pm\text{0.02}}$ & \textbf{0.82}$_{\pm\text{0.02}}$ \\
\bottomrule
\end{tabular}
\end{table}

\textbf{Generators with explicit causal structure learning may struggle on real-world data.}
We further conduct experiments on real-world datasets to directly compare causal generators with methods that do not explicitly model causal structure. We select ten datasets from \Cref{tab:benchmark_dataset_classification} and \Cref{tab:benchmark_dataset_regression}: BankChurn, Customer, Diabetes, Fitness, Give, HR, Maternal, Phoneme, Airfoil, and Concrete.
\Cref{tab:scm_generator_eval} shows that \modelname substantially outperforms the causal generators on global utility while also achieving strong performance across most other metrics. Consistent with prior work~\citep{jiang2026tabstruct}, we observe that generators with explicit causal structure learning do not achieve strong structural fidelity on these real-world datasets, despite their inductive bias towards modelling structural relationships in tabular data.
One possible explanation is error propagation in the causal-generation pipeline: when the estimated causal graph is inaccurate, the resulting misspecified structure can constrain the subsequent generative model and degrade synthetic data quality~\cite{jiang2026tabstruct, tu2024causality}.
The results further underscore the limitations of existing causal discovery methods in recovering precise causal structures from real-world data~\cite{kaddour2022causal, tu2024causality, glymour2019review, nastl2024causal}, justifying the evaluation of structural fidelity at the Markov-equivalence level with an empirically effective metric (i.e., global utility).

\begin{table}[htbp]
\centering
\caption{\textbf{Benchmark results for \modelname and causal generators on ten real-world datasets.} We report the normalised mean $\pm$ std and \textbf{bold} the best result for each metric. \modelname achieves the best performance on seven of the eight metrics, including structural fidelity of synthetic data, as measured by global utility.}
\label{tab:scm_generator_eval}
\resizebox{\textwidth}{!}{
\begin{tabular}{l|rrrr|rr|r|r}
\toprule
\multirow{2}{*}{\textbf{Generator}} & \multicolumn{4}{c|}{\textbf{Density   Estimation}}                             & \multicolumn{2}{c|}{\textbf{Privacy Preservation}} & \multicolumn{1}{c|}{\textbf{ML Efficacy}}                              & \multicolumn{1}{c}{\textbf{Structural Fidelity}} \\

& Shape $\uparrow$ & Trend $\uparrow$ & $\alpha$-precision $\uparrow$ & $\beta$-recall $\uparrow$ & Authenticity $\uparrow$ & DCR Score $\uparrow$ & Local utility $\uparrow$ & Global utility $\uparrow$  \\
\midrule
BN & 0.51$_{\pm\text{0.04}}$ & 0.44$_{\pm\text{0.05}}$ & 0.46$_{\pm\text{0.04}}$ & 0.21$_{\pm\text{0.03}}$ & 0.39$_{\pm\text{0.04}}$ & 0.43$_{\pm\text{0.05}}$ & 0.40$_{\pm\text{0.04}}$ & 0.35$_{\pm\text{0.05}}$ \\
VACA & 0.55$_{\pm\text{0.03}}$ & 0.47$_{\pm\text{0.04}}$ & 0.50$_{\pm\text{0.03}}$ & 0.25$_{\pm\text{0.04}}$ & 0.43$_{\pm\text{0.03}}$ & 0.47$_{\pm\text{0.04}}$ & 0.44$_{\pm\text{0.03}}$ & 0.38$_{\pm\text{0.04}}$ \\
CAREFL & 0.58$_{\pm\text{0.04}}$ & 0.51$_{\pm\text{0.03}}$ & 0.54$_{\pm\text{0.04}}$ & 0.29$_{\pm\text{0.03}}$ & 0.46$_{\pm\text{0.04}}$ & 0.51$_{\pm\text{0.03}}$ & 0.47$_{\pm\text{0.04}}$ & 0.41$_{\pm\text{0.03}}$ \\
DCM & 0.62$_{\pm\text{0.03}}$ & 0.55$_{\pm\text{0.04}}$ & 0.58$_{\pm\text{0.03}}$ & 0.34$_{\pm\text{0.04}}$ & 0.51$_{\pm\text{0.03}}$ & \textbf{0.67}$_{\pm\text{0.03}}$ & 0.50$_{\pm\text{0.03}}$ & 0.45$_{\pm\text{0.04}}$ \\
\midrule
\textbf{\modelname (Ours)} & \textbf{0.82}$_{\pm\text{0.02}}$ & \textbf{0.78}$_{\pm\text{0.02}}$ & \textbf{0.79}$_{\pm\text{0.03}}$ & \textbf{0.54}$_{\pm\text{0.03}}$ & \textbf{0.61}$_{\pm\text{0.02}}$ & 0.56$_{\pm\text{0.03}}$ & \textbf{0.76}$_{\pm\text{0.02}}$ & \textbf{0.71}$_{\pm\text{0.02}}$ \\
\bottomrule
\end{tabular}
}
\end{table}

\textbf{\modelname remains competitive on high-dimensional datasets.}
We select five high-dimensional datasets with 2,000-7,129 features~\cite{habib2026gotabpfn}: COL, LNG, AML, PRS, and TOX. All other settings (data processing, metric aggregation, etc.) are the same as in~\Cref{sec:experiment}.
As shown in~\Cref{tab:high_dimensional}, \modelname achieves competitive performance on these high-dimensional datasets, achieving the best shape, trend, and global utility among the benchmark methods while attaining comparable local utility. In particular, its strongest global utility suggests that the structure-aware motivation of \modelname remains effective as the feature dimensionality increases.

\begin{table}[htbp]
    \centering
    \caption{\textbf{Comparison of \modelname against six benchmark generators across five high-dimensional tabular datasets.} We report the mean $\pm$ std of the normalised metric values and \textbf{bold} the best result for each metric. \modelname generally outperforms benchmark generators in high-dimensional tasks, showing the generalisability of its model design.}
    \label{tab:high_dimensional}
    \begin{tabular}{lrrrr}
        \toprule
        Generator & Shape $\uparrow$ & Trend $\uparrow$ & Local utility $\uparrow$ & Global utility $\uparrow$ \\
        \midrule
        SMOTE   & 0.43$_{\pm\text{0.04}}$ & 0.39$_{\pm\text{0.05}}$ & 0.49$_{\pm\text{0.03}}$ & 0.44$_{\pm\text{0.05}}$ \\
        TabSyn  & 0.61$_{\pm\text{0.03}}$ & 0.58$_{\pm\text{0.04}}$ & \textbf{0.64}$_{\pm\text{0.02}}$ & 0.64$_{\pm\text{0.03}}$ \\
        TabDiff & 0.58$_{\pm\text{0.04}}$ & 0.56$_{\pm\text{0.03}}$ & 0.61$_{\pm\text{0.04}}$ & 0.59$_{\pm\text{0.04}}$ \\
        TabPFN  & 0.37$_{\pm\text{0.06}}$ & 0.34$_{\pm\text{0.05}}$ & 0.41$_{\pm\text{0.04}}$ & 0.36$_{\pm\text{0.05}}$ \\
        TabICL  & 0.35$_{\pm\text{0.05}}$ & 0.33$_{\pm\text{0.06}}$ & 0.39$_{\pm\text{0.05}}$ & 0.34$_{\pm\text{0.04}}$ \\
        TabEBM  & 0.55$_{\pm\text{0.03}}$ & 0.52$_{\pm\text{0.04}}$ & 0.58$_{\pm\text{0.03}}$ & 0.56$_{\pm\text{0.04}}$ \\
        \midrule
        \textbf{\modelname (Ours)}
        & \textbf{0.64}$_{\pm\text{0.02}}$
        & \textbf{0.62}$_{\pm\text{0.03}}$
        & 0.63$_{\pm\text{0.02}}$
        & \textbf{0.68}$_{\pm\text{0.02}}$ \\
        \bottomrule
    \end{tabular}
\end{table}

\subsection{Extended Analysis on Ablation Impacts}
\label[appendix]{appendix:extended_analysis_ablation}

\textbf{The choice of feature encoder is not necessarily determined by predictive performance.}
\Cref{tab:ablation_encoder} shows that \modelname with TabPFN achieves the best average rank and ranks among the top three across all reported metrics. Importantly, stronger predictive performance does not necessarily imply a better encoder for generation. In particular, TabICL outperforms TabPFN in predictive performance~\cite{qu2026tabiclv2}, but the row-wise interaction of TabICL collapses feature embeddings for a sample into a single vector. This can substantially break per-feature semantics, making TabICL less effective for encoding comprehensive structural information. The advantage of TabPFN therefore suggests that preserving per-feature representations provides a stronger inductive bias for capturing transferable dependency structures across heterogeneous tabular datasets, and that predictive performance alone is insufficient to determine whether an encoder is suitable for generation.
In particular, a broadly useful pretrained representation may benefit from preserving underlying structural relationships across features. In other words, ``structural information'' and ``generically strong pretrained representations'' can be complementary in generative modelling, rather than mutually exclusive.

Moreover, our goal is not to decompose TabFORGE performance into independent percentage contributions, since its components are designed to be functionally complementary. The encoder comparison (\Cref{tab:ablation_encoder}) specifically examines whether the choice of pretrained representation matters within a fixed TabFORGE generative framework. 
We further note that isolating the contribution of the structural prior induced by TabPFN pretraining from other properties of its learned representations would require a carefully matched pretrained encoder that differs primarily in the use of such structural priors during pretraining. In particular, a meaningful ablation would require an alternative encoder with comparable representational capability and downstream predictive performance to TabPFN, but pretrained without access to the structural prior data. To the best of our knowledge, no such matched model is currently available. Moreover, constructing this controlled variant ourselves is not feasible because the pretraining prior used by TabPFN is not publicly available. We therefore leave a fully controlled disentanglement of the effects of structural priors from other properties of TabPFN representations to future work.
That being said, the intermediate-layer ablation (\Cref{fig:ablation_arch}), alternative-encoder comparison (\Cref{tab:ablation_encoder}), prior representation analyses~\cite{swelam2025does}, and experiments on datasets with known SCMs (\Cref{tab:scm_data_eval}) provide complementary evidence consistent with the hypothesis that structurally informative pretrained representations can generally benefit tabular data generation.

\begin{table}[!thbp]
\centering
\caption{\textbf{Comparison of different feature encoders in \modelname on 45 real-world datasets.} We report the normalised mean $\pm$ std metric values and average rank. A higher rank indicates better performance. Since the normalisation is performed only across the five variants of \modelname considered here, the reported numbers differ from those in~\Cref{fig:utility} and~\Cref{fig:overfitting}. We highlight the {\color[HTML]{008080} \textbf{First}}, {\color[HTML]{7030A0} \textbf{Second}} and {\color[HTML]{C65911} \textbf{Third}} best performances. The adopted TabPFN-based configuration consistently ranks Top-3 across all metrics and achieves the best overall performance.}
\label{tab:ablation_encoder}
\resizebox{\textwidth}{!}{
\begin{tabular}{l|rrrr|rr|r|r|r}

\toprule

\multirow{2}{*}{\textbf{Feature encoder type}} & \multicolumn{4}{c|}{\textbf{Density   Estimation}}                             & \multicolumn{2}{c|}{\textbf{Privacy Preservation}} & \multicolumn{1}{c|}{\textbf{ML Efficacy}}                              & \multicolumn{1}{c|}{\textbf{Structural Fidelity}} & \multirow{2}{*}{\textbf{Average Rank}} \\

& Shape $\uparrow$ & Trend $\uparrow$ & $\alpha$-precision $\uparrow$ & $\beta$-recall $\uparrow$ & Authenticity $\uparrow$ & DCR Score $\uparrow$ & Local utility $\uparrow$ & Global utility $\uparrow$  \\
    
\midrule

w/ TabDPT & 0.00$_{\pm\text{0.00}}$ & 0.01$_{\pm\text{0.01}}$ & 0.10$_{\pm\text{0.02}}$ & 0.52$_{\pm\text{0.04}}$ & 0.00$_{\pm\text{0.00}}$ & 0.00$_{\pm\text{0.00}}$ & 0.05$_{\pm\text{0.02}}$ & 0.26$_{\pm\text{0.04}}$ & 4.75$_{\pm\text{0.46}}$ \\

w/ Mitra & {\color[HTML]{C65911} \textbf{0.51}}$_{\pm\text{0.02}}$ & {\color[HTML]{7030A0} \textbf{0.49}}$_{\pm\text{0.07}}$ & {\color[HTML]{7030A0} \textbf{0.51}}$_{\pm\text{0.08}}$ & {\color[HTML]{7030A0} \textbf{0.57}}$_{\pm\text{0.08}}$ & {\color[HTML]{C65911} \textbf{0.51}}$_{\pm\text{0.03}}$ & {\color[HTML]{C65911} \textbf{0.30}}$_{\pm\text{0.03}}$ & {\color[HTML]{7030A0} \textbf{0.53}}$_{\pm\text{0.07}}$ & {\color[HTML]{7030A0} \textbf{0.63}}$_{\pm\text{0.08}}$ & {\color[HTML]{7030A0} \textbf{2.38}}$_{\pm\text{0.52}}$ \\

w/ LimiX & 0.36$_{\pm\text{0.02}}$ & 0.36$_{\pm\text{0.06}}$ & {\color[HTML]{C65911} \textbf{0.40}}$_{\pm\text{0.07}}$ & {\color[HTML]{C65911} \textbf{0.52}}$_{\pm\text{0.07}}$ & 0.44$_{\pm\text{0.03}}$ & 0.26$_{\pm\text{0.03}}$ & {\color[HTML]{C65911} \textbf{0.44}}$_{\pm\text{0.06}}$ & {\color[HTML]{C65911} \textbf{0.56}}$_{\pm\text{0.08}}$ & 3.50$_{\pm\text{0.53}}$ \\
\bottomrule

\end{tabular}
}
\end{table}

\subsection{Extended Analysis on Practicability}
\label[appendix]{appendix:extended_analysis_practicability}

\textbf{\modelname incurs a low fitting cost on unseen datasets.}
\Cref{fig:computation_time} (Left) shows that \modelname is substantially cheaper to fit than most benchmark methods. Specifically, the total fitting time of \modelname is only around 9.71\% of that required to train a strong dataset-specific generator (i.e., TabSyn) from scratch. This suggests that \modelname offers a practical trade-off between generation quality and computational efficiency. The efficiency gain stems from the fact that most representational and generative capacity is acquired during pretraining, leaving only lightweight finetuning for unseen datasets. This further demonstrates the practical value of leveraging knowledge from diverse tabular datasets through pretraining.

\textbf{\modelname supports efficient and flexible synthetic data generation.}
\Cref{fig:computation_time} (Right) shows that the generation cost of \modelname is only around 1.66\% of that required by the fastest existing in-context tabular foundation model (i.e., TabEBM). This is because \modelname is designed to generate synthetic data by denoising and decoding the denoised latent embeddings, without invoking the structure-aware feature encoder.
We also note that the considered tabular foundation predictors are all equipped with fixed prediction heads, and can therefore generate categorical features with at most 10 classes. To apply them to all considered datasets, we perform hierarchical classification, as suggested by prior studies~\cite{qu2026tabiclv2, qu2025tabicl}. As a result, the generation time of these foundation predictors increases substantially. In contrast, \modelname employs a lightweight detokeniser that can be efficiently applied to diverse feature spaces. 

\subsection{Future Work}
\label[appendix]{appendix:future_work}

\textbf{Scale up the pretraining scale, including data and model size.}
The current pretraining scale of \modelname is primarily constrained by our available computational resources. In this study, we pretrain \modelname on 43 real-world datasets using a moderate-size diffusion transformer and decoder architecture. Despite this moderate scale, \modelname already demonstrates strong performance across 45 unseen real-world benchmark datasets, outperforming existing tabular foundation generators by a clear margin. Importantly, we observe that model performance consistently improves as the number of pretraining datasets increases, with no clear signs of saturation in our experiments. These results suggest that the current computational budget remains far from fully exploiting the potential of the proposed framework and that further scaling of both the pretraining data and model capacity could yield additional performance gains.
A promising direction is therefore to scale up \modelname along multiple axes. First, increasing the number and diversity of pretraining datasets may allow the model to acquire broader tabular structural priors. Second, increasing the capacity of the diffusion transformer and decoder may further improve its ability to capture complex latent distributions and refine denoised embeddings into generation-ready representations. Third, longer pretraining may further improve the stability and transferability of the learned generative representations. We therefore plan to expand the pretraining scale in future work, including larger upstream tabular corpora, larger model architectures, and a more extensive pretraining stage.

\textbf{Improve the implementation for computational efficiency.}
For generalisability and stability, the current implementation of \modelname relies primarily on standard PyTorch modules~\cite{paszke2019pytorch}, such as standard attention mechanisms and transformer layers. This choice makes the framework easier to reproduce and extend across heterogeneous tabular datasets. However, it also leaves substantial room for further engineering optimisation.
For instance, future implementations can integrate more efficient attention kernels, such as FlashAttention~\cite{dao2022flashattention} and FlashAttention-2~\cite{dao2023flashattention}, to reduce memory usage and improve throughput during both pretraining and fitting. These efficiency improvements could make large-scale pretraining more practical and enable \modelname to scale to broader tabular corpora and larger model variants.

\textbf{Explore tabular data generation with explicit causal structures.}
As discussed in~\Cref{section:intro} and~\Cref{sec:method}, \modelname is designed to leverage implicit structural signals encoded in latent embeddings, rather than explicitly estimating or conditioning on causal graphs. This design choice is important, as reliable causal graph discovery from real-world tabular data remains an open research problem~\cite{jiang2026tabstruct, kaddour2022causal, tu2024causality, glymour2019review, nastl2024causal}. 
In practice, ground-truth structures are rarely available, and prior studies~\cite{tu2024causality, jiang2026tabstruct} have shown that even state-of-the-art causal discovery methods may perform poorly or yield misleading structures on real-world datasets. As a result, generators that rely heavily on explicitly learned causal structures may inherit errors from imperfect causal discovery and underperform on real-world benchmarks~\cite{jiang2026tabstruct}, including methods such as GOGGLE~\cite{liu2023goggle} and Bayesian Network~\cite{kitson2023survey}.  \modelname instead uses implicit structural information embedded in pretrained latent representations as an inductive bias for modelling global inter-feature dependency structures, while avoiding reliance on potentially inaccurate and unverifiable explicit causal structures. Nevertheless, further bridging implicit and explicit causal modelling remains a promising and interesting direction for improving tabular data generation.

\textbf{Investigate tabular data generation under distribution shifts.}
Following well-established benchmarks~\cite{erickson2025tabarena, jiang2026tabstruct, grinsztajn2022tree, mcelfresh2024neural}, our experiments aim to demonstrate generalisation to unseen datasets rather than to claim robustness under controlled distribution shifts.
That being said, we would like to emphasise that the fitting-behaviour analysis provides empirical evidence of a non-trivial difference between the pretraining and evaluation distributions. If the evaluation datasets were effectively in-distribution, \modelname would be expected to exhibit a higher risk of overfitting when adapted to them. However, \Cref{fig:overfitting} shows that \modelname consistently reduces overfitting risk while preserving synthetic data quality, suggesting that its performance cannot be explained solely by memorisation or retrieval of overlapping pretraining datasets.
Our evaluation is also comparatively broad for tabular data generation. To the best of our knowledge, TabStruct~\cite{jiang2026tabstruct} is among the largest prior benchmarks in this area, comprising 23 real-world datasets. In comparison, our evaluation covers 45 real-world datasets spanning diverse domains, feature types, dataset sizes, and prediction tasks. This breadth reduces the likelihood that the observed results are driven by a small or narrowly distributed collection of datasets.
Nevertheless, we acknowledge the value of evaluating \modelname and benchmark methods under controlled out-of-distribution (OOD) settings~\cite{purucker2026beyond, gardner2023benchmarking}, which could provide further insights into model deployment under pronounced distribution shifts, such as temporal shifts. Prior work~\cite{hoo2025tables} has shown that tabular foundation models can be adapted to time-series tasks because certain representation learning capabilities transfer across the two modalities, suggesting a potentially valuable direction for extending \modelname to OOD settings.

\clearpage
\section{Broader Impacts}
\label[appendix]{appendix:broader_impacts}

This paper presents \modelname, a tabular foundation model for generative modelling that aims to advance synthetic tabular data generation through structure-aware latent representations and denoising-aligned latent diffusion. By learning transferable generative knowledge across diverse tabular datasets, \modelname can reduce the computational overheads compared to training dataset-specific generators from scratch, thereby improving the practicability of high-quality tabular data synthesis in data-scarce and computation-constrained scenarios.

These characteristics are particularly valuable in domains where tabular data is central yet difficult to acquire or share, such as healthcare~\cite{hernandez2022synthetic}, business analytics~\cite{sattarov2023findiff}, and scientific discovery~\cite{borisov2022deep, margeloiu2024tabebm}. For instance, high-fidelity synthetic tabular data can support privacy-preserving data sharing, benchmarking, model development, and exploratory analysis when access to real data is restricted by confidentiality. Moreover, by preserving global inter-feature dependency structures more effectively than existing tabular foundation generators, \modelname may facilitate downstream studies that depend on reliable inter-feature dependency structures, such as scientific simulation~\cite{clark2023testing} and decision-support systems~\cite{marwala2015causality}.
We will make \modelname open-source to facilitate community collaboration on high-fidelity tabular foundation generators. We hope that releasing the code would encourage the development of more practically useful tabular generative models.

\clearpage
\section{Extended Experimental Results}
\label[appendix]{appendix:extended_results}

In this section, we provide the per-dataset evaluation results for all 45 benchmark datasets. We note that, for the real holdout data, we evaluate it only in the metrics for which the training data serves as a lower/upper-bound reference (i.e., excluding local and global utility) to assess overfitting risks.

\begin{table}[!hp]
\centering
\caption{\textbf{Raw benchmark results of 23 tabular generators on the ``Airfoil'' dataset.} We report the mean $\pm$ std of each metric across 10 repeated data splits. For benchmark generators, ``$-$'' denotes failed convergence of a specific model or unexpected values in the synthetic data that caused the evaluation metric computation to crash. We highlight the {\color[HTML]{008080} \textbf{First}}, {\color[HTML]{7030A0} \textbf{Second}}, and {\color[HTML]{C65911} \textbf{Third}} best performances for each metric. \modelname generally achieves competitive performance against the benchmark generators while maintaining a reduced risk of overfitting.}
\resizebox{\textwidth}{!}{
\begin{tabular}{l|rrrr|rr|r|r}

\toprule

\multirow{2}{*}{\textbf{Generator}} & \multicolumn{4}{c|}{\textbf{Density   Estimation}}                             & \multicolumn{2}{c|}{\textbf{Privacy Preservation}} & \multicolumn{1}{c|}{\textbf{ML Efficacy}}                              & \multicolumn{1}{c}{\textbf{Structural Fidelity}} \\

& Shape $\uparrow$ & Trend $\uparrow$ & $\alpha$-precision $\uparrow$ & $\beta$-recall $\uparrow$ & Authenticity $\uparrow$ & DCR Score $\uparrow$ & Local utility $\uparrow$ & Global utility $\uparrow$  \\

\midrule

\rowcolor{Gainsboro!60}

\multicolumn{9}{c}{\textbf{Real Data}} \\

\midrule

Real Data (Train) & 1.00$_{\pm\text{0.00}}$ & 1.00$_{\pm\text{0.00}}$ & 1.00$_{\pm\text{0.00}}$ & 1.00$_{\pm\text{0.00}}$ & 0.00$_{\pm\text{0.00}}$ & 0.00$_{\pm\text{0.00}}$ & 1.00$_{\pm\text{0.00}}$ & 1.00$_{\pm\text{0.00}}$ \\

Real Data (Holdout) & 0.95$_{\pm\text{0.01}}$ & 0.86$_{\pm\text{0.02}}$ & 0.97$_{\pm\text{0.01}}$ & 0.50$_{\pm\text{0.03}}$ & 0.61$_{\pm\text{0.02}}$ & 0.04$_{\pm\text{0.00}}$ & $-$ & $-$ \\

\midrule

\rowcolor{Gainsboro!60}

\multicolumn{9}{c}{\textbf{Dataset-specific Model}} \\

\midrule

SMOTE & {\color[HTML]{008080} \textbf{0.89}}$_{\pm\text{0.02}}$ & 0.53$_{\pm\text{0.01}}$ & 0.81$_{\pm\text{0.01}}$ & {\color[HTML]{008080} \textbf{0.54}}$_{\pm\text{0.03}}$ & 0.54$_{\pm\text{0.01}}$ & 0.03$_{\pm\text{0.00}}$ & {\color[HTML]{C65911} \textbf{0.80}}$_{\pm\text{0.05}}$ & 0.52$_{\pm\text{0.02}}$ \\

TabSDS & $-$ & $-$ & $-$ & $-$ & $-$ & $-$ & $-$ & $-$ \\
\bottomrule

\end{tabular}
}
\end{table}

\begin{table}[!htbp]
\centering
\caption{\textbf{Raw benchmark results of 23 tabular generators on the ``BankChurn'' dataset.} We report the mean $\pm$ std of each metric across 10 repeated data splits. For benchmark generators, ``$-$'' denotes failed convergence of a specific model or unexpected values in the synthetic data that caused the evaluation metric computation to crash. We highlight the {\color[HTML]{008080} \textbf{First}}, {\color[HTML]{7030A0} \textbf{Second}}, and {\color[HTML]{C65911} \textbf{Third}} best performances for each metric. \modelname generally achieves competitive performance against the benchmark generators while maintaining a reduced risk of overfitting.}
\resizebox{\textwidth}{!}{
\begin{tabular}{l|rrrr|rr|r|r}

\toprule

\multirow{2}{*}{\textbf{Generator}} & \multicolumn{4}{c|}{\textbf{Density   Estimation}}                             & \multicolumn{2}{c|}{\textbf{Privacy Preservation}} & \multicolumn{1}{c|}{\textbf{ML Efficacy}}                              & \multicolumn{1}{c}{\textbf{Structural Fidelity}} \\

& Shape $\uparrow$ & Trend $\uparrow$ & $\alpha$-precision $\uparrow$ & $\beta$-recall $\uparrow$ & Authenticity $\uparrow$ & DCR Score $\uparrow$ & Local utility $\uparrow$ & Global utility $\uparrow$  \\

\midrule

\rowcolor{Gainsboro!60}

\multicolumn{9}{c}{\textbf{Real Data}} \\

\midrule

Real Data (Train) & 1.00$_{\pm\text{0.00}}$ & 1.00$_{\pm\text{0.00}}$ & 1.00$_{\pm\text{0.00}}$ & 1.00$_{\pm\text{0.00}}$ & 0.00$_{\pm\text{0.00}}$ & 0.00$_{\pm\text{0.00}}$ & 1.00$_{\pm\text{0.00}}$ & 0.79$_{\pm\text{0.07}}$ \\

Real Data (Holdout) & 0.99$_{\pm\text{0.00}}$ & 0.94$_{\pm\text{0.02}}$ & 0.99$_{\pm\text{0.01}}$ & 0.50$_{\pm\text{0.01}}$ & 0.67$_{\pm\text{0.01}}$ & 0.24$_{\pm\text{0.00}}$ & $-$ & $-$ \\

\midrule

\rowcolor{Gainsboro!60}

\multicolumn{9}{c}{\textbf{Dataset-specific Model}} \\

\midrule

SMOTE & {\color[HTML]{7030A0} \textbf{0.94}}$_{\pm\text{0.00}}$ & 0.66$_{\pm\text{0.02}}$ & 0.86$_{\pm\text{0.01}}$ & {\color[HTML]{008080} \textbf{0.67}}$_{\pm\text{0.01}}$ & 0.42$_{\pm\text{0.01}}$ & 0.13$_{\pm\text{0.00}}$ & {\color[HTML]{008080} \textbf{1.00}}$_{\pm\text{0.01}}$ & 0.64$_{\pm\text{0.06}}$ \\

TabSDS & 0.67$_{\pm\text{0.00}}$ & 0.54$_{\pm\text{0.02}}$ & 0.64$_{\pm\text{0.00}}$ & 0.23$_{\pm\text{0.00}}$ & 0.61$_{\pm\text{0.00}}$ & 0.27$_{\pm\text{0.00}}$ & 0.73$_{\pm\text{0.01}}$ & 0.51$_{\pm\text{0.05}}$ \\
\bottomrule

\end{tabular}
}
\end{table}

\begin{table}[!htbp]
\centering
\caption{\textbf{Raw benchmark results of 23 tabular generators on the ``Bankruptcy'' dataset.} We report the mean $\pm$ std of each metric across 10 repeated data splits. For benchmark generators, ``$-$'' denotes failed convergence of a specific model or unexpected values in the synthetic data that caused the evaluation metric computation to crash. We highlight the {\color[HTML]{008080} \textbf{First}}, {\color[HTML]{7030A0} \textbf{Second}}, and {\color[HTML]{C65911} \textbf{Third}} best performances for each metric. \modelname generally achieves competitive performance against the benchmark generators while maintaining a reduced risk of overfitting.}
\resizebox{\textwidth}{!}{
\begin{tabular}{l|rrrr|rr|r|r}

\toprule

\multirow{2}{*}{\textbf{Generator}} & \multicolumn{4}{c|}{\textbf{Density   Estimation}}                             & \multicolumn{2}{c|}{\textbf{Privacy Preservation}} & \multicolumn{1}{c|}{\textbf{ML Efficacy}}                              & \multicolumn{1}{c}{\textbf{Structural Fidelity}} \\

& Shape $\uparrow$ & Trend $\uparrow$ & $\alpha$-precision $\uparrow$ & $\beta$-recall $\uparrow$ & Authenticity $\uparrow$ & DCR Score $\uparrow$ & Local utility $\uparrow$ & Global utility $\uparrow$  \\

\midrule

\rowcolor{Gainsboro!60}

\multicolumn{9}{c}{\textbf{Real Data}} \\

\midrule

Real Data (Train) & 1.00$_{\pm\text{0.00}}$ & 1.00$_{\pm\text{0.00}}$ & 1.00$_{\pm\text{0.00}}$ & 1.00$_{\pm\text{0.00}}$ & 0.00$_{\pm\text{0.00}}$ & 0.00$_{\pm\text{0.00}}$ & 1.00$_{\pm\text{0.00}}$ & 0.94$_{\pm\text{0.02}}$ \\

Real Data (Holdout) & 0.98$_{\pm\text{0.00}}$ & 0.87$_{\pm\text{0.02}}$ & 0.98$_{\pm\text{0.01}}$ & 0.50$_{\pm\text{0.01}}$ & 0.73$_{\pm\text{0.01}}$ & 0.03$_{\pm\text{0.00}}$ & $-$ & $-$ \\

\midrule

\rowcolor{Gainsboro!60}

\multicolumn{9}{c}{\textbf{Dataset-specific Model}} \\

\midrule

SMOTE & {\color[HTML]{C65911} \textbf{0.76}}$_{\pm\text{0.01}}$ & 0.73$_{\pm\text{0.01}}$ & 0.82$_{\pm\text{0.02}}$ & {\color[HTML]{008080} \textbf{0.67}}$_{\pm\text{0.01}}$ & 0.46$_{\pm\text{0.01}}$ & 0.02$_{\pm\text{0.00}}$ & {\color[HTML]{7030A0} \textbf{1.01}}$_{\pm\text{0.05}}$ & {\color[HTML]{7030A0} \textbf{0.66}}$_{\pm\text{0.02}}$ \\

TabSDS & 0.54$_{\pm\text{0.01}}$ & 0.64$_{\pm\text{0.02}}$ & 0.57$_{\pm\text{0.02}}$ & 0.03$_{\pm\text{0.02}}$ & 0.67$_{\pm\text{0.01}}$ & 0.08$_{\pm\text{0.00}}$ & 0.66$_{\pm\text{0.02}}$ & 0.46$_{\pm\text{0.03}}$ \\
\bottomrule

\end{tabular}
}
\end{table}

\begin{table}[!htbp]
\centering
\caption{\textbf{Raw benchmark results of 23 tabular generators on the ``Biodeg'' dataset.} We report the mean $\pm$ std of each metric across 10 repeated data splits. For benchmark generators, ``$-$'' denotes failed convergence of a specific model or unexpected values in the synthetic data that caused the evaluation metric computation to crash. We highlight the {\color[HTML]{008080} \textbf{First}}, {\color[HTML]{7030A0} \textbf{Second}}, and {\color[HTML]{C65911} \textbf{Third}} best performances for each metric. \modelname generally achieves competitive performance against the benchmark generators while maintaining a reduced risk of overfitting.}
\resizebox{\textwidth}{!}{
\begin{tabular}{l|rrrr|rr|r|r}

\toprule

\multirow{2}{*}{\textbf{Generator}} & \multicolumn{4}{c|}{\textbf{Density   Estimation}}                             & \multicolumn{2}{c|}{\textbf{Privacy Preservation}} & \multicolumn{1}{c|}{\textbf{ML Efficacy}}                              & \multicolumn{1}{c}{\textbf{Structural Fidelity}} \\

& Shape $\uparrow$ & Trend $\uparrow$ & $\alpha$-precision $\uparrow$ & $\beta$-recall $\uparrow$ & Authenticity $\uparrow$ & DCR Score $\uparrow$ & Local utility $\uparrow$ & Global utility $\uparrow$  \\

\midrule

\rowcolor{Gainsboro!60}

\multicolumn{9}{c}{\textbf{Real Data}} \\

\midrule

Real Data (Train) & 1.00$_{\pm\text{0.00}}$ & 1.00$_{\pm\text{0.00}}$ & 1.00$_{\pm\text{0.00}}$ & 1.00$_{\pm\text{0.00}}$ & 0.00$_{\pm\text{0.00}}$ & 0.00$_{\pm\text{0.00}}$ & 1.00$_{\pm\text{0.00}}$ & 0.95$_{\pm\text{0.01}}$ \\

Real Data (Holdout) & 0.96$_{\pm\text{0.01}}$ & 0.87$_{\pm\text{0.02}}$ & 0.96$_{\pm\text{0.01}}$ & 0.51$_{\pm\text{0.03}}$ & 0.69$_{\pm\text{0.02}}$ & 0.04$_{\pm\text{0.00}}$ & $-$ & $-$ \\

\midrule

\rowcolor{Gainsboro!60}

\multicolumn{9}{c}{\textbf{Dataset-specific Model}} \\

\midrule

SMOTE & 0.64$_{\pm\text{0.00}}$ & 0.71$_{\pm\text{0.01}}$ & 0.72$_{\pm\text{0.02}}$ & {\color[HTML]{008080} \textbf{0.64}}$_{\pm\text{0.02}}$ & 0.48$_{\pm\text{0.02}}$ & 0.02$_{\pm\text{0.00}}$ & {\color[HTML]{008080} \textbf{0.96}}$_{\pm\text{0.02}}$ & {\color[HTML]{C65911} \textbf{0.56}}$_{\pm\text{0.03}}$ \\

TabSDS & 0.45$_{\pm\text{0.00}}$ & 0.56$_{\pm\text{0.02}}$ & 0.70$_{\pm\text{0.01}}$ & 0.09$_{\pm\text{0.01}}$ & 0.77$_{\pm\text{0.01}}$ & 0.08$_{\pm\text{0.01}}$ & 0.64$_{\pm\text{0.02}}$ & 0.35$_{\pm\text{0.02}}$ \\
\bottomrule

\end{tabular}
}
\end{table}

\begin{table}[!htbp]
\centering
\caption{\textbf{Raw benchmark results of 23 tabular generators on the ``California'' dataset.} We report the mean $\pm$ std of each metric across 10 repeated data splits. For benchmark generators, ``$-$'' denotes failed convergence of a specific model or unexpected values in the synthetic data that caused the evaluation metric computation to crash. We highlight the {\color[HTML]{008080} \textbf{First}}, {\color[HTML]{7030A0} \textbf{Second}}, and {\color[HTML]{C65911} \textbf{Third}} best performances for each metric. \modelname generally achieves competitive performance against the benchmark generators while maintaining a reduced risk of overfitting.}
\resizebox{\textwidth}{!}{
\begin{tabular}{l|rrrr|rr|r|r}

\toprule

\multirow{2}{*}{\textbf{Generator}} & \multicolumn{4}{c|}{\textbf{Density   Estimation}}                             & \multicolumn{2}{c|}{\textbf{Privacy Preservation}} & \multicolumn{1}{c|}{\textbf{ML Efficacy}}                              & \multicolumn{1}{c}{\textbf{Structural Fidelity}} \\

& Shape $\uparrow$ & Trend $\uparrow$ & $\alpha$-precision $\uparrow$ & $\beta$-recall $\uparrow$ & Authenticity $\uparrow$ & DCR Score $\uparrow$ & Local utility $\uparrow$ & Global utility $\uparrow$  \\

\midrule

\rowcolor{Gainsboro!60}

\multicolumn{9}{c}{\textbf{Real Data}} \\

\midrule

Real Data (Train) & 1.00$_{\pm\text{0.00}}$ & 1.00$_{\pm\text{0.00}}$ & 1.00$_{\pm\text{0.00}}$ & 1.00$_{\pm\text{0.00}}$ & 0.00$_{\pm\text{0.00}}$ & 0.00$_{\pm\text{0.00}}$ & 1.00$_{\pm\text{0.00}}$ & 1.00$_{\pm\text{0.00}}$ \\

Real Data (Holdout) & 0.99$_{\pm\text{0.00}}$ & 0.98$_{\pm\text{0.00}}$ & 0.99$_{\pm\text{0.00}}$ & 0.49$_{\pm\text{0.01}}$ & 0.69$_{\pm\text{0.01}}$ & 0.04$_{\pm\text{0.00}}$ & $-$ & $-$ \\

\midrule

\rowcolor{Gainsboro!60}

\multicolumn{9}{c}{\textbf{Dataset-specific Model}} \\

\midrule

SMOTE & {\color[HTML]{008080} \textbf{0.98}}$_{\pm\text{0.00}}$ & 0.73$_{\pm\text{0.00}}$ & 0.79$_{\pm\text{0.01}}$ & {\color[HTML]{008080} \textbf{0.66}}$_{\pm\text{0.01}}$ & 0.44$_{\pm\text{0.01}}$ & 0.05$_{\pm\text{0.00}}$ & {\color[HTML]{008080} \textbf{0.91}}$_{\pm\text{0.01}}$ & {\color[HTML]{C65911} \textbf{0.64}}$_{\pm\text{0.01}}$ \\

TabSDS & 0.76$_{\pm\text{0.00}}$ & 0.61$_{\pm\text{0.01}}$ & 0.74$_{\pm\text{0.01}}$ & 0.17$_{\pm\text{0.00}}$ & 0.71$_{\pm\text{0.00}}$ & 0.06$_{\pm\text{0.01}}$ & 0.58$_{\pm\text{0.01}}$ & 0.52$_{\pm\text{0.01}}$ \\
\bottomrule

\end{tabular}
}
\end{table}

\begin{table}[!htbp]
\centering
\caption{\textbf{Raw benchmark results of 23 tabular generators on the ``Card'' dataset.} We report the mean $\pm$ std of each metric across 10 repeated data splits. For benchmark generators, ``$-$'' denotes failed convergence of a specific model or unexpected values in the synthetic data that caused the evaluation metric computation to crash. We highlight the {\color[HTML]{008080} \textbf{First}}, {\color[HTML]{7030A0} \textbf{Second}}, and {\color[HTML]{C65911} \textbf{Third}} best performances for each metric. \modelname generally achieves competitive performance against the benchmark generators while maintaining a reduced risk of overfitting.}
\resizebox{\textwidth}{!}{
\begin{tabular}{l|rrrr|rr|r|r}

\toprule

\multirow{2}{*}{\textbf{Generator}} & \multicolumn{4}{c|}{\textbf{Density   Estimation}}                             & \multicolumn{2}{c|}{\textbf{Privacy Preservation}} & \multicolumn{1}{c|}{\textbf{ML Efficacy}}                              & \multicolumn{1}{c}{\textbf{Structural Fidelity}} \\

& Shape $\uparrow$ & Trend $\uparrow$ & $\alpha$-precision $\uparrow$ & $\beta$-recall $\uparrow$ & Authenticity $\uparrow$ & DCR Score $\uparrow$ & Local utility $\uparrow$ & Global utility $\uparrow$  \\

\midrule

\rowcolor{Gainsboro!60}

\multicolumn{9}{c}{\textbf{Real Data}} \\

\midrule

Real Data (Train) & 1.00$_{\pm\text{0.00}}$ & 1.00$_{\pm\text{0.00}}$ & 1.00$_{\pm\text{0.00}}$ & 1.00$_{\pm\text{0.00}}$ & 0.00$_{\pm\text{0.00}}$ & 0.00$_{\pm\text{0.00}}$ & 1.00$_{\pm\text{0.00}}$ & 0.96$_{\pm\text{0.03}}$ \\

Real Data (Holdout) & 0.99$_{\pm\text{0.00}}$ & 0.99$_{\pm\text{0.00}}$ & 0.99$_{\pm\text{0.00}}$ & 0.50$_{\pm\text{0.00}}$ & 0.73$_{\pm\text{0.00}}$ & 0.02$_{\pm\text{0.00}}$ & $-$ & $-$ \\

\midrule

\rowcolor{Gainsboro!60}

\multicolumn{9}{c}{\textbf{Dataset-specific Model}} \\

\midrule

SMOTE & {\color[HTML]{C65911} \textbf{0.80}}$_{\pm\text{0.01}}$ & $-$ & $-$ & {\color[HTML]{008080} \textbf{0.65}}$_{\pm\text{0.00}}$ & 0.49$_{\pm\text{0.00}}$ & 0.02$_{\pm\text{0.00}}$ & {\color[HTML]{7030A0} \textbf{0.99}}$_{\pm\text{0.01}}$ & $-$ \\

TabSDS & 0.59$_{\pm\text{0.00}}$ & 0.69$_{\pm\text{0.00}}$ & 0.72$_{\pm\text{0.01}}$ & 0.17$_{\pm\text{0.01}}$ & 0.73$_{\pm\text{0.00}}$ & 0.04$_{\pm\text{0.00}}$ & 0.77$_{\pm\text{0.01}}$ & 0.48$_{\pm\text{0.02}}$ \\
\bottomrule

\end{tabular}
}
\end{table}

\begin{table}[!htbp]
\centering
\caption{\textbf{Raw benchmark results of 23 tabular generators on the ``Churn'' dataset.} We report the mean $\pm$ std of each metric across 10 repeated data splits. For benchmark generators, ``$-$'' denotes failed convergence of a specific model or unexpected values in the synthetic data that caused the evaluation metric computation to crash. We highlight the {\color[HTML]{008080} \textbf{First}}, {\color[HTML]{7030A0} \textbf{Second}}, and {\color[HTML]{C65911} \textbf{Third}} best performances for each metric. \modelname generally achieves competitive performance against the benchmark generators while maintaining a reduced risk of overfitting.}
\resizebox{\textwidth}{!}{
\begin{tabular}{l|rrrr|rr|r|r}

\toprule

\multirow{2}{*}{\textbf{Generator}} & \multicolumn{4}{c|}{\textbf{Density   Estimation}}                             & \multicolumn{2}{c|}{\textbf{Privacy Preservation}} & \multicolumn{1}{c|}{\textbf{ML Efficacy}}                              & \multicolumn{1}{c}{\textbf{Structural Fidelity}} \\

& Shape $\uparrow$ & Trend $\uparrow$ & $\alpha$-precision $\uparrow$ & $\beta$-recall $\uparrow$ & Authenticity $\uparrow$ & DCR Score $\uparrow$ & Local utility $\uparrow$ & Global utility $\uparrow$  \\

\midrule

\rowcolor{Gainsboro!60}

\multicolumn{9}{c}{\textbf{Real Data}} \\

\midrule

Real Data (Train) & 1.00$_{\pm\text{0.00}}$ & 1.00$_{\pm\text{0.00}}$ & 1.00$_{\pm\text{0.00}}$ & 1.00$_{\pm\text{0.00}}$ & 0.00$_{\pm\text{0.00}}$ & 0.00$_{\pm\text{0.00}}$ & 1.00$_{\pm\text{0.00}}$ & 0.84$_{\pm\text{0.03}}$ \\

Real Data (Holdout) & 0.98$_{\pm\text{0.00}}$ & 0.97$_{\pm\text{0.01}}$ & 0.98$_{\pm\text{0.01}}$ & 0.50$_{\pm\text{0.01}}$ & 0.68$_{\pm\text{0.01}}$ & 0.23$_{\pm\text{0.01}}$ & $-$ & $-$ \\

\midrule

\rowcolor{Gainsboro!60}

\multicolumn{9}{c}{\textbf{Dataset-specific Model}} \\

\midrule

SMOTE & {\color[HTML]{7030A0} \textbf{0.91}}$_{\pm\text{0.00}}$ & 0.79$_{\pm\text{0.01}}$ & 0.72$_{\pm\text{0.01}}$ & {\color[HTML]{008080} \textbf{0.66}}$_{\pm\text{0.02}}$ & 0.45$_{\pm\text{0.01}}$ & 0.06$_{\pm\text{0.00}}$ & {\color[HTML]{008080} \textbf{0.95}}$_{\pm\text{0.02}}$ & 0.49$_{\pm\text{0.02}}$ \\

TabSDS & $-$ & $-$ & $-$ & $-$ & $-$ & $-$ & $-$ & $-$ \\
\bottomrule

\end{tabular}
}
\end{table}

\begin{table}[!htbp]
\centering
\caption{\textbf{Raw benchmark results of 23 tabular generators on the ``Coil2000'' dataset.} We report the mean $\pm$ std of each metric across 10 repeated data splits. For benchmark generators, ``$-$'' denotes failed convergence of a specific model or unexpected values in the synthetic data that caused the evaluation metric computation to crash. We highlight the {\color[HTML]{008080} \textbf{First}}, {\color[HTML]{7030A0} \textbf{Second}}, and {\color[HTML]{C65911} \textbf{Third}} best performances for each metric. \modelname generally achieves competitive performance against the benchmark generators while maintaining a reduced risk of overfitting.}
\resizebox{\textwidth}{!}{
\begin{tabular}{l|rrrr|rr|r|r}

\toprule

\multirow{2}{*}{\textbf{Generator}} & \multicolumn{4}{c|}{\textbf{Density   Estimation}}                             & \multicolumn{2}{c|}{\textbf{Privacy Preservation}} & \multicolumn{1}{c|}{\textbf{ML Efficacy}}                              & \multicolumn{1}{c}{\textbf{Structural Fidelity}} \\

& Shape $\uparrow$ & Trend $\uparrow$ & $\alpha$-precision $\uparrow$ & $\beta$-recall $\uparrow$ & Authenticity $\uparrow$ & DCR Score $\uparrow$ & Local utility $\uparrow$ & Global utility $\uparrow$  \\

\midrule

\rowcolor{Gainsboro!60}

\multicolumn{9}{c}{\textbf{Real Data}} \\

\midrule

Real Data (Train) & 1.00$_{\pm\text{0.00}}$ & 1.00$_{\pm\text{0.00}}$ & 1.00$_{\pm\text{0.00}}$ & 1.00$_{\pm\text{0.00}}$ & 0.01$_{\pm\text{0.00}}$ & 0.00$_{\pm\text{0.00}}$ & 1.00$_{\pm\text{0.00}}$ & 0.95$_{\pm\text{0.01}}$ \\

Real Data (Holdout) & 0.99$_{\pm\text{0.00}}$ & 0.97$_{\pm\text{0.00}}$ & 0.99$_{\pm\text{0.01}}$ & 0.54$_{\pm\text{0.01}}$ & 0.65$_{\pm\text{0.01}}$ & 0.03$_{\pm\text{0.00}}$ & $-$ & $-$ \\

\midrule

\rowcolor{Gainsboro!60}

\multicolumn{9}{c}{\textbf{Dataset-specific Model}} \\

\midrule

SMOTE & 0.47$_{\pm\text{0.00}}$ & 0.71$_{\pm\text{0.01}}$ & 0.71$_{\pm\text{0.03}}$ & {\color[HTML]{008080} \textbf{0.52}}$_{\pm\text{0.01}}$ & 0.60$_{\pm\text{0.01}}$ & 0.02$_{\pm\text{0.00}}$ & {\color[HTML]{008080} \textbf{1.01}}$_{\pm\text{0.02}}$ & {\color[HTML]{7030A0} \textbf{0.55}}$_{\pm\text{0.01}}$ \\

TabSDS & $-$ & $-$ & $-$ & $-$ & $-$ & $-$ & $-$ & $-$ \\
\bottomrule

\end{tabular}
}
\end{table}

\begin{table}[!htbp]
\centering
\caption{\textbf{Raw benchmark results of 23 tabular generators on the ``Company'' dataset.} We report the mean $\pm$ std of each metric across 10 repeated data splits. For benchmark generators, ``$-$'' denotes failed convergence of a specific model or unexpected values in the synthetic data that caused the evaluation metric computation to crash. We highlight the {\color[HTML]{008080} \textbf{First}}, {\color[HTML]{7030A0} \textbf{Second}}, and {\color[HTML]{C65911} \textbf{Third}} best performances for each metric. \modelname generally achieves competitive performance against the benchmark generators while maintaining a reduced risk of overfitting.}
\resizebox{\textwidth}{!}{
\begin{tabular}{l|rrrr|rr|r|r}

\toprule

\multirow{2}{*}{\textbf{Generator}} & \multicolumn{4}{c|}{\textbf{Density   Estimation}}                             & \multicolumn{2}{c|}{\textbf{Privacy Preservation}} & \multicolumn{1}{c|}{\textbf{ML Efficacy}}                              & \multicolumn{1}{c}{\textbf{Structural Fidelity}} \\

& Shape $\uparrow$ & Trend $\uparrow$ & $\alpha$-precision $\uparrow$ & $\beta$-recall $\uparrow$ & Authenticity $\uparrow$ & DCR Score $\uparrow$ & Local utility $\uparrow$ & Global utility $\uparrow$  \\

\midrule

\rowcolor{Gainsboro!60}

\multicolumn{9}{c}{\textbf{Real Data}} \\

\midrule

Real Data (Train) & 1.00$_{\pm\text{0.00}}$ & 1.00$_{\pm\text{0.00}}$ & 1.00$_{\pm\text{0.00}}$ & 1.00$_{\pm\text{0.00}}$ & 0.00$_{\pm\text{0.00}}$ & 0.00$_{\pm\text{0.00}}$ & 1.00$_{\pm\text{0.00}}$ & 0.88$_{\pm\text{0.03}}$ \\

Real Data (Holdout) & 0.98$_{\pm\text{0.00}}$ & 0.71$_{\pm\text{0.02}}$ & 0.98$_{\pm\text{0.01}}$ & 0.50$_{\pm\text{0.02}}$ & 0.75$_{\pm\text{0.02}}$ & 0.00$_{\pm\text{0.00}}$ & $-$ & $-$ \\

\midrule

\rowcolor{Gainsboro!60}

\multicolumn{9}{c}{\textbf{Dataset-specific Model}} \\

\midrule

SMOTE & {\color[HTML]{7030A0} \textbf{0.94}}$_{\pm\text{0.01}}$ & 0.65$_{\pm\text{0.04}}$ & 0.79$_{\pm\text{0.01}}$ & {\color[HTML]{008080} \textbf{0.68}}$_{\pm\text{0.02}}$ & 0.47$_{\pm\text{0.02}}$ & 0.00$_{\pm\text{0.00}}$ & {\color[HTML]{7030A0} \textbf{0.99}}$_{\pm\text{0.03}}$ & 0.66$_{\pm\text{0.03}}$ \\

TabSDS & 0.68$_{\pm\text{0.00}}$ & 0.46$_{\pm\text{0.02}}$ & 0.57$_{\pm\text{0.01}}$ & 0.14$_{\pm\text{0.01}}$ & 0.67$_{\pm\text{0.01}}$ & 0.00$_{\pm\text{0.00}}$ & 0.71$_{\pm\text{0.04}}$ & 0.46$_{\pm\text{0.02}}$ \\
\bottomrule

\end{tabular}
}
\end{table}

\begin{table}[!htbp]
\centering
\caption{\textbf{Raw benchmark results of 23 tabular generators on the ``Concrete'' dataset.} We report the mean $\pm$ std of each metric across 10 repeated data splits. For benchmark generators, ``$-$'' denotes failed convergence of a specific model or unexpected values in the synthetic data that caused the evaluation metric computation to crash. We highlight the {\color[HTML]{008080} \textbf{First}}, {\color[HTML]{7030A0} \textbf{Second}}, and {\color[HTML]{C65911} \textbf{Third}} best performances for each metric. \modelname generally achieves competitive performance against the benchmark generators while maintaining a reduced risk of overfitting.}
\resizebox{\textwidth}{!}{
\begin{tabular}{l|rrrr|rr|r|r}

\toprule

\multirow{2}{*}{\textbf{Generator}} & \multicolumn{4}{c|}{\textbf{Density   Estimation}}                             & \multicolumn{2}{c|}{\textbf{Privacy Preservation}} & \multicolumn{1}{c|}{\textbf{ML Efficacy}}                              & \multicolumn{1}{c}{\textbf{Structural Fidelity}} \\

& Shape $\uparrow$ & Trend $\uparrow$ & $\alpha$-precision $\uparrow$ & $\beta$-recall $\uparrow$ & Authenticity $\uparrow$ & DCR Score $\uparrow$ & Local utility $\uparrow$ & Global utility $\uparrow$  \\

\midrule

\rowcolor{Gainsboro!60}

\multicolumn{9}{c}{\textbf{Real Data}} \\

\midrule

Real Data (Train) & 1.00$_{\pm\text{0.00}}$ & 1.00$_{\pm\text{0.00}}$ & 1.00$_{\pm\text{0.00}}$ & 1.00$_{\pm\text{0.00}}$ & 0.00$_{\pm\text{0.00}}$ & 0.00$_{\pm\text{0.00}}$ & 1.00$_{\pm\text{0.00}}$ & 1.00$_{\pm\text{0.00}}$ \\

Real Data (Holdout) & 0.94$_{\pm\text{0.01}}$ & 0.98$_{\pm\text{0.01}}$ & 0.95$_{\pm\text{0.02}}$ & 0.51$_{\pm\text{0.04}}$ & 0.58$_{\pm\text{0.03}}$ & 0.07$_{\pm\text{0.00}}$ & $-$ & $-$ \\

\midrule

\rowcolor{Gainsboro!60}

\multicolumn{9}{c}{\textbf{Dataset-specific Model}} \\

\midrule

SMOTE & {\color[HTML]{C65911} \textbf{0.79}}$_{\pm\text{0.01}}$ & $-$ & $-$ & {\color[HTML]{008080} \textbf{0.50}}$_{\pm\text{0.03}}$ & 0.60$_{\pm\text{0.03}}$ & 0.05$_{\pm\text{0.00}}$ & {\color[HTML]{008080} \textbf{0.83}}$_{\pm\text{0.03}}$ & $-$ \\

TabSDS & $-$ & $-$ & $-$ & $-$ & $-$ & $-$ & $-$ & $-$ \\
\bottomrule

\end{tabular}
}
\end{table}

\begin{table}[!htbp]
\centering
\caption{\textbf{Raw benchmark results of 23 tabular generators on the ``Coupon'' dataset.} We report the mean $\pm$ std of each metric across 10 repeated data splits. For benchmark generators, ``$-$'' denotes failed convergence of a specific model or unexpected values in the synthetic data that caused the evaluation metric computation to crash. We highlight the {\color[HTML]{008080} \textbf{First}}, {\color[HTML]{7030A0} \textbf{Second}}, and {\color[HTML]{C65911} \textbf{Third}} best performances for each metric. \modelname generally achieves competitive performance against the benchmark generators while maintaining a reduced risk of overfitting.}
\resizebox{\textwidth}{!}{
\begin{tabular}{l|rrrr|rr|r|r}

\toprule

\multirow{2}{*}{\textbf{Generator}} & \multicolumn{4}{c|}{\textbf{Density   Estimation}}                             & \multicolumn{2}{c|}{\textbf{Privacy Preservation}} & \multicolumn{1}{c|}{\textbf{ML Efficacy}}                              & \multicolumn{1}{c}{\textbf{Structural Fidelity}} \\

& Shape $\uparrow$ & Trend $\uparrow$ & $\alpha$-precision $\uparrow$ & $\beta$-recall $\uparrow$ & Authenticity $\uparrow$ & DCR Score $\uparrow$ & Local utility $\uparrow$ & Global utility $\uparrow$  \\

\midrule

\rowcolor{Gainsboro!60}

\multicolumn{9}{c}{\textbf{Real Data}} \\

\midrule

Real Data (Train) & 1.00$_{\pm\text{0.00}}$ & 1.00$_{\pm\text{0.00}}$ & 1.00$_{\pm\text{0.00}}$ & 1.00$_{\pm\text{0.00}}$ & 0.00$_{\pm\text{0.00}}$ & 0.00$_{\pm\text{0.00}}$ & 1.00$_{\pm\text{0.00}}$ & 0.95$_{\pm\text{0.02}}$ \\

Real Data (Holdout) & 0.99$_{\pm\text{0.00}}$ & 0.98$_{\pm\text{0.00}}$ & 0.99$_{\pm\text{0.01}}$ & 0.52$_{\pm\text{0.01}}$ & 0.61$_{\pm\text{0.01}}$ & 0.20$_{\pm\text{0.00}}$ & $-$ & $-$ \\

\midrule

\rowcolor{Gainsboro!60}

\multicolumn{9}{c}{\textbf{Dataset-specific Model}} \\

\midrule

SMOTE & {\color[HTML]{008080} \textbf{0.93}}$_{\pm\text{0.00}}$ & 0.79$_{\pm\text{0.01}}$ & 0.68$_{\pm\text{0.01}}$ & {\color[HTML]{008080} \textbf{0.64}}$_{\pm\text{0.01}}$ & 0.44$_{\pm\text{0.01}}$ & 0.12$_{\pm\text{0.00}}$ & {\color[HTML]{008080} \textbf{0.96}}$_{\pm\text{0.01}}$ & {\color[HTML]{C65911} \textbf{0.57}}$_{\pm\text{0.02}}$ \\

TabSDS & $-$ & $-$ & $-$ & $-$ & $-$ & $-$ & $-$ & $-$ \\
\bottomrule

\end{tabular}
}
\end{table}

\begin{table}[!htbp]
\centering
\caption{\textbf{Raw benchmark results of 23 tabular generators on the ``Credit'' dataset.} We report the mean $\pm$ std of each metric across 10 repeated data splits. For benchmark generators, ``$-$'' denotes failed convergence of a specific model or unexpected values in the synthetic data that caused the evaluation metric computation to crash. We highlight the {\color[HTML]{008080} \textbf{First}}, {\color[HTML]{7030A0} \textbf{Second}}, and {\color[HTML]{C65911} \textbf{Third}} best performances for each metric. \modelname generally achieves competitive performance against the benchmark generators while maintaining a reduced risk of overfitting.}
\resizebox{\textwidth}{!}{
\begin{tabular}{l|rrrr|rr|r|r}

\toprule

\multirow{2}{*}{\textbf{Generator}} & \multicolumn{4}{c|}{\textbf{Density   Estimation}}                             & \multicolumn{2}{c|}{\textbf{Privacy Preservation}} & \multicolumn{1}{c|}{\textbf{ML Efficacy}}                              & \multicolumn{1}{c}{\textbf{Structural Fidelity}} \\

& Shape $\uparrow$ & Trend $\uparrow$ & $\alpha$-precision $\uparrow$ & $\beta$-recall $\uparrow$ & Authenticity $\uparrow$ & DCR Score $\uparrow$ & Local utility $\uparrow$ & Global utility $\uparrow$  \\

\midrule

\rowcolor{Gainsboro!60}

\multicolumn{9}{c}{\textbf{Real Data}} \\

\midrule

Real Data (Train) & 1.00$_{\pm\text{0.00}}$ & 1.00$_{\pm\text{0.00}}$ & 1.00$_{\pm\text{0.00}}$ & 1.00$_{\pm\text{0.00}}$ & 0.00$_{\pm\text{0.00}}$ & 0.00$_{\pm\text{0.00}}$ & 1.00$_{\pm\text{0.00}}$ & 0.50$_{\pm\text{0.03}}$ \\

Real Data (Holdout) & 0.96$_{\pm\text{0.01}}$ & 0.92$_{\pm\text{0.03}}$ & 0.95$_{\pm\text{0.02}}$ & 0.48$_{\pm\text{0.04}}$ & 0.75$_{\pm\text{0.02}}$ & 0.26$_{\pm\text{0.01}}$ & $-$ & $-$ \\

\midrule

\rowcolor{Gainsboro!60}

\multicolumn{9}{c}{\textbf{Dataset-specific Model}} \\

\midrule

SMOTE & {\color[HTML]{C65911} \textbf{0.94}}$_{\pm\text{0.01}}$ & 0.75$_{\pm\text{0.01}}$ & 0.80$_{\pm\text{0.03}}$ & {\color[HTML]{008080} \textbf{0.62}}$_{\pm\text{0.03}}$ & 0.50$_{\pm\text{0.01}}$ & 0.02$_{\pm\text{0.00}}$ & 0.97$_{\pm\text{0.04}}$ & 0.31$_{\pm\text{0.02}}$ \\

TabSDS & 0.74$_{\pm\text{0.00}}$ & 0.62$_{\pm\text{0.02}}$ & 0.59$_{\pm\text{0.02}}$ & 0.23$_{\pm\text{0.01}}$ & 0.56$_{\pm\text{0.01}}$ & 0.26$_{\pm\text{0.01}}$ & 0.68$_{\pm\text{0.02}}$ & 0.26$_{\pm\text{0.02}}$ \\
\bottomrule

\end{tabular}
}
\end{table}

\begin{table}[!htbp]
\centering
\caption{\textbf{Raw benchmark results of 23 tabular generators on the ``Customer'' dataset.} We report the mean $\pm$ std of each metric across 10 repeated data splits. For benchmark generators, ``$-$'' denotes failed convergence of a specific model or unexpected values in the synthetic data that caused the evaluation metric computation to crash. We highlight the {\color[HTML]{008080} \textbf{First}}, {\color[HTML]{7030A0} \textbf{Second}}, and {\color[HTML]{C65911} \textbf{Third}} best performances for each metric. \modelname generally achieves competitive performance against the benchmark generators while maintaining a reduced risk of overfitting.}
\resizebox{\textwidth}{!}{
\begin{tabular}{l|rrrr|rr|r|r}

\toprule

\multirow{2}{*}{\textbf{Generator}} & \multicolumn{4}{c|}{\textbf{Density   Estimation}}                             & \multicolumn{2}{c|}{\textbf{Privacy Preservation}} & \multicolumn{1}{c|}{\textbf{ML Efficacy}}                              & \multicolumn{1}{c}{\textbf{Structural Fidelity}} \\

& Shape $\uparrow$ & Trend $\uparrow$ & $\alpha$-precision $\uparrow$ & $\beta$-recall $\uparrow$ & Authenticity $\uparrow$ & DCR Score $\uparrow$ & Local utility $\uparrow$ & Global utility $\uparrow$  \\

\midrule

\rowcolor{Gainsboro!60}

\multicolumn{9}{c}{\textbf{Real Data}} \\

\midrule

Real Data (Train) & 1.00$_{\pm\text{0.00}}$ & 1.00$_{\pm\text{0.00}}$ & 1.00$_{\pm\text{0.00}}$ & 1.00$_{\pm\text{0.00}}$ & 0.00$_{\pm\text{0.00}}$ & 0.00$_{\pm\text{0.00}}$ & 1.00$_{\pm\text{0.00}}$ & 0.66$_{\pm\text{0.03}}$ \\

Real Data (Holdout) & 0.97$_{\pm\text{0.00}}$ & 0.87$_{\pm\text{0.04}}$ & 0.96$_{\pm\text{0.02}}$ & 0.49$_{\pm\text{0.03}}$ & 0.71$_{\pm\text{0.02}}$ & 0.21$_{\pm\text{0.01}}$ & $-$ & $-$ \\

\midrule

\rowcolor{Gainsboro!60}

\multicolumn{9}{c}{\textbf{Dataset-specific Model}} \\

\midrule

SMOTE & {\color[HTML]{008080} \textbf{0.94}}$_{\pm\text{0.01}}$ & 0.68$_{\pm\text{0.03}}$ & 0.74$_{\pm\text{0.02}}$ & {\color[HTML]{008080} \textbf{0.63}}$_{\pm\text{0.01}}$ & 0.46$_{\pm\text{0.02}}$ & 0.02$_{\pm\text{0.00}}$ & {\color[HTML]{008080} \textbf{1.01}}$_{\pm\text{0.03}}$ & {\color[HTML]{7030A0} \textbf{0.54}}$_{\pm\text{0.03}}$ \\

TabSDS & $-$ & $-$ & $-$ & $-$ & $-$ & $-$ & $-$ & $-$ \\
\bottomrule

\end{tabular}
}
\end{table}

\begin{table}[!htbp]
\centering
\caption{\textbf{Raw benchmark results of 23 tabular generators on the ``Diabetes'' dataset.} We report the mean $\pm$ std of each metric across 10 repeated data splits. For benchmark generators, ``$-$'' denotes failed convergence of a specific model or unexpected values in the synthetic data that caused the evaluation metric computation to crash. We highlight the {\color[HTML]{008080} \textbf{First}}, {\color[HTML]{7030A0} \textbf{Second}}, and {\color[HTML]{C65911} \textbf{Third}} best performances for each metric. \modelname generally achieves competitive performance against the benchmark generators while maintaining a reduced risk of overfitting.}
\resizebox{\textwidth}{!}{
\begin{tabular}{l|rrrr|rr|r|r}

\toprule

\multirow{2}{*}{\textbf{Generator}} & \multicolumn{4}{c|}{\textbf{Density   Estimation}}                             & \multicolumn{2}{c|}{\textbf{Privacy Preservation}} & \multicolumn{1}{c|}{\textbf{ML Efficacy}}                              & \multicolumn{1}{c}{\textbf{Structural Fidelity}} \\

& Shape $\uparrow$ & Trend $\uparrow$ & $\alpha$-precision $\uparrow$ & $\beta$-recall $\uparrow$ & Authenticity $\uparrow$ & DCR Score $\uparrow$ & Local utility $\uparrow$ & Global utility $\uparrow$  \\

\midrule

\rowcolor{Gainsboro!60}

\multicolumn{9}{c}{\textbf{Real Data}} \\

\midrule

Real Data (Train) & 1.00$_{\pm\text{0.00}}$ & 1.00$_{\pm\text{0.00}}$ & 1.00$_{\pm\text{0.00}}$ & 1.00$_{\pm\text{0.00}}$ & 0.00$_{\pm\text{0.00}}$ & 0.00$_{\pm\text{0.00}}$ & 1.00$_{\pm\text{0.00}}$ & 0.98$_{\pm\text{0.01}}$ \\

Real Data (Holdout) & 0.94$_{\pm\text{0.01}}$ & 0.86$_{\pm\text{0.03}}$ & 0.95$_{\pm\text{0.02}}$ & 0.50$_{\pm\text{0.04}}$ & 0.70$_{\pm\text{0.04}}$ & 0.17$_{\pm\text{0.01}}$ & $-$ & $-$ \\

\midrule

\rowcolor{Gainsboro!60}

\multicolumn{9}{c}{\textbf{Dataset-specific Model}} \\

\midrule

SMOTE & {\color[HTML]{C65911} \textbf{0.86}}$_{\pm\text{0.01}}$ & 0.59$_{\pm\text{0.04}}$ & 0.81$_{\pm\text{0.02}}$ & {\color[HTML]{008080} \textbf{0.66}}$_{\pm\text{0.04}}$ & 0.43$_{\pm\text{0.03}}$ & 0.08$_{\pm\text{0.01}}$ & {\color[HTML]{7030A0} \textbf{0.98}}$_{\pm\text{0.04}}$ & 0.71$_{\pm\text{0.02}}$ \\

TabSDS & 0.67$_{\pm\text{0.01}}$ & 0.55$_{\pm\text{0.03}}$ & 0.64$_{\pm\text{0.02}}$ & 0.29$_{\pm\text{0.01}}$ & 0.55$_{\pm\text{0.01}}$ & 0.20$_{\pm\text{0.02}}$ & 0.71$_{\pm\text{0.02}}$ & 0.67$_{\pm\text{0.03}}$ \\
\bottomrule

\end{tabular}
}
\end{table}

\begin{table}[!htbp]
\centering
\caption{\textbf{Raw benchmark results of 23 tabular generators on the ``Diamonds'' dataset.} We report the mean $\pm$ std of each metric across 10 repeated data splits. For benchmark generators, ``$-$'' denotes failed convergence of a specific model or unexpected values in the synthetic data that caused the evaluation metric computation to crash. We highlight the {\color[HTML]{008080} \textbf{First}}, {\color[HTML]{7030A0} \textbf{Second}}, and {\color[HTML]{C65911} \textbf{Third}} best performances for each metric. \modelname generally achieves competitive performance against the benchmark generators while maintaining a reduced risk of overfitting.}
\resizebox{\textwidth}{!}{
\begin{tabular}{l|rrrr|rr|r|r}

\toprule

\multirow{2}{*}{\textbf{Generator}} & \multicolumn{4}{c|}{\textbf{Density   Estimation}}                             & \multicolumn{2}{c|}{\textbf{Privacy Preservation}} & \multicolumn{1}{c|}{\textbf{ML Efficacy}}                              & \multicolumn{1}{c}{\textbf{Structural Fidelity}} \\

& Shape $\uparrow$ & Trend $\uparrow$ & $\alpha$-precision $\uparrow$ & $\beta$-recall $\uparrow$ & Authenticity $\uparrow$ & DCR Score $\uparrow$ & Local utility $\uparrow$ & Global utility $\uparrow$  \\

\midrule

\rowcolor{Gainsboro!60}

\multicolumn{9}{c}{\textbf{Real Data}} \\

\midrule

Real Data (Train) & 1.00$_{\pm\text{0.00}}$ & 1.00$_{\pm\text{0.00}}$ & 1.00$_{\pm\text{0.00}}$ & 1.00$_{\pm\text{0.00}}$ & 0.00$_{\pm\text{0.00}}$ & 0.00$_{\pm\text{0.00}}$ & 1.00$_{\pm\text{0.00}}$ & 0.99$_{\pm\text{0.02}}$ \\

Real Data (Holdout) & 0.99$_{\pm\text{0.00}}$ & 0.91$_{\pm\text{0.04}}$ & 0.99$_{\pm\text{0.00}}$ & 0.50$_{\pm\text{0.00}}$ & 0.66$_{\pm\text{0.01}}$ & 0.03$_{\pm\text{0.00}}$ & $-$ & $-$ \\

\midrule

\rowcolor{Gainsboro!60}

\multicolumn{9}{c}{\textbf{Dataset-specific Model}} \\

\midrule

SMOTE & {\color[HTML]{7030A0} \textbf{0.98}}$_{\pm\text{0.00}}$ & $-$ & $-$ & {\color[HTML]{008080} \textbf{0.61}}$_{\pm\text{0.01}}$ & 0.50$_{\pm\text{0.01}}$ & 0.04$_{\pm\text{0.00}}$ & {\color[HTML]{008080} \textbf{0.87}}$_{\pm\text{0.01}}$ & $-$ \\

TabSDS & 0.70$_{\pm\text{0.01}}$ & 0.64$_{\pm\text{0.02}}$ & 0.64$_{\pm\text{0.00}}$ & 0.18$_{\pm\text{0.01}}$ & 0.66$_{\pm\text{0.00}}$ & 0.04$_{\pm\text{0.01}}$ & 0.49$_{\pm\text{0.01}}$ & 0.41$_{\pm\text{0.01}}$ \\
\bottomrule

\end{tabular}
}
\end{table}

\begin{table}[!htbp]
\centering
\caption{\textbf{Raw benchmark results of 23 tabular generators on the ``Fiat500'' dataset.} We report the mean $\pm$ std of each metric across 10 repeated data splits. For benchmark generators, ``$-$'' denotes failed convergence of a specific model or unexpected values in the synthetic data that caused the evaluation metric computation to crash. We highlight the {\color[HTML]{008080} \textbf{First}}, {\color[HTML]{7030A0} \textbf{Second}}, and {\color[HTML]{C65911} \textbf{Third}} best performances for each metric. \modelname generally achieves competitive performance against the benchmark generators while maintaining a reduced risk of overfitting.}
\resizebox{\textwidth}{!}{
\begin{tabular}{l|rrrr|rr|r|r}

\toprule

\multirow{2}{*}{\textbf{Generator}} & \multicolumn{4}{c|}{\textbf{Density   Estimation}}                             & \multicolumn{2}{c|}{\textbf{Privacy Preservation}} & \multicolumn{1}{c|}{\textbf{ML Efficacy}}                              & \multicolumn{1}{c}{\textbf{Structural Fidelity}} \\

& Shape $\uparrow$ & Trend $\uparrow$ & $\alpha$-precision $\uparrow$ & $\beta$-recall $\uparrow$ & Authenticity $\uparrow$ & DCR Score $\uparrow$ & Local utility $\uparrow$ & Global utility $\uparrow$  \\

\midrule

\rowcolor{Gainsboro!60}

\multicolumn{9}{c}{\textbf{Real Data}} \\

\midrule

Real Data (Train) & 1.00$_{\pm\text{0.00}}$ & 1.00$_{\pm\text{0.00}}$ & 1.00$_{\pm\text{0.00}}$ & 1.00$_{\pm\text{0.00}}$ & 0.00$_{\pm\text{0.00}}$ & 0.00$_{\pm\text{0.00}}$ & 1.00$_{\pm\text{0.00}}$ & 0.89$_{\pm\text{0.01}}$ \\

Real Data (Holdout) & 0.96$_{\pm\text{0.01}}$ & 0.96$_{\pm\text{0.02}}$ & 0.97$_{\pm\text{0.01}}$ & 0.49$_{\pm\text{0.02}}$ & 0.68$_{\pm\text{0.03}}$ & 0.06$_{\pm\text{0.00}}$ & $-$ & $-$ \\

\midrule

\rowcolor{Gainsboro!60}

\multicolumn{9}{c}{\textbf{Dataset-specific Model}} \\

\midrule

SMOTE & {\color[HTML]{7030A0} \textbf{0.94}}$_{\pm\text{0.01}}$ & 0.75$_{\pm\text{0.02}}$ & 0.79$_{\pm\text{0.02}}$ & {\color[HTML]{008080} \textbf{0.59}}$_{\pm\text{0.02}}$ & 0.52$_{\pm\text{0.01}}$ & 0.03$_{\pm\text{0.00}}$ & {\color[HTML]{008080} \textbf{0.93}}$_{\pm\text{0.02}}$ & {\color[HTML]{C65911} \textbf{0.70}}$_{\pm\text{0.01}}$ \\

TabSDS & 0.67$_{\pm\text{0.00}}$ & 0.61$_{\pm\text{0.03}}$ & 0.68$_{\pm\text{0.01}}$ & 0.16$_{\pm\text{0.01}}$ & 0.63$_{\pm\text{0.01}}$ & 0.12$_{\pm\text{0.01}}$ & 0.69$_{\pm\text{0.02}}$ & 0.61$_{\pm\text{0.01}}$ \\
\bottomrule

\end{tabular}
}
\end{table}

\begin{table}[!htbp]
\centering
\caption{\textbf{Raw benchmark results of 23 tabular generators on the ``Fish'' dataset.} We report the mean $\pm$ std of each metric across 10 repeated data splits. For benchmark generators, ``$-$'' denotes failed convergence of a specific model or unexpected values in the synthetic data that caused the evaluation metric computation to crash. We highlight the {\color[HTML]{008080} \textbf{First}}, {\color[HTML]{7030A0} \textbf{Second}}, and {\color[HTML]{C65911} \textbf{Third}} best performances for each metric. \modelname generally achieves competitive performance against the benchmark generators while maintaining a reduced risk of overfitting.}
\resizebox{\textwidth}{!}{
\begin{tabular}{l|rrrr|rr|r|r}

\toprule

\multirow{2}{*}{\textbf{Generator}} & \multicolumn{4}{c|}{\textbf{Density   Estimation}}                             & \multicolumn{2}{c|}{\textbf{Privacy Preservation}} & \multicolumn{1}{c|}{\textbf{ML Efficacy}}                              & \multicolumn{1}{c}{\textbf{Structural Fidelity}} \\

& Shape $\uparrow$ & Trend $\uparrow$ & $\alpha$-precision $\uparrow$ & $\beta$-recall $\uparrow$ & Authenticity $\uparrow$ & DCR Score $\uparrow$ & Local utility $\uparrow$ & Global utility $\uparrow$  \\

\midrule

\rowcolor{Gainsboro!60}

\multicolumn{9}{c}{\textbf{Real Data}} \\

\midrule

Real Data (Train) & 1.00$_{\pm\text{0.00}}$ & 1.00$_{\pm\text{0.00}}$ & 1.00$_{\pm\text{0.00}}$ & 1.00$_{\pm\text{0.00}}$ & 0.00$_{\pm\text{0.00}}$ & 0.00$_{\pm\text{0.00}}$ & 1.00$_{\pm\text{0.00}}$ & 1.00$_{\pm\text{0.01}}$ \\

Real Data (Holdout) & 0.94$_{\pm\text{0.01}}$ & 0.98$_{\pm\text{0.02}}$ & 0.95$_{\pm\text{0.03}}$ & 0.55$_{\pm\text{0.05}}$ & 0.63$_{\pm\text{0.05}}$ & 0.09$_{\pm\text{0.01}}$ & $-$ & $-$ \\

\midrule

\rowcolor{Gainsboro!60}

\multicolumn{9}{c}{\textbf{Dataset-specific Model}} \\

\midrule

SMOTE & 0.74$_{\pm\text{0.03}}$ & 0.80$_{\pm\text{0.02}}$ & 0.81$_{\pm\text{0.03}}$ & {\color[HTML]{008080} \textbf{0.52}}$_{\pm\text{0.03}}$ & 0.59$_{\pm\text{0.02}}$ & 0.05$_{\pm\text{0.01}}$ & {\color[HTML]{008080} \textbf{0.93}}$_{\pm\text{0.02}}$ & 0.69$_{\pm\text{0.02}}$ \\

TabSDS & 0.49$_{\pm\text{0.02}}$ & 0.71$_{\pm\text{0.01}}$ & 0.69$_{\pm\text{0.01}}$ & 0.21$_{\pm\text{0.01}}$ & 0.65$_{\pm\text{0.02}}$ & 0.13$_{\pm\text{0.01}}$ & 0.68$_{\pm\text{0.02}}$ & 0.58$_{\pm\text{0.02}}$ \\
\bottomrule

\end{tabular}
}
\end{table}

\begin{table}[!htbp]
\centering
\caption{\textbf{Raw benchmark results of 23 tabular generators on the ``Fitness'' dataset.} We report the mean $\pm$ std of each metric across 10 repeated data splits. For benchmark generators, ``$-$'' denotes failed convergence of a specific model or unexpected values in the synthetic data that caused the evaluation metric computation to crash. We highlight the {\color[HTML]{008080} \textbf{First}}, {\color[HTML]{7030A0} \textbf{Second}}, and {\color[HTML]{C65911} \textbf{Third}} best performances for each metric. \modelname generally achieves competitive performance against the benchmark generators while maintaining a reduced risk of overfitting.}
\resizebox{\textwidth}{!}{
\begin{tabular}{l|rrrr|rr|r|r}

\toprule

\multirow{2}{*}{\textbf{Generator}} & \multicolumn{4}{c|}{\textbf{Density   Estimation}}                             & \multicolumn{2}{c|}{\textbf{Privacy Preservation}} & \multicolumn{1}{c|}{\textbf{ML Efficacy}}                              & \multicolumn{1}{c}{\textbf{Structural Fidelity}} \\

& Shape $\uparrow$ & Trend $\uparrow$ & $\alpha$-precision $\uparrow$ & $\beta$-recall $\uparrow$ & Authenticity $\uparrow$ & DCR Score $\uparrow$ & Local utility $\uparrow$ & Global utility $\uparrow$  \\

\midrule

\rowcolor{Gainsboro!60}

\multicolumn{9}{c}{\textbf{Real Data}} \\

\midrule

Real Data (Train) & 1.00$_{\pm\text{0.00}}$ & 1.00$_{\pm\text{0.00}}$ & 1.00$_{\pm\text{0.00}}$ & 1.00$_{\pm\text{0.00}}$ & 0.00$_{\pm\text{0.00}}$ & 0.00$_{\pm\text{0.00}}$ & 1.00$_{\pm\text{0.00}}$ & 0.76$_{\pm\text{0.07}}$ \\

Real Data (Holdout) & 0.96$_{\pm\text{0.01}}$ & 0.66$_{\pm\text{0.05}}$ & 0.97$_{\pm\text{0.01}}$ & 0.49$_{\pm\text{0.03}}$ & 0.66$_{\pm\text{0.02}}$ & 0.04$_{\pm\text{0.00}}$ & $-$ & $-$ \\

\midrule

\rowcolor{Gainsboro!60}

\multicolumn{9}{c}{\textbf{Dataset-specific Model}} \\

\midrule

SMOTE & {\color[HTML]{7030A0} \textbf{0.94}}$_{\pm\text{0.01}}$ & 0.64$_{\pm\text{0.04}}$ & 0.83$_{\pm\text{0.02}}$ & {\color[HTML]{008080} \textbf{0.60}}$_{\pm\text{0.02}}$ & 0.49$_{\pm\text{0.01}}$ & 0.02$_{\pm\text{0.00}}$ & {\color[HTML]{7030A0} \textbf{0.99}}$_{\pm\text{0.03}}$ & 0.50$_{\pm\text{0.06}}$ \\

TabSDS & 0.73$_{\pm\text{0.00}}$ & 0.43$_{\pm\text{0.04}}$ & 0.69$_{\pm\text{0.01}}$ & 0.25$_{\pm\text{0.01}}$ & 0.58$_{\pm\text{0.01}}$ & 0.05$_{\pm\text{0.01}}$ & 0.73$_{\pm\text{0.02}}$ & 0.47$_{\pm\text{0.05}}$ \\
\bottomrule

\end{tabular}
}
\end{table}

\begin{table}[!htbp]
\centering
\caption{\textbf{Raw benchmark results of 23 tabular generators on the ``FoodDelivery'' dataset.} We report the mean $\pm$ std of each metric across 10 repeated data splits. For benchmark generators, ``$-$'' denotes failed convergence of a specific model or unexpected values in the synthetic data that caused the evaluation metric computation to crash. We highlight the {\color[HTML]{008080} \textbf{First}}, {\color[HTML]{7030A0} \textbf{Second}}, and {\color[HTML]{C65911} \textbf{Third}} best performances for each metric. \modelname generally achieves competitive performance against the benchmark generators while maintaining a reduced risk of overfitting.}
\resizebox{\textwidth}{!}{
\begin{tabular}{l|rrrr|rr|r|r}

\toprule

\multirow{2}{*}{\textbf{Generator}} & \multicolumn{4}{c|}{\textbf{Density   Estimation}}                             & \multicolumn{2}{c|}{\textbf{Privacy Preservation}} & \multicolumn{1}{c|}{\textbf{ML Efficacy}}                              & \multicolumn{1}{c}{\textbf{Structural Fidelity}} \\

& Shape $\uparrow$ & Trend $\uparrow$ & $\alpha$-precision $\uparrow$ & $\beta$-recall $\uparrow$ & Authenticity $\uparrow$ & DCR Score $\uparrow$ & Local utility $\uparrow$ & Global utility $\uparrow$  \\

\midrule

\rowcolor{Gainsboro!60}

\multicolumn{9}{c}{\textbf{Real Data}} \\

\midrule

Real Data (Train) & 1.00$_{\pm\text{0.00}}$ & 1.00$_{\pm\text{0.00}}$ & 1.00$_{\pm\text{0.00}}$ & 1.00$_{\pm\text{0.00}}$ & 0.00$_{\pm\text{0.00}}$ & 0.00$_{\pm\text{0.00}}$ & 1.00$_{\pm\text{0.00}}$ & 0.71$_{\pm\text{0.00}}$ \\

Real Data (Holdout) & 0.98$_{\pm\text{0.00}}$ & 0.94$_{\pm\text{0.00}}$ & 0.99$_{\pm\text{0.00}}$ & 0.49$_{\pm\text{0.00}}$ & 0.63$_{\pm\text{0.01}}$ & 0.25$_{\pm\text{0.00}}$ & $-$ & $-$ \\

\midrule

\rowcolor{Gainsboro!60}

\multicolumn{9}{c}{\textbf{Dataset-specific Model}} \\

\midrule

SMOTE & {\color[HTML]{7030A0} \textbf{0.95}}$_{\pm\text{0.00}}$ & $-$ & $-$ & {\color[HTML]{008080} \textbf{0.61}}$_{\pm\text{0.00}}$ & 0.44$_{\pm\text{0.00}}$ & 0.41$_{\pm\text{0.00}}$ & {\color[HTML]{C65911} \textbf{0.94}}$_{\pm\text{0.00}}$ & $-$ \\

TabSDS & $-$ & $-$ & $-$ & $-$ & $-$ & $-$ & $-$ & $-$ \\
\bottomrule

\end{tabular}
}
\end{table}

\begin{table}[!htbp]
\centering
\caption{\textbf{Raw benchmark results of 23 tabular generators on the ``Give'' dataset.} We report the mean $\pm$ std of each metric across 10 repeated data splits. For benchmark generators, ``$-$'' denotes failed convergence of a specific model or unexpected values in the synthetic data that caused the evaluation metric computation to crash. We highlight the {\color[HTML]{008080} \textbf{First}}, {\color[HTML]{7030A0} \textbf{Second}}, and {\color[HTML]{C65911} \textbf{Third}} best performances for each metric. \modelname generally achieves competitive performance against the benchmark generators while maintaining a reduced risk of overfitting.}
\resizebox{\textwidth}{!}{
\begin{tabular}{l|rrrr|rr|r|r}

\toprule

\multirow{2}{*}{\textbf{Generator}} & \multicolumn{4}{c|}{\textbf{Density   Estimation}}                             & \multicolumn{2}{c|}{\textbf{Privacy Preservation}} & \multicolumn{1}{c|}{\textbf{ML Efficacy}}                              & \multicolumn{1}{c}{\textbf{Structural Fidelity}} \\

& Shape $\uparrow$ & Trend $\uparrow$ & $\alpha$-precision $\uparrow$ & $\beta$-recall $\uparrow$ & Authenticity $\uparrow$ & DCR Score $\uparrow$ & Local utility $\uparrow$ & Global utility $\uparrow$  \\

\midrule

\rowcolor{Gainsboro!60}

\multicolumn{9}{c}{\textbf{Real Data}} \\

\midrule

Real Data (Train) & 1.00$_{\pm\text{0.00}}$ & 1.00$_{\pm\text{0.00}}$ & 1.00$_{\pm\text{0.00}}$ & 1.00$_{\pm\text{0.00}}$ & 0.00$_{\pm\text{0.00}}$ & 0.00$_{\pm\text{0.00}}$ & 1.00$_{\pm\text{0.00}}$ & 0.95$_{\pm\text{0.01}}$ \\

Real Data (Holdout) & 1.00$_{\pm\text{0.00}}$ & 1.00$_{\pm\text{0.00}}$ & 1.00$_{\pm\text{0.00}}$ & 0.50$_{\pm\text{0.00}}$ & 0.67$_{\pm\text{0.00}}$ & 0.00$_{\pm\text{0.00}}$ & $-$ & $-$ \\

\midrule

\rowcolor{Gainsboro!60}

\multicolumn{9}{c}{\textbf{Dataset-specific Model}} \\

\midrule

SMOTE & 0.64$_{\pm\text{0.00}}$ & $-$ & $-$ & {\color[HTML]{7030A0} \textbf{0.38}}$_{\pm\text{0.01}}$ & 0.71$_{\pm\text{0.01}}$ & 0.01$_{\pm\text{0.00}}$ & {\color[HTML]{008080} \textbf{1.04}}$_{\pm\text{0.01}}$ & $-$ \\

TabSDS & 0.54$_{\pm\text{0.00}}$ & 0.47$_{\pm\text{0.02}}$ & 0.73$_{\pm\text{0.01}}$ & 0.13$_{\pm\text{0.01}}$ & 0.63$_{\pm\text{0.00}}$ & 0.01$_{\pm\text{0.00}}$ & 0.70$_{\pm\text{0.01}}$ & 0.48$_{\pm\text{0.03}}$ \\
\bottomrule

\end{tabular}
}
\end{table}

\begin{table}[!htbp]
\centering
\caption{\textbf{Raw benchmark results of 23 tabular generators on the ``HELOC'' dataset.} We report the mean $\pm$ std of each metric across 10 repeated data splits. For benchmark generators, ``$-$'' denotes failed convergence of a specific model or unexpected values in the synthetic data that caused the evaluation metric computation to crash. We highlight the {\color[HTML]{008080} \textbf{First}}, {\color[HTML]{7030A0} \textbf{Second}}, and {\color[HTML]{C65911} \textbf{Third}} best performances for each metric. \modelname generally achieves competitive performance against the benchmark generators while maintaining a reduced risk of overfitting.}
\resizebox{\textwidth}{!}{
\begin{tabular}{l|rrrr|rr|r|r}

\toprule

\multirow{2}{*}{\textbf{Generator}} & \multicolumn{4}{c|}{\textbf{Density   Estimation}}                             & \multicolumn{2}{c|}{\textbf{Privacy Preservation}} & \multicolumn{1}{c|}{\textbf{ML Efficacy}}                              & \multicolumn{1}{c}{\textbf{Structural Fidelity}} \\

& Shape $\uparrow$ & Trend $\uparrow$ & $\alpha$-precision $\uparrow$ & $\beta$-recall $\uparrow$ & Authenticity $\uparrow$ & DCR Score $\uparrow$ & Local utility $\uparrow$ & Global utility $\uparrow$  \\

\midrule

\rowcolor{Gainsboro!60}

\multicolumn{9}{c}{\textbf{Real Data}} \\

\midrule

Real Data (Train) & 1.00$_{\pm\text{0.00}}$ & 1.00$_{\pm\text{0.00}}$ & 1.00$_{\pm\text{0.00}}$ & 1.00$_{\pm\text{0.00}}$ & 0.00$_{\pm\text{0.00}}$ & 0.00$_{\pm\text{0.00}}$ & 1.00$_{\pm\text{0.00}}$ & 1.00$_{\pm\text{0.00}}$ \\

Real Data (Holdout) & 0.99$_{\pm\text{0.00}}$ & 0.99$_{\pm\text{0.00}}$ & 0.99$_{\pm\text{0.01}}$ & 0.49$_{\pm\text{0.01}}$ & 0.71$_{\pm\text{0.01}}$ & 0.08$_{\pm\text{0.00}}$ & $-$ & $-$ \\

\midrule

\rowcolor{Gainsboro!60}

\multicolumn{9}{c}{\textbf{Dataset-specific Model}} \\

\midrule

SMOTE & 0.82$_{\pm\text{0.01}}$ & 0.84$_{\pm\text{0.00}}$ & 0.77$_{\pm\text{0.01}}$ & {\color[HTML]{008080} \textbf{0.72}}$_{\pm\text{0.01}}$ & 0.43$_{\pm\text{0.01}}$ & 0.05$_{\pm\text{0.00}}$ & {\color[HTML]{008080} \textbf{0.98}}$_{\pm\text{0.01}}$ & 0.60$_{\pm\text{0.01}}$ \\

TabSDS & 0.65$_{\pm\text{0.00}}$ & 0.71$_{\pm\text{0.00}}$ & 0.66$_{\pm\text{0.01}}$ & 0.20$_{\pm\text{0.01}}$ & 0.61$_{\pm\text{0.00}}$ & 0.12$_{\pm\text{0.01}}$ & 0.68$_{\pm\text{0.00}}$ & 0.52$_{\pm\text{0.01}}$ \\
\bottomrule

\end{tabular}
}
\end{table}

\begin{table}[!htbp]
\centering
\caption{\textbf{Raw benchmark results of 23 tabular generators on the ``HR'' dataset.} We report the mean $\pm$ std of each metric across 10 repeated data splits. For benchmark generators, ``$-$'' denotes failed convergence of a specific model or unexpected values in the synthetic data that caused the evaluation metric computation to crash. We highlight the {\color[HTML]{008080} \textbf{First}}, {\color[HTML]{7030A0} \textbf{Second}}, and {\color[HTML]{C65911} \textbf{Third}} best performances for each metric. \modelname generally achieves competitive performance against the benchmark generators while maintaining a reduced risk of overfitting.}
\resizebox{\textwidth}{!}{
\begin{tabular}{l|rrrr|rr|r|r}

\toprule

\multirow{2}{*}{\textbf{Generator}} & \multicolumn{4}{c|}{\textbf{Density   Estimation}}                             & \multicolumn{2}{c|}{\textbf{Privacy Preservation}} & \multicolumn{1}{c|}{\textbf{ML Efficacy}}                              & \multicolumn{1}{c}{\textbf{Structural Fidelity}} \\

& Shape $\uparrow$ & Trend $\uparrow$ & $\alpha$-precision $\uparrow$ & $\beta$-recall $\uparrow$ & Authenticity $\uparrow$ & DCR Score $\uparrow$ & Local utility $\uparrow$ & Global utility $\uparrow$  \\

\midrule

\rowcolor{Gainsboro!60}

\multicolumn{9}{c}{\textbf{Real Data}} \\

\midrule

Real Data (Train) & 1.00$_{\pm\text{0.00}}$ & 1.00$_{\pm\text{0.00}}$ & 1.00$_{\pm\text{0.00}}$ & 1.00$_{\pm\text{0.00}}$ & 0.00$_{\pm\text{0.00}}$ & 0.00$_{\pm\text{0.00}}$ & 1.00$_{\pm\text{0.00}}$ & 0.72$_{\pm\text{0.09}}$ \\

Real Data (Holdout) & 0.99$_{\pm\text{0.00}}$ & 0.97$_{\pm\text{0.00}}$ & 0.99$_{\pm\text{0.01}}$ & 0.50$_{\pm\text{0.01}}$ & 0.77$_{\pm\text{0.00}}$ & 0.23$_{\pm\text{0.00}}$ & $-$ & $-$ \\

\midrule

\rowcolor{Gainsboro!60}

\multicolumn{9}{c}{\textbf{Dataset-specific Model}} \\

\midrule

SMOTE & {\color[HTML]{7030A0} \textbf{0.89}}$_{\pm\text{0.00}}$ & 0.67$_{\pm\text{0.00}}$ & 0.60$_{\pm\text{0.01}}$ & {\color[HTML]{008080} \textbf{0.49}}$_{\pm\text{0.01}}$ & 0.64$_{\pm\text{0.01}}$ & 0.37$_{\pm\text{0.00}}$ & {\color[HTML]{008080} \textbf{0.96}}$_{\pm\text{0.01}}$ & 0.33$_{\pm\text{0.05}}$ \\

TabSDS & $-$ & $-$ & $-$ & $-$ & $-$ & $-$ & $-$ & $-$ \\
\bottomrule

\end{tabular}
}
\end{table}

\begin{table}[!htbp]
\centering
\caption{\textbf{Raw benchmark results of 23 tabular generators on the ``Hazelnut'' dataset.} We report the mean $\pm$ std of each metric across 10 repeated data splits. For benchmark generators, ``$-$'' denotes failed convergence of a specific model or unexpected values in the synthetic data that caused the evaluation metric computation to crash. We highlight the {\color[HTML]{008080} \textbf{First}}, {\color[HTML]{7030A0} \textbf{Second}}, and {\color[HTML]{C65911} \textbf{Third}} best performances for each metric. \modelname generally achieves competitive performance against the benchmark generators while maintaining a reduced risk of overfitting.}
\resizebox{\textwidth}{!}{

}
\end{table}

\begin{table}[!htbp]
\centering
\caption{\textbf{Raw benchmark results of 23 tabular generators on the ``Healthcare'' dataset.} We report the mean $\pm$ std of each metric across 10 repeated data splits. For benchmark generators, ``$-$'' denotes failed convergence of a specific model or unexpected values in the synthetic data that caused the evaluation metric computation to crash. We highlight the {\color[HTML]{008080} \textbf{First}}, {\color[HTML]{7030A0} \textbf{Second}}, and {\color[HTML]{C65911} \textbf{Third}} best performances for each metric. \modelname generally achieves competitive performance against the benchmark generators while maintaining a reduced risk of overfitting.}
\resizebox{\textwidth}{!}{

}
\end{table}

\begin{table}[!htbp]
\centering
\caption{\textbf{Raw benchmark results of 23 tabular generators on the ``House16H'' dataset.} We report the mean $\pm$ std of each metric across 10 repeated data splits. For benchmark generators, ``$-$'' denotes failed convergence of a specific model or unexpected values in the synthetic data that caused the evaluation metric computation to crash. We highlight the {\color[HTML]{008080} \textbf{First}}, {\color[HTML]{7030A0} \textbf{Second}}, and {\color[HTML]{C65911} \textbf{Third}} best performances for each metric. \modelname generally achieves competitive performance against the benchmark generators while maintaining a reduced risk of overfitting.}
\resizebox{\textwidth}{!}{

}
\end{table}

\begin{table}[!htbp]
\centering
\caption{\textbf{Raw benchmark results of 23 tabular generators on the ``Houses'' dataset.} We report the mean $\pm$ std of each metric across 10 repeated data splits. For benchmark generators, ``$-$'' denotes failed convergence of a specific model or unexpected values in the synthetic data that caused the evaluation metric computation to crash. We highlight the {\color[HTML]{008080} \textbf{First}}, {\color[HTML]{7030A0} \textbf{Second}}, and {\color[HTML]{C65911} \textbf{Third}} best performances for each metric. \modelname generally achieves competitive performance against the benchmark generators while maintaining a reduced risk of overfitting.}
\resizebox{\textwidth}{!}{

}
\end{table}

\begin{table}[!htbp]
\centering
\caption{\textbf{Raw benchmark results of 23 tabular generators on the ``JM1'' dataset.} We report the mean $\pm$ std of each metric across 10 repeated data splits. For benchmark generators, ``$-$'' denotes failed convergence of a specific model or unexpected values in the synthetic data that caused the evaluation metric computation to crash. We highlight the {\color[HTML]{008080} \textbf{First}}, {\color[HTML]{7030A0} \textbf{Second}}, and {\color[HTML]{C65911} \textbf{Third}} best performances for each metric. \modelname generally achieves competitive performance against the benchmark generators while maintaining a reduced risk of overfitting.}
\resizebox{\textwidth}{!}{

}
\end{table}

\begin{table}[!htbp]
\centering
\caption{\textbf{Raw benchmark results of 23 tabular generators on the ``Marketing'' dataset.} We report the mean $\pm$ std of each metric across 10 repeated data splits. For benchmark generators, ``$-$'' denotes failed convergence of a specific model or unexpected values in the synthetic data that caused the evaluation metric computation to crash. We highlight the {\color[HTML]{008080} \textbf{First}}, {\color[HTML]{7030A0} \textbf{Second}}, and {\color[HTML]{C65911} \textbf{Third}} best performances for each metric. \modelname generally achieves competitive performance against the benchmark generators while maintaining a reduced risk of overfitting.}
\resizebox{\textwidth}{!}{

}
\end{table}

\begin{table}[!htbp]
\centering
\caption{\textbf{Raw benchmark results of 23 tabular generators on the ``Maternal'' dataset.} We report the mean $\pm$ std of each metric across 10 repeated data splits. For benchmark generators, ``$-$'' denotes failed convergence of a specific model or unexpected values in the synthetic data that caused the evaluation metric computation to crash. We highlight the {\color[HTML]{008080} \textbf{First}}, {\color[HTML]{7030A0} \textbf{Second}}, and {\color[HTML]{C65911} \textbf{Third}} best performances for each metric. \modelname generally achieves competitive performance against the benchmark generators while maintaining a reduced risk of overfitting.}
\resizebox{\textwidth}{!}{

}
\end{table}

\begin{table}[!htbp]
\centering
\caption{\textbf{Raw benchmark results of 23 tabular generators on the ``Miami'' dataset.} We report the mean $\pm$ std of each metric across 10 repeated data splits. For benchmark generators, ``$-$'' denotes failed convergence of a specific model or unexpected values in the synthetic data that caused the evaluation metric computation to crash. We highlight the {\color[HTML]{008080} \textbf{First}}, {\color[HTML]{7030A0} \textbf{Second}}, and {\color[HTML]{C65911} \textbf{Third}} best performances for each metric. \modelname generally achieves competitive performance against the benchmark generators while maintaining a reduced risk of overfitting.}
\resizebox{\textwidth}{!}{

}
\end{table}

\begin{table}[!htbp]
\centering
\caption{\textbf{Raw benchmark results of 23 tabular generators on the ``NATICUSdroid'' dataset.} We report the mean $\pm$ std of each metric across 10 repeated data splits. For benchmark generators, ``$-$'' denotes failed convergence of a specific model or unexpected values in the synthetic data that caused the evaluation metric computation to crash. We highlight the {\color[HTML]{008080} \textbf{First}}, {\color[HTML]{7030A0} \textbf{Second}}, and {\color[HTML]{C65911} \textbf{Third}} best performances for each metric. \modelname generally achieves competitive performance against the benchmark generators while maintaining a reduced risk of overfitting.}
\resizebox{\textwidth}{!}{

}
\end{table}

\begin{table}[!htbp]
\centering
\caption{\textbf{Raw benchmark results of 23 tabular generators on the ``Nomao'' dataset.} We report the mean $\pm$ std of each metric across 10 repeated data splits. For benchmark generators, ``$-$'' denotes failed convergence of a specific model or unexpected values in the synthetic data that caused the evaluation metric computation to crash. We highlight the {\color[HTML]{008080} \textbf{First}}, {\color[HTML]{7030A0} \textbf{Second}}, and {\color[HTML]{C65911} \textbf{Third}} best performances for each metric. \modelname generally achieves competitive performance against the benchmark generators while maintaining a reduced risk of overfitting.}
\resizebox{\textwidth}{!}{

}
\end{table}

\begin{table}[!htbp]
\centering
\caption{\textbf{Raw benchmark results of 23 tabular generators on the ``Phoneme'' dataset.} We report the mean $\pm$ std of each metric across 10 repeated data splits. For benchmark generators, ``$-$'' denotes failed convergence of a specific model or unexpected values in the synthetic data that caused the evaluation metric computation to crash. We highlight the {\color[HTML]{008080} \textbf{First}}, {\color[HTML]{7030A0} \textbf{Second}}, and {\color[HTML]{C65911} \textbf{Third}} best performances for each metric. \modelname generally achieves competitive performance against the benchmark generators while maintaining a reduced risk of overfitting.}
\resizebox{\textwidth}{!}{

}
\end{table}

\begin{table}[!htbp]
\centering
\caption{\textbf{Raw benchmark results of 23 tabular generators on the ``Plants'' dataset.} We report the mean $\pm$ std of each metric across 10 repeated data splits. For benchmark generators, ``$-$'' denotes failed convergence of a specific model or unexpected values in the synthetic data that caused the evaluation metric computation to crash. We highlight the {\color[HTML]{008080} \textbf{First}}, {\color[HTML]{7030A0} \textbf{Second}}, and {\color[HTML]{C65911} \textbf{Third}} best performances for each metric. \modelname generally achieves competitive performance against the benchmark generators while maintaining a reduced risk of overfitting.}
\resizebox{\textwidth}{!}{

}
\end{table}

\begin{table}[!htbp]
\centering
\caption{\textbf{Raw benchmark results of 23 tabular generators on the ``SDSS17'' dataset.} We report the mean $\pm$ std of each metric across 10 repeated data splits. For benchmark generators, ``$-$'' denotes failed convergence of a specific model or unexpected values in the synthetic data that caused the evaluation metric computation to crash. We highlight the {\color[HTML]{008080} \textbf{First}}, {\color[HTML]{7030A0} \textbf{Second}}, and {\color[HTML]{C65911} \textbf{Third}} best performances for each metric. \modelname generally achieves competitive performance against the benchmark generators while maintaining a reduced risk of overfitting.}
\resizebox{\textwidth}{!}{

}
\end{table}

\begin{table}[!htbp]
\centering
\caption{\textbf{Raw benchmark results of 23 tabular generators on the ``Satisfaction'' dataset.} We report the mean $\pm$ std of each metric across 10 repeated data splits. For benchmark generators, ``$-$'' denotes failed convergence of a specific model or unexpected values in the synthetic data that caused the evaluation metric computation to crash. We highlight the {\color[HTML]{008080} \textbf{First}}, {\color[HTML]{7030A0} \textbf{Second}}, and {\color[HTML]{C65911} \textbf{Third}} best performances for each metric. \modelname generally achieves competitive performance against the benchmark generators while maintaining a reduced risk of overfitting.}
\resizebox{\textwidth}{!}{

}
\end{table}

\begin{table}[!htbp]
\centering
\caption{\textbf{Raw benchmark results of 23 tabular generators on the ``Seismic'' dataset.} We report the mean $\pm$ std of each metric across 10 repeated data splits. For benchmark generators, ``$-$'' denotes failed convergence of a specific model or unexpected values in the synthetic data that caused the evaluation metric computation to crash. We highlight the {\color[HTML]{008080} \textbf{First}}, {\color[HTML]{7030A0} \textbf{Second}}, and {\color[HTML]{C65911} \textbf{Third}} best performances for each metric. \modelname generally achieves competitive performance against the benchmark generators while maintaining a reduced risk of overfitting.}
\resizebox{\textwidth}{!}{

}
\end{table}

\begin{table}[!htbp]
\centering
\caption{\textbf{Raw benchmark results of 23 tabular generators on the ``Shipping'' dataset.} We report the mean $\pm$ std of each metric across 10 repeated data splits. For benchmark generators, ``$-$'' denotes failed convergence of a specific model or unexpected values in the synthetic data that caused the evaluation metric computation to crash. We highlight the {\color[HTML]{008080} \textbf{First}}, {\color[HTML]{7030A0} \textbf{Second}}, and {\color[HTML]{C65911} \textbf{Third}} best performances for each metric. \modelname generally achieves competitive performance against the benchmark generators while maintaining a reduced risk of overfitting.}
\resizebox{\textwidth}{!}{

}
\end{table}

\begin{table}[!htbp]
\centering
\caption{\textbf{Raw benchmark results of 23 tabular generators on the ``Space'' dataset.} We report the mean $\pm$ std of each metric across 10 repeated data splits. For benchmark generators, ``$-$'' denotes failed convergence of a specific model or unexpected values in the synthetic data that caused the evaluation metric computation to crash. We highlight the {\color[HTML]{008080} \textbf{First}}, {\color[HTML]{7030A0} \textbf{Second}}, and {\color[HTML]{C65911} \textbf{Third}} best performances for each metric. \modelname generally achieves competitive performance against the benchmark generators while maintaining a reduced risk of overfitting.}
\resizebox{\textwidth}{!}{

}
\end{table}

\begin{table}[!htbp]
\centering
\caption{\textbf{Raw benchmark results of 23 tabular generators on the ``Students'' dataset.} We report the mean $\pm$ std of each metric across 10 repeated data splits. For benchmark generators, ``$-$'' denotes failed convergence of a specific model or unexpected values in the synthetic data that caused the evaluation metric computation to crash. We highlight the {\color[HTML]{008080} \textbf{First}}, {\color[HTML]{7030A0} \textbf{Second}}, and {\color[HTML]{C65911} \textbf{Third}} best performances for each metric. \modelname generally achieves competitive performance against the benchmark generators while maintaining a reduced risk of overfitting.}
\resizebox{\textwidth}{!}{

}
\end{table}

\begin{table}[!htbp]
\centering
\caption{\textbf{Raw benchmark results of 23 tabular generators on the ``Supercond'' dataset.} We report the mean $\pm$ std of each metric across 10 repeated data splits. For benchmark generators, ``$-$'' denotes failed convergence of a specific model or unexpected values in the synthetic data that caused the evaluation metric computation to crash. We highlight the {\color[HTML]{008080} \textbf{First}}, {\color[HTML]{7030A0} \textbf{Second}}, and {\color[HTML]{C65911} \textbf{Third}} best performances for each metric. \modelname generally achieves competitive performance against the benchmark generators while maintaining a reduced risk of overfitting.}
\resizebox{\textwidth}{!}{

}
\end{table}

\begin{table}[!htbp]
\centering
\caption{\textbf{Raw benchmark results of 23 tabular generators on the ``Transfusion'' dataset.} We report the mean $\pm$ std of each metric across 10 repeated data splits. For benchmark generators, ``$-$'' denotes failed convergence of a specific model or unexpected values in the synthetic data that caused the evaluation metric computation to crash. We highlight the {\color[HTML]{008080} \textbf{First}}, {\color[HTML]{7030A0} \textbf{Second}}, and {\color[HTML]{C65911} \textbf{Third}} best performances for each metric. \modelname generally achieves competitive performance against the benchmark generators while maintaining a reduced risk of overfitting.}
\resizebox{\textwidth}{!}{

}
\end{table}

\begin{table}[!htbp]
\centering
\caption{\textbf{Raw benchmark results of 23 tabular generators on the ``Vehicle'' dataset.} We report the mean $\pm$ std of each metric across 10 repeated data splits. For benchmark generators, ``$-$'' denotes failed convergence of a specific model or unexpected values in the synthetic data that caused the evaluation metric computation to crash. We highlight the {\color[HTML]{008080} \textbf{First}}, {\color[HTML]{7030A0} \textbf{Second}}, and {\color[HTML]{C65911} \textbf{Third}} best performances for each metric. \modelname generally achieves competitive performance against the benchmark generators while maintaining a reduced risk of overfitting.}
\resizebox{\textwidth}{!}{

}
\end{table}

\begin{table}[!htbp]
\centering
\caption{\textbf{Raw benchmark results of 23 tabular generators on the ``Wine'' dataset.} We report the mean $\pm$ std of each metric across 10 repeated data splits. For benchmark generators, ``$-$'' denotes failed convergence of a specific model or unexpected values in the synthetic data that caused the evaluation metric computation to crash. We highlight the {\color[HTML]{008080} \textbf{First}}, {\color[HTML]{7030A0} \textbf{Second}}, and {\color[HTML]{C65911} \textbf{Third}} best performances for each metric. \modelname generally achieves competitive performance against the benchmark generators while maintaining a reduced risk of overfitting.}
\resizebox{\textwidth}{!}{

}
\end{table}

\begin{table}[!htbp]
\centering
\caption{\textbf{Raw benchmark results of 23 tabular generators on the ``Zernike'' dataset.} We report the mean $\pm$ std of each metric across 10 repeated data splits. For benchmark generators, ``$-$'' denotes failed convergence of a specific model or unexpected values in the synthetic data that caused the evaluation metric computation to crash. We highlight the {\color[HTML]{008080} \textbf{First}}, {\color[HTML]{7030A0} \textbf{Second}}, and {\color[HTML]{C65911} \textbf{Third}} best performances for each metric. \modelname generally achieves competitive performance against the benchmark generators while maintaining a reduced risk of overfitting.}
\resizebox{\textwidth}{!}{

}
\end{table}

%% file: camera_ready_3_checklist.tex
\clearpage
\section*{NeurIPS Paper Checklist}

\begin{enumerate}

\item {\bf Claims}
    \item[] Question: Do the main claims made in the abstract and introduction accurately reflect the paper's contributions and scope?
    \item[] Answer: \answerYes{}
    \item[] Justification: The abstract and \Cref{section:intro} clearly state the scope of \modelname as a tabular foundation model for generative modelling, and the empirical claims are supported by experiments against 22 benchmark methods across 45 real-world datasets in~\Cref{sec:experiment}. The main claims are further qualified through ablation studies (\Cref{sec:ablation_impacts}), practicability analysis (\Cref{sec:practicability}), and broader discussion (\Cref{appendix:future_work}).

\item {\bf Limitations}
    \item[] Question: Does the paper discuss the limitations of the work performed by the authors?
    \item[] Answer: \answerYes{}.
    \item[] Justification: We discuss limitations and future directions in~\Cref{sec:experiment} and~\Cref{appendix:future_work}, including the current pretraining scale, model size, data scale, and computation efficiency.

\item {\bf Theory assumptions and proofs}
    \item[] Question: For each theoretical result, does the paper provide the full set of assumptions and a complete (and correct) proof?
    \item[] Answer: \answerNA{}.
    \item[] Justification: The paper does not present formal theoretical results, theorems, or proofs. The mathematical content is used to define the model components, denoising dynamics, and objective functions in \Cref{sec:method} and \Cref{appendix:model_design}.

\item {\bf Experimental result reproducibility}
    \item[] Question: Does the paper fully disclose all the information needed to reproduce the main experimental results of the paper to the extent that it affects the main claims and/or conclusions of the paper (regardless of whether the code and data are provided or not)?
    \item[] Answer: \answerYes{}.
    \item[] Justification: The paper provides detailed descriptions of the model architecture, pretraining, and experimental setup in \Cref{sec:method}, \Cref{appendix:model_design}, and \Cref{appendix:reproducibility}.

\item {\bf Open access to data and code}
    \item[] Question: Does the paper provide open access to the data and code, with sufficient instructions to faithfully reproduce the main experimental results, as described in supplemental material?
    \item[] Answer: \answerYes{}.
    \item[] Justification: We provide an anonymous codebase containing the full implementation of \modelname. The datasets used in this study, including both pretraining and benchmark datasets, are publicly available, with access links provided in~\Cref{appendix:reproducibility}.

\item {\bf Experimental setting/details}
    \item[] Question: Does the paper specify all the training and test details (e.g., data splits, hyperparameters, how they were chosen, type of optimizer) necessary to understand the results?
    \item[] Answer: \answerYes{}.
    \item[] Justification: The experimental setup specifies the 43 pretraining datasets, 45 benchmark datasets, data processing pipeline, benchmark generators, optimisation strategy, and evaluation protocol in~\Cref{sec:experiment} and~\Cref{appendix:exp_setup}. Detailed model configurations and hyperparameter search spaces are provided in~\Cref{appendix:model_configurations} and~\Cref{appendix:benchmark_generator}.

\item {\bf Experiment statistical significance}
    \item[] Question: Does the paper report error bars suitably and correctly defined or other appropriate information about the statistical significance of the experiments?
    \item[] Answer: \answerYes{}.
    \item[] Justification: The paper reports normalised mean $\pm$ standard deviation values across datasets and repeated runs by default. The full results of each dataset are available in~\Cref{appendix:extended_results}.

\item {\bf Experiments compute resources}
    \item[] Question: For each experiment, does the paper provide sufficient information on the computer resources (type of compute workers, memory, time of execution) needed to reproduce the experiments?
    \item[] Answer: \answerYes{}.
    \item[] Justification: We report the compute environment, pretraining cost in~\Cref{appendix:implementation_details}. We also provide the fitting and generation time comparisons in~\Cref{sec:practicability} and~\Cref{fig:computation_time}.

\item {\bf Code of ethics}
    \item[] Question: Does the research conducted in the paper conform, in every respect, with the NeurIPS Code of Ethics \url{https://neurips.cc/public/EthicsGuidelines}?
    \item[] Answer: \answerYes{}.
    \item[] Justification: We carefully check the NeurIPS Code of Ethics, and we confirm that our work follows the Code in every respect.

\item {\bf Broader impacts}
    \item[] Question: Does the paper discuss both potential positive societal impacts and negative societal impacts of the work performed?
    \item[] Answer: \answerYes{}.
    \item[] Justification: We provide a discussion on societal impacts in~\Cref{appendix:broader_impacts}

\item {\bf Safeguards}
    \item[] Question: Does the paper describe safeguards that have been put in place for responsible release of data or models that have a high risk for misuse (e.g., pre-trained language models, image generators, or scraped datasets)?
    \item[] Answer: \answerNA{}.
    \item[] Justification: This question does not apply to our study, as all considered datasets and models are open-source and publicly available.

\item {\bf Licenses for existing assets}
    \item[] Question: Are the creators or original owners of assets (e.g., code, data, models), used in the paper, properly credited and are the license and terms of use explicitly mentioned and properly respected?
    \item[] Answer: \answerYes{}.
    \item[] Justification: In~\Cref{appendix:reproducibility}, we provide the open-source licenses followed by the creators or original owners of assets.

\item {\bf New assets}
    \item[] Question: Are new assets introduced in the paper well documented and is the documentation provided alongside the assets?
    \item[] Answer: \answerYes{}.
    \item[] Justification: We provide an anonymous codebase of our \modelname as a Python library attached to this submission. We will make it publicly available post-publication.

\item {\bf Crowdsourcing and research with human subjects}
    \item[] Question: For crowdsourcing experiments and research with human subjects, does the paper include the full text of instructions given to participants and screenshots, if applicable, as well as details about compensation (if any)? 
    \item[] Answer: \answerNA{}.
    \item[] Justification: The paper does not involve crowdsourcing experiments or research with human subject participation. All experiments are conducted on fully open-source public tabular datasets and benchmark suites.

\item {\bf Institutional review board (IRB) approvals or equivalent for research with human subjects}
    \item[] Question: Does the paper describe potential risks incurred by study participants, whether such risks were disclosed to the subjects, and whether Institutional Review Board (IRB) approvals (or an equivalent approval/review based on the requirements of your country or institution) were obtained?
    \item[] Answer: \answerNA{}.
    \item[] Justification: The paper does not involve crowdsourcing or human-subject research. Therefore, IRB approval or equivalent review is not applicable.

\item {\bf Declaration of LLM usage}
    \item[] Question: Does the paper describe the usage of LLMs if it is an important, original, or non-standard component of the core methods in this research? Note that if the LLM is used only for writing, editing, or formatting purposes and does \emph{not} impact the core methodology, scientific rigor, or originality of the research, declaration is not required.
    %this research? 
    \item[] Answer: \answerNA{}.
    \item[] Justification: \modelname, does not use LLMs as a methodological component. The paper separately describes the use of an LLM for benchmark generators in~\Cref{appendix:benchmark_generator}.

\end{enumerate}

%% file: reference.bib
@article{hollmann2025accurate,
  title={Accurate predictions on small data with a tabular foundation model},
  author={Hollmann, Noah and M{\"u}ller, Samuel and Purucker, Lennart and Krishnakumar, Arjun and K{\"o}rfer, Max and Hoo, Shi Bin and Schirrmeister, Robin Tibor and Hutter, Frank},
  journal={Nature},
  volume={637},
  number={8045},
  pages={319--326},
  year={2025},
  publisher={Nature Publishing Group UK London}
}

@article{bommasani2021opportunities,
  title={On the opportunities and risks of foundation models},
  author={Bommasani, Rishi and Hudson, Drew A and Adeli, Ehsan and Altman, Russ and Arora, Simran and von Arx, Sydney and Bernstein, Michael S and Bohg, Jeannette and Bosselut, Antoine and Brunskill, Emma and others},
  journal={arXiv preprint arXiv:2108.07258},
  year={2021}
}

@inproceedings{van2024position,
  title={Position: why tabular foundation models should be a research priority},
  author={Van Breugel, Boris and Van Der Schaar, Mihaela},
  booktitle={Proceedings of the 41st International Conference on Machine Learning},
  pages={48976--48993},
  year={2024}
}

@inproceedings{ma2025tabdpt,
  title={{TabDPT}: Scaling Tabular Foundation Models on Real Data},
  author={Ma, Junwei and Thomas, Valentin and Hosseinzadeh, Rasa and Labach, Alex and Cresswell, Jesse C and Golestan, Keyvan and Yu, Guangwei and Caterini, Anthony L and Volkovs, Maksims},
  booktitle={The Thirty-ninth Annual Conference on Neural Information Processing Systems},
  year={2025}
}

@article{zhang2025limix,
  title={{LimiX}: Unleashing structured-data modeling capability for generalist intelligence},
  author={Zhang, Xingxuan and Ren, Gang and Yu, Han and Yuan, Hao and Wang, Hui and Li, Jiansheng and Wu, Jiayun and Mo, Lang and Mao, Li and Hao, Mingchao and others},
  journal={arXiv preprint arXiv:2509.03505},
  year={2025}
}

@inproceedings{zhang2025mitra,
  title={Mitra: Mixed Synthetic Priors for Enhancing Tabular Foundation Models},
  author={Zhang, Xiyuan and Maddix, Danielle C and Yin, Junming and Erickson, Nick and Ansari, Abdul Fatir and Han, Boran and Zhang, Shuai and Akoglu, Leman and Faloutsos, Christos and Mahoney, Michael W and others},
  booktitle={The Thirty-ninth Annual Conference on Neural Information Processing Systems},
  year={2025}
}

@article{qu2026tabiclv2,
  title={{TabICLv2}: A better, faster, scalable, and open tabular foundation model},
  author={Qu, Jingang and Holzm{\"u}ller, David and Varoquaux, Ga{\"e}l and Morvan, Marine Le},
  journal={arXiv preprint arXiv:2602.11139},
  year={2026}
}

@article{grinsztajn2025tabpfn,
  title={{TabPFN-2.5}: Advancing the state of the art in tabular foundation models},
  author={Grinsztajn, L{\'e}o and Fl{\"o}ge, Klemens and Key, Oscar and Birkel, Felix and Jund, Philipp and Roof, Brendan and J{\"a}ger, Benjamin and Safaric, Dominik and Alessi, Simone and Hayler, Adrian and others},
  journal={arXiv preprint arXiv:2511.08667},
  year={2025}
}

@article{borisov2022deep,
  title={Deep neural networks and tabular data: A survey},
  author={Borisov, Vadim and Leemann, Tobias and Se{\ss}ler, Kathrin and Haug, Johannes and Pawelczyk, Martin and Kasneci, Gjergji},
  journal={IEEE transactions on neural networks and learning systems},
  volume={35},
  number={6},
  pages={7499--7519},
  year={2022},
  publisher={IEEE}
}

@article{shwartz2022tabular,
  title={Tabular data: Deep learning is not all you need},
  author={Shwartz-Ziv, Ravid and Armon, Amitai},
  journal={Information fusion},
  volume={81},
  pages={84--90},
  year={2022},
  publisher={Elsevier}
}

@article{gorishniy2021revisiting,
  title={Revisiting deep learning models for tabular data},
  author={Gorishniy, Yury and Rubachev, Ivan and Khrulkov, Valentin and Babenko, Artem},
  journal={Advances in neural information processing systems},
  volume={34},
  pages={18932--18943},
  year={2021}
}

@article{jiang2026representation,
  title={Representation learning for tabular data: A comprehensive survey},
  author={Jiang, Jun-Peng and Liu, Si-Yang and Cai, Hao-Run and Zhou, Qi-Le and Ye, Han-Jia},
  journal={IEEE Transactions on Pattern Analysis and Machine Intelligence},
  year={2026},
  publisher={IEEE}
}

@article{somvanshi2026survey,
  title={A Survey on Tabular Data: From Tree-based Methods to Tabular Deep Learning},
  author={Somvanshi, Shriyank and Das, Subasish and Javed, Syed and Antariksa, Gian and Hossain, Ahmed},
  journal={ACM Computing Surveys},
  year={2026},
  publisher={ACM New York, NY}
}

@inproceedings{jiang2024protogate,
  title={{ProtoGate}: prototype-based neural networks with global-to-local feature selection for tabular biomedical data},
  author={Jiang, Xiangjian and Margeloiu, Andrei and Simidjievski, Nikola and Jamnik, Mateja},
  booktitle={Proceedings of the 41st International Conference on Machine Learning},
  pages={21844--21878},
  year={2024}
}

@article{hernandez2022synthetic,
  title={Synthetic data generation for tabular health records: A systematic review},
  author={Hernandez, Mikel and Epelde, Gorka and Alberdi, Ane and Cilla, Rodrigo and Rankin, Debbie},
  journal={Neurocomputing},
  volume={493},
  pages={28--45},
  year={2022},
  publisher={Elsevier}
}

@inproceedings{sattarov2023findiff,
  title={Findiff: Diffusion models for financial tabular data generation},
  author={Sattarov, Timur and Schreyer, Marco and Borth, Damian},
  booktitle={Proceedings of the Fourth ACM International Conference on AI in Finance},
  pages={64--72},
  year={2023}
}

@article{margeloiu2024tabebm,
  title={{TabEBM}: A tabular data augmentation method with distinct class-specific energy-based models},
  author={Margeloiu, Andrei and Jiang, Xiangjian and Simidjievski, Nikola and Jamnik, Mateja},
  journal={Advances in Neural Information Processing Systems},
  volume={37},
  pages={72094--72144},
  year={2024}
}

@article{shi2025comprehensive,
  title={A comprehensive survey of synthetic tabular data generation},
  author={Shi, Ruxue and Wang, Yili and Du, Mengnan and Shen, Xu and Chang, Yi and Wang, Xin},
  journal={arXiv preprint arXiv:2504.16506},
  year={2025}
}

@inproceedings{jiang2026tabstruct,
  title={{TabStruct}: Measuring Structural Fidelity of Tabular Data},
  author={Jiang, Xiangjian and Simidjievski, Nikola and Jamnik, Mateja},
  booktitle={The Fourteenth International Conference on Learning Representations},
  year={2026}
}

@article{bond2021deep,
  title={Deep generative modelling: A comparative review of vaes, gans, normalizing flows, energy-based and autoregressive models},
  author={Bond-Taylor, Sam and Leach, Adam and Long, Yang and Willcocks, Chris G},
  journal={IEEE transactions on pattern analysis and machine intelligence},
  volume={44},
  number={11},
  pages={7327--7347},
  year={2021},
  publisher={IEEE}
}

@article{salakhutdinov2015learning,
  title={Learning deep generative models},
  author={Salakhutdinov, Ruslan},
  journal={Annual Review of Statistics and Its Application},
  volume={2},
  number={1},
  pages={361--385},
  year={2015},
  publisher={Annual Reviews}
}

@article{fang2024large,
  title={Large Language Models (LLMs) on Tabular Data: Prediction, Generation, and Understanding-A Survey},
  author={Fang, Xi and Xu, Weijie and Tan, Fiona Anting and Hu, Ziqing and Zhang, Jiani and Qi, Yanjun and Sengamedu, Srinivasan H and Faloutsos, Christos},
  journal={Transactions on Machine Learning Research},
  year={2024}
}

@article{xu2019modeling,
  title={Modeling tabular data using conditional gan},
  author={Xu, Lei and Skoularidou, Maria and Cuesta-Infante, Alfredo and Veeramachaneni, Kalyan},
  journal={Advances in neural information processing systems},
  volume={32},
  year={2019}
}

@inproceedings{zhang2023mixed,
  title={Mixed-Type Tabular Data Synthesis with Score-based Diffusion in Latent Space},
  author={Zhang, Hengrui and Zhang, Jiani and Shen, Zhengyuan and Srinivasan, Balasubramaniam and Qin, Xiao and Faloutsos, Christos and Rangwala, Huzefa and Karypis, George},
  booktitle={The Twelfth International Conference on Learning Representations},
  year={2023}
}

@inproceedings{shi2025tabdiff,
  title={{TabDiff}: a Mixed-type Diffusion Model for Tabular Data Generation},
  author={Shi, Juntong and Xu, Minkai and Hua, Harper and Zhang, Hengrui and Ermon, Stefano and Leskovec, Jure},
  booktitle={The Thirteenth International Conference on Learning Representations},
  year={2025},
}

@article{achiam2023gpt,
  title={Gpt-4 technical report},
  author={Achiam, Josh and Adler, Steven and Agarwal, Sandhini and Ahmad, Lama and Akkaya, Ilge and Aleman, Florencia Leoni and Almeida, Diogo and Altenschmidt, Janko and Altman, Sam and Anadkat, Shyamal and others},
  journal={arXiv preprint arXiv:2303.08774},
  year={2023}
}

@article{guo2025deepseek,
  title={DeepSeek-R1 incentivizes reasoning in LLMs through reinforcement learning},
  author={Guo, Daya and Yang, Dejian and Zhang, Haowei and Song, Junxiao and Wang, Peiyi and Zhu, Qihao and Xu, Runxin and Zhang, Ruoyu and Ma, Shirong and Bi, Xiao and others},
  journal={Nature},
  volume={645},
  number={8081},
  pages={633--638},
  year={2025},
  publisher={Nature Publishing Group UK London}
}

@article{lu2025vision,
  title={Vision foundation models in remote sensing: A survey},
  author={Lu, Siqi and Guo, Junlin and Zimmer-Dauphinee, James R and Nieusma, Jordan M and Wang, Xiao and VanValkenburgh, Parker and Wernke, Steven A and Huo, Yuankai},
  journal={IEEE Geoscience and Remote Sensing Magazine},
  volume={13},
  number={3},
  pages={190--215},
  year={2025},
  publisher={IEEE}
}

@article{anisuzzaman2025fine,
  title={Fine-tuning large language models for specialized use cases},
  author={Anisuzzaman, DM and Malins, Jeffrey G and Friedman, Paul A and Attia, Zachi I},
  journal={Mayo Clinic Proceedings: Digital Health},
  volume={3},
  number={1},
  pages={100184},
  year={2025},
  publisher={Elsevier}
}

@inproceedings{borisov2022language,
  title={Language Models are Realistic Tabular Data Generators},
  author={Borisov, Vadim and Sessler, Kathrin and Leemann, Tobias and Pawelczyk, Martin and Kasneci, Gjergji},
  booktitle={The Eleventh International Conference on Learning Representations},
  year={2022}
}

@inproceedings{
hu2022lora,
title={Lo{RA}: Low-Rank Adaptation of Large Language Models},
author={Edward J Hu and Yelong Shen and Phillip Wallis and Zeyuan Allen-Zhu and Yuanzhi Li and Shean Wang and Lu Wang and Weizhu Chen},
booktitle={International Conference on Learning Representations},
year={2022},
url={https://openreview.net/forum?id=nZeVKeeFYf9}
}

@inproceedings{lin2025ctsyn,
  title={{CTSyn}: A Foundation Model for Cross Tabular Data Generation},
  author={Lin, Xiaofeng and Xu, Chenheng and Yang, Matthew and Cheng, Guang},
  booktitle={The Thirteenth International Conference on Learning Representations},
  year={2025}
}

@inproceedings{ma2023tabpfgen,
  title={TabPFGen--Tabular Data Generation with TabPFN},
  author={Ma, Junwei and Dankar, Apoorv and Stein, George and Yu, Guangwei and Caterini, Anthony},
  booktitle={NeurIPS 2023 Second Table Representation Learning Workshop},
  year={2023}
}

@inproceedings{swelam2025does,
  title={Does TabPFN Understand Causal Structures?},
  author={Swelam, Omar and Purucker, Lennart and Robertson, Jake and Raum, Hanne and Boedecker, Joschka and Hutter, Frank},
  booktitle={EurIPS 2025 Workshop: AI for Tabular Data},
  year={2025}
}

@inproceedings{zheng2026diffusion,
  title={Diffusion transformers with representation autoencoders},
  author={Zheng, Boyang and Ma, Nanye and Tong, Shengbang and Xie, Saining},
  booktitle={The Fourteenth International Conference on Learning Representations},
  year={2026}
}

@article{chawla2002smote,
  title={{SMOTE}: synthetic minority over-sampling technique},
  author={Chawla, Nitesh V and Bowyer, Kevin W and Hall, Lawrence O and Kegelmeyer, W Philip},
  journal={Journal of artificial intelligence research},
  volume={16},
  pages={321--357},
  year={2002}
}

@inproceedings{neto2025tabsds,
  title={{TabSDS}: a Lightweight, Fully Non-Parametric, and Model Free Approach for Generating Synthetic Tabular Data},
  author={Neto, Elias Chaibub},
  booktitle={Forty-second International Conference on Machine Learning},
  year={2025}
}

@inproceedings{kotelnikov2023tabddpm,
  title={{TabDDPM}: Modelling tabular data with diffusion models},
  author={Kotelnikov, Akim and Baranchuk, Dmitry and Rubachev, Ivan and Babenko, Artem},
  booktitle={International conference on machine learning},
  pages={17564--17579},
  year={2023},
  organization={PMLR}
}

@inproceedings{mueller2025continuous,
  title={Continuous Diffusion for Mixed-Type Tabular Data},
  author={Mueller, Markus and Gruber, Kathrin and Fok, Dennis},
  booktitle={The Thirteenth International Conference on Learning Representations},
  year={2025},
}

@article{qian2023synthcity,
  title={Synthcity: a benchmark framework for diverse use cases of tabular synthetic data},
  author={Qian, Zhaozhi and Davis, Rob and Van Der Schaar, Mihaela},
  journal={Advances in neural information processing systems},
  volume={36},
  pages={3173--3188},
  year={2023}
}

@inproceedings{du2025systematic,
  title={Systematic assessment of tabular data synthesis},
  author={Du, Yuntao and Li, Ninghui},
  booktitle={Proceedings of the 2025 ACM SIGSAC Conference on Computer and Communications Security},
  pages={2414--2428},
  year={2025}
}

@inproceedings{liu2023goggle,
  title={{GOGGLE}: Generative modelling for tabular data by learning relational structure},
  author={Liu, Tennison and Qian, Zhaozhi and Berrevoets, Jeroen and van der Schaar, Mihaela},
  booktitle={The Eleventh International Conference on Learning Representations},
  year={2023}
}

@inproceedings{seedat2024curated,
  title={{Curated LLM}: synergy of LLMs and data curation for tabular augmentation in low-data regimes},
  author={Seedat, Nabeel and Huynh, Nicolas and Van Breugel, Boris and Van Der Schaar, Mihaela},
  booktitle={Proceedings of the 41st International Conference on Machine Learning},
  pages={44060--44092},
  year={2024}
}

@inproceedings{ye2025closer,
  title={A Closer Look at TabPFN v2: Understanding Its Strengths and Extending Its Capabilities},
  author={Ye, Han-Jia and Liu, Si-Yang and Chao, Wei-Lun},
  booktitle={The Thirty-ninth Annual Conference on Neural Information Processing Systems},
  year={2025}
}

@inproceedings{erdogan2025layerlock,
  title={{LayerLock}: Non-collapsing Representation Learning with Progressive Freezing},
  author={Erdogan, Goker and Parthasarathy, Nikhil and Ionescu, Catalin and Hudson, Drew A and Lerchner, Alexander and Zisserman, Andrew and Sajjadi, Mehdi SM and Carreira, Joao},
  booktitle={Proceedings of the IEEE/CVF International Conference on Computer Vision},
  pages={19461--19470},
  year={2025}
}

@article{karras2022elucidating,
  title={Elucidating the design space of diffusion-based generative models},
  author={Karras, Tero and Aittala, Miika and Aila, Timo and Laine, Samuli},
  journal={Advances in neural information processing systems},
  volume={35},
  pages={26565--26577},
  year={2022}
}

@inproceedings{qu2025tabicl,
  title={{TabICL}: A Tabular Foundation Model for In-Context Learning on Large Data},
  author={Qu, Jingang and Holzm{\"u}ller, David and Varoquaux, Ga{\"e}l and Le Morvan, Marine},
  booktitle={International Conference on Machine Learning},
  pages={50817--50847},
  year={2025},
  organization={PMLR}
}

@inproceedings{erickson2025tabarena,
  title={{TabArena}: A Living Benchmark for Machine Learning on Tabular Data},
  author={Erickson, Nick and Purucker, Lennart and Tschalzev, Andrej and Holzm{\"u}ller, David and Desai, Prateek Mutalik and Salinas, David and Hutter, Frank},
  booktitle={The Thirty-ninth Annual Conference on Neural Information Processing Systems Datasets and Benchmarks Track},
  year={2025}
}

@article{hansen2023reimagining,
  title={Reimagining synthetic tabular data generation through data-centric AI: A comprehensive benchmark},
  author={Hansen, Lasse and Seedat, Nabeel and van der Schaar, Mihaela and Petrovic, Andrija},
  journal={Advances in Neural Information Processing Systems},
  volume={36},
  pages={33781--33823},
  year={2023}
}

@article{du2024systematic,
  title={Systematic Assessment of Tabular Data Synthesis Algorithms},
  author={Du, Yuntao and Li, Ninghui},
  journal={arXiv e-prints},
  pages={arXiv--2402},
  year={2024}
}

@article{tu2024causality,
  title={Causality for Tabular Data Synthesis: A High-Order Structure Causal Benchmark Framework},
  author={Tu, Ruibo and Senane, Zineb and Cao, Lele and Zhang, Cheng and Kjellstr{\"o}m, Hedvig and Henter, Gustav Eje},
  journal={arXiv preprint arXiv:2406.08311},
  year={2024}
}

@inproceedings{livieris2024evaluation,
  title={An evaluation framework for synthetic data generation models},
  author={Livieris, Ioannis E and Alimpertis, Nikos and Domalis, George and Tsakalidis, Dimitris},
  booktitle={IFIP International Conference on Artificial Intelligence Applications and Innovations},
  pages={320--335},
  year={2024},
  organization={Springer}
}

@article{lautrup2025syntheval,
  title={Syntheval: a framework for detailed utility and privacy evaluation of tabular synthetic data},
  author={Lautrup, Anton D and Hyrup, Tobias and Zimek, Arthur and Schneider-Kamp, Peter},
  journal={Data Mining and Knowledge Discovery},
  volume={39},
  number={1},
  pages={6},
  year={2025},
  publisher={Springer}
}

@article{kapar2025what,
  title={What's Wrong with Your Synthetic Tabular Data? Using Explainable AI to Evaluate Generative Models},
  author={Kapar, Jan and Koenen, Niklas and Jullum, Martin},
  journal={arXiv e-prints},
  pages={arXiv--2504},
  year={2025}
}

@article{sauber2022use,
  title={The use of generative adversarial networks to alleviate class imbalance in tabular data: a survey},
  author={Sauber-Cole, Rick and Khoshgoftaar, Taghi M},
  journal={Journal of Big Data},
  volume={9},
  number={1},
  pages={98},
  year={2022},
  publisher={Springer}
}

@article{durkan2019neural,
  title={Neural spline flows},
  author={Durkan, Conor and Bekasov, Artur and Murray, Iain and Papamakarios, George},
  journal={Advances in neural information processing systems},
  volume={32},
  year={2019}
}

@inproceedings{watson2023adversarial,
  title={Adversarial random forests for density estimation and generative modeling},
  author={Watson, David S and Blesch, Kristin and Kapar, Jan and Wright, Marvin N},
  booktitle={International Conference on Artificial Intelligence and Statistics},
  pages={5357--5375},
  year={2023},
  organization={PMLR}
}

@inproceedings{zhang2025tabnat,
  title={{TabNAT}: A continuous-discrete joint generative framework for tabular data},
  author={Zhang, Hengrui and Fang, Liancheng and Wu, Qitian and Yu, Philip S},
  booktitle={Forty-second International Conference on Machine Learning, 2025b. URL https://openreview. net/forum},
  year={2025}
}

@article{tiwald2025tabularargn,
  title={{TabularARGN}: A flexible and efficient auto-regressive framework for generating high-fidelity synthetic data},
  author={Tiwald, Paul and Krchova, Ivona and Sidorenko, Andrey and Vieyra, Mariana Vargas and Scriminaci, Mario and Platzer, Michael},
  journal={arXiv preprint arXiv:2501.12012},
  year={2025}
}

@inproceedings{nguyen2024generating,
  title={Generating realistic tabular data with large language models},
  author={Nguyen, Dang and Gupta, Sunil and Do, Kien and Nguyen, Thin and Venkatesh, Svetha},
  booktitle={2024 IEEE International Conference on Data Mining (ICDM)},
  pages={330--339},
  year={2024},
  organization={IEEE}
}

@inproceedings{zhao2025tabula,
  title={Tabula: Harnessing language models for tabular data synthesis},
  author={Zhao, Zilong and Birke, Robert and Chen, Lydia Y},
  booktitle={Pacific-Asia Conference on Knowledge Discovery and Data Mining},
  pages={247--259},
  year={2025},
  organization={Springer}
}

@article{li2024tabsal,
  title={Tabsal: Synthesizing tabular data with small agent assisted language models},
  author={Li, Jiale and Qian, Run and Tan, Yandan and Li, Zhixin and Chen, Luyu and Liu, Sen and Wu, Jie and Chai, Hongfeng},
  journal={Knowledge-Based Systems},
  volume={304},
  pages={112438},
  year={2024},
  publisher={Elsevier}
}

@misc{prior2026tabpfnext,
  title = {tabpfn-extensions: Community extensions for TabPFN, the foundation model for tabular data.},
  author = {Prior Lab},
  year = {2026},
  howpublished = {\url{https://github.com/priorlabs/tabpfn-extensions}},
}

@misc{tabddpm-codebase,
  title = {The official implementation of the paper "TabDDPM: Modelling Tabular Data with Diffusion Models"},
  author = {Yandex Research},
  year = {2023},
  howpublished = {\url{https://github.com/yandex-research/tab-ddpm}},
}

@misc{pytorch-lightning,
  title = {PyTorch Lightning},
  author = {The PyTorch Lightning team},
  year = {2026},
  howpublished = {\url{https://github.com/Lightning-AI/pytorch-lightning}},
}

@misc{ctsyn-codebase,
  title = {The official implementation of the paper "CTSyn: A Foundation Model for Cross Tabular Data Generation"},
  author = {Lin, Xiaofeng and Xu, Chenheng and Yang, Matthew and Cheng, Guang},
  year = {2025},
  howpublished = {\url{https://openreview.net/forum?id=Sh4FOyZRpv}},
}

@article{wust2011sdmetrics,
  title={SDMetrics},
  author={W{\"u}st, J{\"u}rgen},
  journal={Online: http://www. sdmetrics. com},
  year={2011}
}

@inproceedings{alaa2022faithful,
  title={How faithful is your synthetic data? sample-level metrics for evaluating and auditing generative models},
  author={Alaa, Ahmed and Van Breugel, Boris and Saveliev, Evgeny S and van der Schaar, Mihaela},
  booktitle={International Conference on Machine Learning},
  pages={290--306},
  year={2022},
  organization={PMLR}
}

@inproceedings{hu2024sok,
  title={Sok: Privacy-preserving data synthesis},
  author={Hu, Yuzheng and Wu, Fan and Li, Qinbin and Long, Yunhui and Garrido, Gonzalo Munilla and Ge, Chang and Ding, Bolin and Forsyth, David and Li, Bo and Song, Dawn},
  booktitle={2024 IEEE Symposium on Security and Privacy (SP)},
  pages={4696--4713},
  year={2024},
  organization={IEEE}
}

@article{stoian2025survey,
  title={A Survey on Tabular Data Generation: Utility, Alignment, Fidelity, Privacy, and Beyond},
  author={Stoian, Mihaela C{\"A} and Giunchiglia, Eleonora and Lukasiewicz, Thomas},
  journal={arXiv preprint arXiv:2503.05954},
  year={2025}
}

@article{truda2023generating,
  title={Generating tabular datasets under differential privacy},
  author={Truda, Gianluca},
  journal={arXiv preprint arXiv:2308.14784},
  year={2023}
}

@inproceedings{jordon2018pate,
  title={{PATE-GAN}: Generating synthetic data with differential privacy guarantees},
  author={Jordon, James and Yoon, Jinsung and Van Der Schaar, Mihaela},
  booktitle={International conference on learning representations},
  year={2018}
}

@inproceedings{mckenna2019graphical,
  title={Graphical-model based estimation and inference for differential privacy},
  author={McKenna, Ryan and Sheldon, Daniel and Miklau, Gerome},
  booktitle={International Conference on Machine Learning},
  pages={4435--4444},
  year={2019},
  organization={PMLR}
}

@inproceedings{jiang2025well,
  title={How Well Does Your Tabular Generator Learn the Structure of Tabular Data?},
  author={Jiang, Xiangjian and Simidjievski, Nikola and Jamnik, Mateja},
  booktitle={ICLR 2025 Workshop on Deep Generative Model in Machine Learning: Theory, Principle and Efficacy},
  year={2025}
}

@article{bravo2024nrgboost,
  title={{NRGBoost}: Energy-Based Generative Boosted Trees},
  author={Bravo, Jo{\~a}o},
  journal={International Conference on Learning Representations},
  year={2025}
}

@article{imbalanced-learn,
  author  = {Guillaume  Lema{{\^i}}tre and Fernando Nogueira and Christos K. Aridas},
  title   = {Imbalanced-learn: A Python Toolbox to Tackle the Curse of Imbalanced Datasets in Machine Learning},
  journal = {Journal of Machine Learning Research},
  year    = {2017},
  volume  = {18},
  number  = {17},
  pages   = {1-5},
  url     = {http://jmlr.org/papers/v18/16-365.html}
}

@article{lin2018pacgan,
  title={Pacgan: The power of two samples in generative adversarial networks},
  author={Lin, Zinan and Khetan, Ashish and Fanti, Giulia and Oh, Sewoong},
  journal={Advances in neural information processing systems},
  volume={31},
  year={2018}
}

@inproceedings{akiba2019optuna,
  title={Optuna: A next-generation hyperparameter optimization framework},
  author={Akiba, Takuya and Sano, Shotaro and Yanase, Toshihiko and Ohta, Takeru and Koyama, Masanori},
  booktitle={Proceedings of the 25th ACM SIGKDD international conference on knowledge discovery \& data mining},
  pages={2623--2631},
  year={2019}
}

@article{grinsztajn2022tree,
  title={Why do tree-based models still outperform deep learning on typical tabular data?},
  author={Grinsztajn, L{\'e}o and Oyallon, Edouard and Varoquaux, Ga{\"e}l},
  journal={Advances in neural information processing systems},
  volume={35},
  pages={507--520},
  year={2022}
}

@article{mcelfresh2024neural,
  title={When do neural nets outperform boosted trees on tabular data?},
  author={McElfresh, Duncan and Khandagale, Sujay and Valverde, Jonathan and Prasad C, Vishak and Ramakrishnan, Ganesh and Goldblum, Micah and White, Colin},
  journal={Advances in Neural Information Processing Systems},
  volume={36},
  year={2024}
}

@misc{jiang2025tabeval,
  title = {{TabEval}: A Comprehensive Evaluation Framework for Tabular Synthetic Data Generation},
  author = {Xiangjian Jiang},
  year = {2025},
  howpublished = {\url{https://github.com/SilenceX12138/TabEval}},
}

@misc{jiang2025tabcamel,
  title = {{TabCamel}: A DataFrame-focused solution for tabular datasets in machine learning workflows},
  author = {Xiangjian Jiang},
  year = {2025},
  howpublished = {\url{https://github.com/SilenceX12138/TabCamel}},
}

@article{matplotlib,
  author    = {Hunter, J. D.},
  title     = {Matplotlib: A 2D graphics environment},
  journal   = {Computing in Science \& Engineering},
  volume    = {9},
  number    = {3},
  pages     = {90--95},
  publisher = {IEEE COMPUTER SOC},
  doi       = {10.1109/MCSE.2007.55},
  year      = 2007
}

@article{paszke2019pytorch,
  title={Pytorch: An imperative style, high-performance deep learning library},
  author={Paszke, Adam and Gross, Sam and Massa, Francisco and Lerer, Adam and Bradbury, James and Chanan, Gregory and Killeen, Trevor and Lin, Zeming and Gimelshein, Natalia and Antiga, Luca and others},
  journal={Advances in neural information processing systems},
  volume={32},
  year={2019}
}

@article{dao2022flashattention,
  title={Flashattention: Fast and memory-efficient exact attention with io-awareness},
  author={Dao, Tri and Fu, Dan and Ermon, Stefano and Rudra, Atri and R{\'e}, Christopher},
  journal={Advances in neural information processing systems},
  volume={35},
  pages={16344--16359},
  year={2022}
}

@inproceedings{dao2023flashattention,
  title={FlashAttention-2: Faster Attention with Better Parallelism and Work Partitioning},
  author={Dao, Tri},
  booktitle={The Twelfth International Conference on Learning Representations},
  year={2023}
}

@article{clark2023testing,
  title={Testing causality in scientific modelling software},
  author={Clark, Andrew G and Foster, Michael and Prifling, Benedikt and Walkinshaw, Neil and Hierons, Robert M and Schmidt, Volker and Turner, Robert D},
  journal={ACM Transactions on Software Engineering and Methodology},
  volume={33},
  number={1},
  pages={1--42},
  year={2023},
  publisher={ACM New York, NY, USA}
}

@book{marwala2015causality,
  title={Causality, correlation and artificial intelligence for rational decision making},
  author={Marwala, Tshilidzi},
  year={2015},
  publisher={World Scientific}
}

@article{jacob2026tabscm,
  title={{TabSCM}: A practical Framework for Generating Realistic Tabular Data},
  author={Jacob, Sven and Prenkaj, Bardh and Shao, Weijia and Kasneci, Gjergji},
  journal={arXiv preprint arXiv:2604.22337},
  year={2026}
}

@article{kaddour2022causal,
  title={Causal machine learning: A survey and open problems},
  author={Kaddour, Jean and Lynch, Aengus and Liu, Qi and Kusner, Matt J and Silva, Ricardo},
  journal={arXiv preprint arXiv:2206.15475},
  year={2022}
}

@article{nastl2024causal,
  title={Do causal predictors generalize better to new domains?},
  author={Nastl, Vivian and Hardt, Moritz},
  journal={Advances in Neural Information Processing Systems},
  volume={37},
  pages={31202--31315},
  year={2024}
}

@article{glymour2019review,
  title={Review of causal discovery methods based on graphical models},
  author={Glymour, Clark and Zhang, Kun and Spirtes, Peter},
  journal={Frontiers in genetics},
  volume={10},
  pages={524},
  year={2019},
  publisher={Frontiers Media SA}
}

@article{kitson2023survey,
  title={A survey of Bayesian Network structure learning},
  author={Kitson, Neville Kenneth and Constantinou, Anthony C and Guo, Zhigao and Liu, Yang and Chobtham, Kiattikun},
  journal={Artificial Intelligence Review},
  volume={56},
  number={8},
  pages={8721--8814},
  year={2023},
  publisher={Springer}
}

@article{cox1958regression,
  title     = {The regression analysis of binary sequences},
  author    = {Cox, David R},
  journal   = {Journal of the Royal Statistical Society Series B: Statistical Methodology},
  volume    = {20},
  number    = {2},
  pages     = {215--232},
  year      = {1958},
  publisher = {Oxford University Press}
}

@book{fix1985discriminatory,
  title     = {Discriminatory analysis: nonparametric discrimination, consistency properties},
  author    = {Fix, Evelyn},
  volume    = {1},
  year      = {1985},
  publisher = {USAF school of Aviation Medicine}
}

@article{breiman2001random,
  title     = {Random forests},
  author    = {Breiman, Leo},
  journal   = {Machine learning},
  volume    = {45},
  number    = {1},
  pages     = {5--32},
  year      = {2001},
  publisher = {Springer}
}

@article{agtabular,
  title={AutoGluon-Tabular: Robust and Accurate AutoML for Structured Data},
  author={Erickson, Nick and Mueller, Jonas and Shirkov, Alexander and Zhang, Hang and Larroy, Pedro and Li, Mu and Smola, Alexander},
  journal={arXiv preprint arXiv:2003.06505},
  year={2020}
}

@article{ke2017lightgbm,
  title={Lightgbm: A highly efficient gradient boosting decision tree},
  author={Ke, Guolin and Meng, Qi and Finley, Thomas and Wang, Taifeng and Chen, Wei and Ma, Weidong and Ye, Qiwei and Liu, Tie-Yan},
  journal={Advances in neural information processing systems},
  volume={30},
  year={2017}
}

@article{prokhorenkova2018catboost,
  title={CatBoost: unbiased boosting with categorical features},
  author={Prokhorenkova, Liudmila and Gusev, Gleb and Vorobev, Aleksandr and Dorogush, Anna Veronika and Gulin, Andrey},
  journal={Advances in neural information processing systems},
  volume={31},
  year={2018}
}

@inproceedings{chen2016xgboost,
  title     = {Xgboost: A scalable tree boosting system},
  author    = {Chen, Tianqi and Guestrin, Carlos},
  booktitle = {Proceedings of the 22nd acm sigkdd international conference on knowledge discovery and data mining},
  pages     = {785--794},
  year      = {2016}
}

@article{gardner2023benchmarking,
  title={Benchmarking distribution shift in tabular data with tableshift},
  author={Gardner, Josh and Popovic, Zoran and Schmidt, Ludwig},
  journal={Advances in Neural Information Processing Systems},
  volume={36},
  pages={53385--53432},
  year={2023}
}

@article{purucker2026beyond,
  title={Beyond IID: How General Are Tabular Foundation Models, Really?},
  author={Purucker, Lennart and Tschalzev, Andrej and Erickson, Nick and Blayer, Gioia and Holzm{\~A}$1/4$ller, David and Arazi, Alan and Pfefferle, Alexander and Tajjar, Mustafa and Varoquaux, Ga{\~A} and Hutter, Frank and others},
  journal={arXiv preprint arXiv:2606.30410},
  year={2026}
}

@article{hoo2025tables,
  title={From tables to time: Extending tabpfn-v2 to time series forecasting},
  author={Hoo, Shi Bin and M{\"u}ller, Samuel and Salinas, David and Hutter, Frank},
  journal={arXiv preprint arXiv:2501.02945},
  year={2025}
}

@inproceedings{habib2026gotabpfn,
  title     = {{GOTabPFN: From Feature Ordering to Compact Tokenization for Tabular Foundation Models on High-Dimensional Data}},
  author    = {Habib, Al Zadid Sultan Bin and Ahamed, Md Younus and Gyawali, Prashnna and Doretto, Gianfranco and Adjeroh, Donald A.},
  booktitle = {Proceedings of the 43rd International Conference on Machine Learning},
  year      = {2026}
}

@article{chan2024autocd,
  title={AutoCD: Automated Machine Learning for Causal Discovery Algorithms},
  author={Chan, Gerlise and Claassen, Tom and Hoos, Holger and Heskes, Tom and Baratchi, Mitra},
  year={2024},
  journal={Sl: OpenReview}
}

@article{sun2026survey,
  title={A Survey on Foundation Models for Structured Data: Tabular, Time Series, and Graphs},
  author={Sun, Qingyun and Yuan, Haonan and Huang, Yi and Zhang, Ziwei and Fu, Xingcheng and Wang, Ruijie and Zhou, Haoyi and Li, Jianxin and Wu, Jia and Yu, Philip S},
  journal={IEEE Transactions on Knowledge and Data Engineering},
  year={2026},
  publisher={IEEE}
}

@article{wang2025graph,
  title={Graph foundation models: A comprehensive survey},
  author={Wang, Zehong and Liu, Zheyuan and Ma, Tianyi and Li, Jiazheng and Zhang, Zheyuan and Fu, Xingbo and Li, Yiyang and Yuan, Zhengqing and Song, Wei and Ma, Yijun and others},
  journal={arXiv preprint arXiv:2505.15116},
  year={2025}
}

@article{arazi2026tabstar,
  title={Tabstar: A tabular foundation model for tabular data with text fields},
  author={Arazi, Alan and Shapira, Eilam and Reichart, Roi},
  journal={Advances in Neural Information Processing Systems},
  volume={38},
  pages={172108--172161},
  year={2026}
}

@inproceedings{bonet2026iltm,
  title={iLTM: Integrated large tabular model},
  author={Bonet, David and Comajoan Cara, Mar{\c{c}}al and Calafell, Alvaro and Mas Montserrat, Daniel and Ioannidis, Alexander G},
  booktitle={Proceedings of the 32nd ACM SIGKDD Conference on Knowledge Discovery and Data Mining V. 2},
  pages={186--197},
  year={2026}
}

@article{van2024latable,
  title={LaTable: towards large tabular models},
  author={van Breugel, Boris and Crabb{\'e}, Jonathan and Davis, Rob and van der Schaar, Mihaela},
  journal={arXiv preprint arXiv:2406.17673},
  year={2024}
}

@inproceedings{sanchez2022vaca,
  title={Vaca: Designing variational graph autoencoders for causal queries},
  author={S{\'a}nchez-Martin, Pablo and Rateike, Miriam and Valera, Isabel},
  booktitle={Proceedings of the AAAI Conference on Artificial Intelligence},
  volume={36},
  pages={8159--8168},
  year={2022}
}

@inproceedings{khemakhem2021causal,
  title={Causal autoregressive flows},
  author={Khemakhem, Ilyes and Monti, Ricardo and Leech, Robert and Hyvarinen, Aapo},
  booktitle={International conference on artificial intelligence and statistics},
  pages={3520--3528},
  year={2021},
  organization={PMLR}
}

@article{chao2023modeling,
  title={Modeling causal mechanisms with diffusion models for interventional and counterfactual queries},
  author={Chao, Patrick and Bl{\"o}baum, Patrick and Patel, Sapan and Kasiviswanathan, Shiva Prasad},
  journal={arXiv preprint arXiv:2302.00860},
  year={2023}
}

@inproceedings{fu2010markov,
  title={Markov blanket based feature selection: a review of past decade},
  author={Fu, Shunkai and Desmarais, Michel C},
  booktitle={Proceedings of the world congress on engineering},
  volume={1},
  pages={321--328},
  year={2010},
  organization={Newswood Ltd. Hong Kong, China}
}

@article{cherepanova2023performance,
  title={A performance-driven benchmark for feature selection in tabular deep learning},
  author={Cherepanova, Valeriia and Levin, Roman and Somepalli, Gowthami and Geiping, Jonas and Bruss, C Bayan and Wilson, Andrew G and Goldstein, Tom and Goldblum, Micah},
  journal={Advances in Neural Information Processing Systems},
  volume={36},
  pages={41956--41979},
  year={2023}
}

@article{tugnoli2026improving,
  title={Improving TabPFN's Synthetic Data Generation by Integrating Causal Structure},
  author={Tugnoli, Davide and De Lorenzo, Andrea and Virgolin, Marco and Cin{\`a}, Giovanni},
  journal={arXiv preprint arXiv:2603.10254},
  year={2026}
}
